\documentclass[12pt,fleqn,a4paper,ChapStyle1,twoside,openright]{book}

\usepackage{etex}
\usepackage{url}
\usepackage{amsmath,amssymb,amscd,amsbsy,array,color,epsfig}
\usepackage{fancyhdr,framed,latexsym,multicol,pstricks,slashed,xcolor}
\usepackage{picture}
\usepackage{indentfirst}
\usepackage{enumitem}
\usepackage{amsmath}
\usepackage{amssymb}
\usepackage{pdfpages}
\usepackage{rotating}
\usepackage{lscape}
\usepackage{amsthm}
\usepackage{makeidx} 
\usepackage{fmtcount}
\usepackage{textcomp}
\usepackage{fancyhdr}
\usepackage{txfonts} \usepackage{algorithm,algorithmic}

\usepackage{multirow}
\usepackage{minitoc}
\usepackage{graphicx}
\usepackage{amsmath}
\usepackage{epsfig}
\usepackage{epstopdf}
\usepackage{subfig}
\usepackage{fancyhdr}
\usepackage{multirow}
\usepackage{algorithm}
\usepackage{algorithmic}

\renewcommand{\chaptermark}[1]%
{\markboth{\itshape{Chapter~\thechapter~: \ #1}}{}}
\renewcommand{\sectionmark}[1]%
{\markright{\itshape{Section~\thesection~--~\ #1}}}
\usepackage{latexsym}
\usepackage{amssymb, euscript} 
\usepackage{calc}
\usepackage{lettrine}
\usepackage{a4wide}
\usepackage{verbatim}
\usepackage{tabularx}
\usepackage{pict2e}
\usepackage{floatflt}
\usepackage{moreverb}
\usepackage{pdfpages}

\newtheorem{example}{Example} [chapter]
\newtheorem{c-Example}{Counter-example} [chapter]
\newtheorem{proposition}{Proposition} [chapter]
\newtheorem{definition}{Definition} [chapter]

\usepackage{fncychap}
\usepackage{color}
\usepackage{tikz}
\usepackage{forest}
\usepackage{lscape}
\usepackage{tikz-qtree}
\usepackage{graphicx}
\usepackage{nameref}
\usepackage{colortbl}
\usepackage{minitoc}
\usepackage{mdwmath}
\usepackage{amsmath}
\usepackage{mdwtab}
\usepackage{forest}
\usepackage{booktabs}
\usepackage{lscape}
\usetikzlibrary{calc}
\usetikzlibrary{snakes,arrows,shapes}
\usepgflibrary{shapes.multipart}
\usetikzlibrary{trees}
\ChNumVar{\fontsize{76}{80}\usefont{OT1}{pzc}{m}{n}
\selectfont}
\ChTitleVar{\raggedleft\Large\sffamily\bfseries}
\usepackage{float}
\usepackage{caption}
\usepackage{array,multirow,makecell}
\setcellgapes{1pt}
\makegapedcells
\newcolumntype{R}[1]{>{\raggedleft\arraybackslash }b{#1}}
\newcolumntype{L}[1]{>{\raggedright\arraybackslash }b{#1}}
\newcolumntype{C}[1]{>{\centering\arraybackslash }b{#1}}
  \makeatletter
\renewcommand{\thealgorithm}{\ifnum \c@chapter>\z@ \thechapter.\fi
 \@arabic\c@algorithm}
\@addtoreset{algorithm}{chapter}
\makeatother
\usepackage{amsmath, amsfonts}
\usepackage{dsfont}
\definecolor{coolblack}{rgb}{0.0, 0.18, 0.39}
\definecolor{bluegray}{rgb}{0.4, 0.6, 0.8}
\definecolor{ceruleanblue}{rgb}{0.16, 0.32, 0.75}
\definecolor{light-gray}{gray}{0.95}
\definecolor{mongris}{gray}{0.75}

\usepackage{pgfplots}

\usepackage{pst-grad,pst-node,pstricks,tikz}
\usetikzlibrary{fit}
\usepackage{pgflibraryshapes}
\usepackage{pgfplots}
 \usepgfplotslibrary{external}
 \usepgfplotslibrary{units}
\usepackage{pgf, tikz}
\usepackage{tikz}
\tikzset{
  treenode/.style = {shape=rectangle, rounded corners,
                     draw, align=center,
                     top color=white, bottom color=blue!20},
  root/.style     = {treenode, font=\Large, bottom color=red!30},
  env/.style      = {treenode, font=\ttfamily\normalsize},
  dummy/.style    = {circle,draw}
}

\usetikzlibrary{arrows,shapes,positioning,shadows,trees}

\tikzset{
  basic/.style  = {draw, text width=2cm, drop shadow, font=\sffamily, rectangle},
  root/.style   = {basic, rounded corners=2pt, thin, align=center,
                   fill=green!30},
  level 2/.style = {basic, rounded corners=6pt, thin,align=center, fill=green!60,
                   text width=8em},
  level 3/.style = {basic, thin, align=left, fill=pink!60, text width=6.5em}
} 
\begin{document}

\thispagestyle{empty}
\begin{center} 
  \includegraphics[width=24mm,height=24mm]{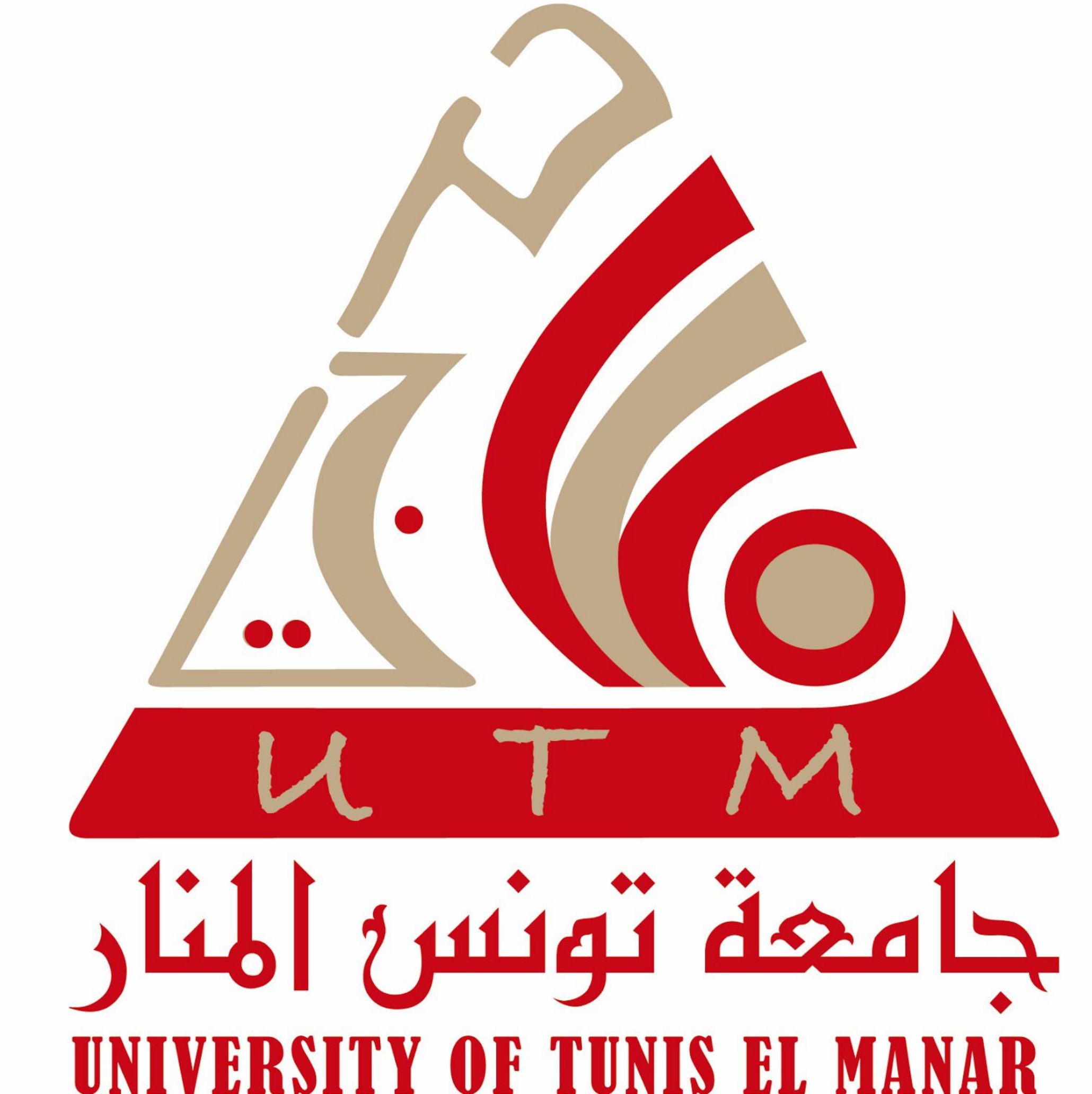}
  \hskip 20mm 
  \includegraphics[width=30mm]{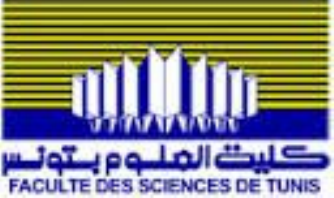} 
  \hskip 15mm 
  \includegraphics[width=45mm]{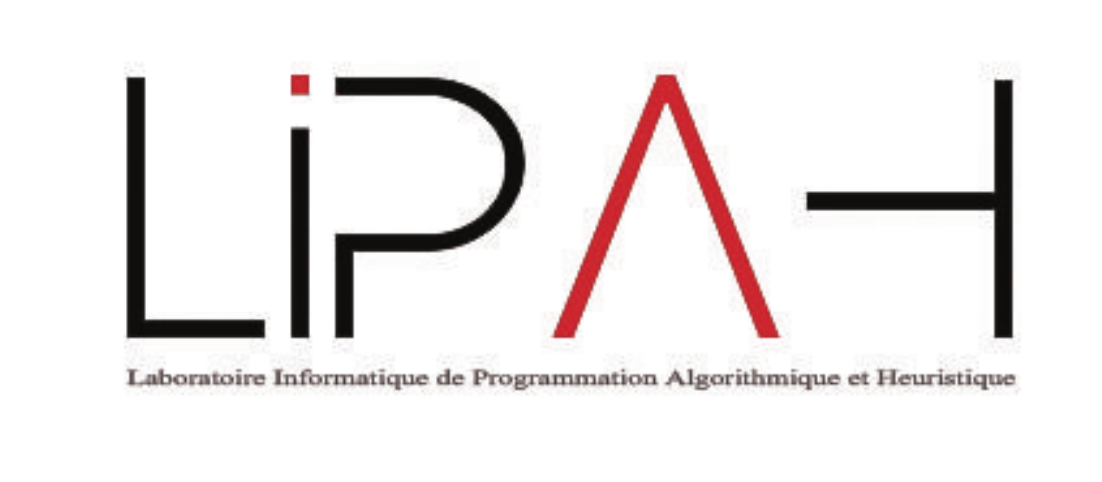}
\end{center} 
\begin{center}

\center{\large{\textbf{UNIVERSITY OF TUNIS EL MANAR}}}
\vspace{-0.3cm}
\center{\large{\textbf{FACULTY OF SCIENCES OF TUNIS}}}


\vspace{2cm}
\textbf{ \Large DOCTORAL THESIS}\\[0.3cm] 

\center{A thesis submitted in fulfillment of the requirements}
\vspace{-0.3cm}
\center{for the degree of Doctor of }
\vspace{-0.3cm}
\center{\textsc{Computer Science}}

\center{Defended by \large{\textsc{\textbf{Amina Houari}}}}


\vspace{1cm}
{\color{blue!35!black}\textbf{ \huge Contributions to Biclustering of Microarray Data Using Formal Concept Analysis}}
\vspace{0.08cm}
\vspace{1cm}
\\ \par
\hrule height 4pt
\par

\begin{table} [H]
\renewcommand{\footnoterule}{}
\renewcommand{\arraystretch}{1}
\setlength\tabcolsep{5pt}
\begin{tabular}{lll}
\\  \multicolumn{3} {l} {\small{\textsc{Defended on 07/05/2018 in front of jury members:}}}
 \\ \\ \textsc{Amel GRISSA-TOUZI}  & \textsc{Prof, National Engineering School of Tunis (ENIT)} & \textsc{President}
  \\ \textsc{Jin-Kao HAO}  & \textsc{Prof, University of Angers (France)} & \textsc{Reviewer}
  \\ \textsc{Nadjet KAMEL}  & \textsc{Prof, University of Setif (Algeria)} &\textsc{Reviewer}
  \\ \textsc{Nedra MELLOULI}  & \textsc{Prof, University of Paris-VIII (France)} & \textsc{Examiner}
  \\ \textsc{Sadok BEN YAHIA}  & \textsc{Prof, University of Tunis El Manar } & \textsc{Director}
\\ 
\end{tabular}
\end{table}

\hrule height 4pt

\vspace{-0.08cm}

\end{center}
  \footnotesize{LIPAH Research Laboratory,
  Department of Computer Science }  

\clearpage
\thispagestyle{empty}
\cleardoublepage
 \newpage

\setcounter{page}{1} \pagenumbering{roman}
\chapter*{\begin{center}
Acknowledgments
\end{center}}\normalsize
I thank all who contributed in a way or another in the completion of this thesis.

All praise to be for Allah, Lord of the worlds. I give thanks to God for protection and the ability to do this work. Alhamdulillah, Allah gave me enough strength and patience to tackle every problem with calm and ease.

I would like to express my sincere gratitude to my advisor Professor \textsc{Sadok Ben Yahia} for the continuous support of my Ph.D study and related research, for his patience, motivation, and immense knowledge. His guidance helped me in all the time of research and writing of this thesis. I could not have imagined having a better advisor and mentor for my Ph.D study. Thank you for having oriented and supported me with patience and encouragement, for your great sense of responsibility and professionalism. Thank you also for being understanding and for your support in many difficult times I lived during my PhD life.

Besides my advisor, I would like to thank the rest of my thesis committee, I thank Professor \textsc{Amel Grissa-Touzi} for agreeing to chair my thesis committee. I would like to thank Professor \textsc{Jin-Kao Hao} and Professor \textsc{Nadjet Kamel} for accepting to review my thesis report. I would like to thanks Professor \textsc{Nedra Mellouli} for participating to the thesis committee.

I want to express my special thanks to Dr. \textsc{Wassim Ayadi} for collaborating in the realization of my thesis project. I am very grateful for all the offered efforts to ensure the high-quality of this research.

I would like to thank MESRS (Minist\`{e}re Alg\'{e}rien de l'Enseignement Sup\'{e}rieur et de la Recherche Scientifique) for funding my PhD thesis.
 
My sincere thanks go to all the members of the LIPAH laboratory of the Faculty of Sciences of Tunis, for the encouraging ambiance and emotional atmosphere during my years in Tunisia.

Many thanks to all my friends who have been around these four years, who have  been  helping  me  at different occasions and in one way or another have influenced this thesis. In particular, I would like to thank my housemates: \textsc{Amina} and \textsc{Meriem}, for creating a nice, quiet living atmosphere, and for their friendship. My thanks also go to my colleagues and friends \textsc{Amina Jarraya}, \textsc{Malek Hajjem}, \textsc{Fatma Dhaou}, \textsc{Thoraya ben Chattah}, \textsc{Soumaya Guesmi} and \textsc{Mehdi Zitouni} for their continuous support. To them, I express my deepest gratitude, friendship and love, and wish them the greatest of success.

I would never forget all the chats and beautiful moments I shared with some of my cousins, friends and classmates in Algeria. They were fundamental in supporting me during these stressful and difficult moments. Other close friends I would like to mention: \textsc{Hadia}, \textsc{Hanane} and \textsc{Houda}.

 I wish to express my appreciation to my brothers and their wives, for supporting me spiritually throughout writing this thesis and my life in general.

 Special mention goes to my joyful and lovely nephews and nieces \textsc{Sohaib, Sarah, Sabrina, Salssabil, Youcef, Mayssa, Ayoub, Ismahan, Ahmed, Safaa and Anfel}, for their love and "dua". In their company I forget all worries. I just adore you!

And last but not the least, my dear parents, thanks for giving me solid roots from which I can grow, for being a constant and active presence in my life, for accepting my dreams and for doing your best in helping me realize them. Mama you taught me that a good education gives me wings, and by personal example, instilled in me this love of learning; never-endings thanks for you.

 To those who did not feel appreciated at that time - I apologize.

\newpage
\begin{flushright}
\textit{To my parents for their endless love.\\
I am especially thankful for your love \\and your continuous
support.\\ Thanks for always believing in me.\\
I know you wait this special day from my birth.
}
\end{flushright} 

\newpage
\def\cleardoublepage{\clearpage
 \if@twoside
  \ifodd\c@page\else
   \null\thispagestyle{empty}\newpage
   \if@twocolumn\null\newpage\fi
   \fi
  \fi
 }%
\def\ps@chapterverso{\ps@empty}%
\chapter*{\begin{center}
Abstract
\end{center}}\normalsize

{\large Biclustering is an unsupervised data mining technique that aims to unveil patterns (biclusters) from gene expression data matrices. In the framework of this thesis, we propose new biclustering algorithms for microarray data. The latter is done using data mining techniques. The objective is to identify positively and negatively correlated biclusters.

This thesis is divided into two part: In the first part, we present an overview of the pattern-mining techniques and the biclustering of microarray data. In the second part, we present our proposed biclustering algorithms where we rely on two axes. In the first axis, we initially focus on extracting biclusters of positive correlations. For this, we use both Formal Concept Analysis and Association Rules. In the second axis, we focus on the extraction of negatively correlated biclusters.

The performed experimental studies highlight the very promising results offered by the proposed algorithms. Our biclustering algorithms are evaluated and compared statistically and biologically.


\textbf{Key words:} Data mining, Bioinformatics, Microarray data analysis, Biclustering, Formal Concept Analysis, Positive correlations, Negative correlations, Association Rules.
} 
\newpage
\def\cleardoublepage{\clearpage
 \if@twoside
  \ifodd\c@page\else
   \null\thispagestyle{empty}\newpage
   \if@twocolumn\null\newpage\fi
   \fi
  \fi
 }%
\def\ps@chapterverso{\ps@empty}%
 \chapter*{\begin{center}
R\'{e}sum\'{e}
\end{center}}\normalsize
{\large
Le biclustering est une technique de fouille de donn\'{e}es non supervis\'{e}e qui vise \`{a} d\'{e}voiler des motifs (biclusters) \`{a} partir des donn\'{e}es biopuces. Dans le cadre de cette th\`{e}se, nous avons propos\'{e} de nouveaux algorithmes de biclustering pour analyser les donn\'{e}es d'expressions de g\`{e}nes \`{a} travers les techniques de fouille de donn\'{e}es. L'objectif est d'identifier des biclusters positivement et n\'{e}gativement corr\'{e}l\'{e}s.

Cette th\`{e}se est d\'{e}visie en deux parties: Dans la premi\`{e}re partie, nous pr\'{e}sentons un aper\c{c}u sur les techniques de fouilles de donn\'{e}es et sur le biclustering des donn\'{e}es biopuces. Dans la deuxi\`{e}me partie, nous pr\'{e}sentons nos algorithmes de biclustering o\`{u} nous nous appuyons sur deux axes. Dans le premier axe, nous proposons d'abord des algorithmes de biclustering permettant d'identifier des biclusters de corr\'{e}lations positives. Pour cela, nous avons utilis\'{e} l'analyse formelle de concepts et les r\`{e}gles d'associations. Dans le deuxi\`{e}me axe, nous nous sommes focalis\'{e} sur l'extraction des biclusters de corr\'{e}lations n\'{e}gatives.

L'\'{e}valuation exp\'{e}rimentale men\'{e}e met en valeur les taux de compacit\'{e}s tr\`{e}s int\'{e}ressants offerts par les diff\'{e}rents algorithmes propos\'{e}s. L'\'{e}valuation de nos algorithmes est bas\'{e}e sur des crit\`{e}res statistique et biologique.

\textbf{Mots cl\'{e}s:} Fouille de donn\'{e}es, Bioinformatique, Analyse des donn\'{e}es biopuces, Biregroupement, Analyse Formelle de Concepts, Corr\'{e}lations positives, Corr\'{e}lations n\'{e}gatives, R\`{e}gles associatives.

}
\newpage
\def\cleardoublepage{\clearpage
 \if@twoside
  \ifodd\c@page\else
   \null\thispagestyle{empty}\newpage
   \if@twocolumn\null\newpage\fi
   \fi
  \fi
 }%
\def\ps@chapterverso{\ps@empty}%

\thispagestyle{empty}\null\newpage
\cleardoublepage
\thispagestyle{empty}
\dominitoc \tableofcontents \mtcaddchapter
\newpage
\def\cleardoublepage{\clearpage
 \if@twoside
  \ifodd\c@page\else
   \null\thispagestyle{empty}\newpage
   \if@twocolumn\null\newpage\fi
   \fi
  \fi
 }%
\def\ps@chapterverso{\ps@empty}%

\thispagestyle{empty}\null\newpage
\cleardoublepage
\thispagestyle{empty}
 \dominilof \listoffigures \mtcaddchapter
\cleardoublepage
\newpage
\listoftables
\newpage
 \cleardoublepage
\listofalgorithms
 \newpage
\def\cleardoublepage{\clearpage
 \if@twoside
  \ifodd\c@page\else
   \null\thispagestyle{empty}\newpage
   \if@twocolumn\null\newpage\fi
   \fi
  \fi
 }%
\def\ps@chapterverso{\ps@empty}%

\cleardoublepage
\thispagestyle{empty}

\pagenumbering{arabic}
 
\chapter*{Introduction} \markboth{Introduction}{Introduction}

\addcontentsline{toc}{chapter}{Introduction}
\section*{Context and motivations}
A biological network is a linked collection of biological entities like genes, proteins and metabolites\cite{Henriques2016}. Analyzing information and extracting biologically relevant knowledge, from these entities, is one of the key issues of bioinformatics. For instance, DNA microarray technologies help to measure the expression levels of thousands of genes under experimental conditions \cite{C.Madeira2004}. To do so, gene expression data are arranged in a data matrix. In the latter, rows represent genes, columns represent samples (experimental conditions), and each entry of the matrix denotes the expression level of a gene under a certain experimental condition. In this respect, the discovery of transcriptional modules of genes that are co-regulated in a set of experiments is of paramount importance \cite{C.Madeira2004}. Thus, we need novel ways to efficiently unveil such a type of data.

In this context, the process of Knowledge Discovery from Databases (KDD) is a whole process aiming to extract useful, hidden knowledge from a huge amount of data \cite{Agrawal1994a}. One of the main steps of this process is \textit{data mining}. This latter is dedicated to offer the necessary tools needed for data exploration. The KDD methods are widely used in the literature \cite{Fayyad1996,Fayyad1996a,Michalski1998,Othman2008,Brahmi2010,Hamdi2013,Brahmi2011}. To overcome the lack of extracted knowledge from stored data, new methods have been hence proposed, gathered under the generic term of KDD.

Interestingly enough, a first data mining technique applied to gene expression data is \textit{clustering}. In fact, the clustering techniques have been shown to be of benefit in many challenges in bioinformatics. Although useful, these approaches suffer from two major drawbacks: \textit{(i)} They consider the whole set of samples. However, genes may not be relevant to every sample. Instead, they can be relevant to only a subset of samples, which is a fundamental aspect for numerous problems in the biomedicine field \cite{Wang2002}. Thus, \textit{clustering} should be performed simultaneously on both genes and conditions.
 \textit{(ii)} Each gene can only be clustered into one group. However, many genes can belong to several clusters depending on their influence in different biological processes \cite{Gasch2002}.

 In this respect, \textit{biclustering}, which is a particular clustering type, has been palliating these drawbacks. Hence, biclustering aims to identify maximal sub-matrices (\textit{aka biclusters}) where a subset of genes expresses highly correlated behaviors over a range of conditions \cite{C.Madeira2004}. In fact, the use of biclustering in biological data is widespread owing to the presence of local patterns in them. In particular, biclustering is very relevant within the filed of gene-expression-data analysis. This includes its employment in the discovery of transcriptional modules described by correlated subsets of genes in subsets of samples.

Despite the large number of biclustering algorithms proposed in the literature, most of them are based on greedy or stochastic approaches. Furthermore, they provide sub-higher-quality answers with restrictions on the quality of obtained biclusters \cite{C.Madeira2004,Hochreiter2010,Henriques2016a}. Some attempts to palliate such drawbacks have relied on pattern-mining approaches \cite{Martinez2008,Kaytoue2011a,Mondal2012,Kaytoue2014}. Pattern-mining-based biclustering approaches aim to perform efficient and flexible searches with better solutions in terms of coherency and quality \cite{Henriques2015}. Their capabilities, among
others, include (1) an efficient search with better results guaranteed \cite{Henriques2016a}; (2) biclusters with flexible coherency strength and assumptions \cite{Henriques2014,Henriques2015a}; (3) well-designed against noise, missing values and discretization problems due to the possibility of assigning or imputing multiple symbols to a single data element; and (4) the absence of pre-fixed numbers for biclusters \cite{Serin2011}. These advantages will bring these algorithms into the spotlight when it comes to biological data analysis \cite{Martinez2008,Kaytoue2011b,Henriques2014,Henriques2014a,Kaytoue2014,Henriques2016}.

Among these pattern-mining-based algorithms are those relying on Formal Concept Analysis (FCA). FCA is a mathematical tool for analyzing data and formally representing conceptual knowledge \cite{ganter2005}. FCA helps form conceptual structures from data. Such structures consist of units, which are formal abstractions of concepts of human thought allowing meaningful and comprehensible interpretation. Interestingly enough, a distinguishing feature of FCA is an inherent integration of components of conceptual processing of data and knowledge \cite{radim2008,Mouakher2016}. Through the integration of these components, FCA's mathematical settings have recently been shown to act as a powerful tool by providing a theoretical framework for the efficient resolution of many practical problems including data mining, software engineering and information retrieval \cite{TarekYN13}.

Fundamentally concerned with FCA, it is arguably a type of biclustering method for binary data since FCA fundamentally applies to formal context (\textit{aka} binary data). \textit{Gene Expression Data} are numerical data that can be binarized in order to conciliate their objects using FCA.
Thus, motivated by this issue, we propose in this thesis to benefit from the knowledge returned from both formal concepts and association rules, especially the \textit{IGB} basis \cite{Gasmi2005} to solve the biclustering task.

\section*{Research contributions}
Biclustering has been very relevant within the field of gene-expression-data analysis. In fact, its main thrust  stands in its ability to identify groups of genes that behave in the same way under a subset of samples (conditions). With respect to this objective, a set of main contributions is presented within this dissertation and is listed as follows: 

\begin{enumerate}
\item BiARM: A new approach for the extraction of low overlapping biclusters. The driving idea is to use generic association rules and the Jaccard measure in order to remove the biclusters that have a high overlap.
\item BiFCA+: A new algorithm that aims to an efficient mining of biclusters from gene expression data. BiFCA+ heavily relies on the mathematical  background of FCA, in order to extract the biclusters' set. In addition, the \textit{Bond} correlation measure is of use to filter out the overlapping biclusters.
\item BiFCA: A new algorithm that aims for the efficient discovery of positively correlated biclusters. The main thrust of the BiFCA algorithm is the use of FCA, which has been shown to be an efficient methodology for biclustering binary data.
\item NBic-ARM: A new proposed method where we introduce a biclustering algorithm to discover biclusters of negative corrlations for gene expression data. NBic-ARM is based on generic association rules.
\item NBF: This contribution answers the same research axe. The aim is to unveil biclusters of negative correlations using FCA.
\end{enumerate}

The evaluation protocol of these contributions consists of experimental studies carried out on real-life datasets commonly used for evaluating data biclustering algorithms as well as a comparison with other approaches reported in the literature.
\section*{Thesis organization}
This Thesis is structured into two main parts.  

The first part, Theoretical aspects, is composed of two main chapters which are the following:
\begin{itemize}
\item[$\bullet$] Chapter 1: In this chapter, we present several important notions and properties that will be used in this thesis. We also recall the mathematical background of FCA; ARM and some correlations measures. 
\item[$\bullet$] Chapter 2: In this chapter, first, we describe how gene expression data are constructed to understand the data used in this thesis. Next, we detail the biclustering problem. In addition, we scrutinize pioneering work that has addressed the extraction of biclusters. We present also in this chapter some web tools of biclustering algorithms and present some statistical and biological validation. 
\end{itemize}

The second part of this thesis presents our contributions. We can split this part into two main chapters. While we dedicate the third chapter to the extraction of positively correlated biclusters, we devote the entirety of the fourth chapter to the extraction of negatively correlated ones.
\begin{itemize}
\item[$\bullet$] Chapter 3: In this chapter, we present our proposed approaches to improve the biclustering task . This is achieved through the use of ARM and FCA.     
\end{itemize}

\begin{itemize}
\item[$\bullet$] Chapter 4: In this chapter, we focus on biclustering gene expression data based on negative correlations and we present our proposed methods.
\end{itemize}

Finally, our dissertation ends with a conclusion. The conclusion summarizes all the work presented in this report and proposes further work to be done with the biclustering problem.

\part{{\Huge Theoretical aspects}}
\chapter{Overview of pattern mining}
\minitoc
\newpage
\section{Introduction}
Within the traditional framework of Formal Concept Analysis (FCA) \cite{wille1982restructuring} and Association Rule Mining (ARM)~\cite{ceglar_06}, managing the high number of frequent patterns extracted from real-life datasets becomes an important topic. In addition, providing efficient and easy-to-use tools to users is a promising challenge of data mining.

In this chapter, we present several important notions and properties that will be used in the remainder of the thesis. We also recall the mathematical background of FCA; ARM and some correlation measures.

The organization of the chapter is as follows: Section~\ref{firstchapterFCA} presents the basic definitions and mathematical settings on FCA. Section~\ref{firstchapterARM} details the association rule framework and presents some interesting measures. After that, this chapter is concluded by Section~\ref{firstchapterconclusion}.
\section{Formal Concept Analysis}
\label{firstchapterFCA}
FCA, initially introduced by Will in 1982 \cite{wille1982restructuring}, treats formal concepts. A formal concept is a set of objects to which we apply a set of attributes.

In this section, we sketch the key notions used in the remainder of this thesis. In the following, we recall some basic definitions borrowed from FCA.
\subsection{Background on Formal Concept Analysis}
\begin{definition} \textsc{\textbf{\textsc{(}Formal context\textsc{)}}}\mbox{} \\A formal context is a triplet
$\mathds{K}$ = \textsc{(}$\mathcal{O}$, $\mathcal{I}$,
$\mathcal{R}$\textsc{)}, where $\mathcal{O}$ represents a finite
set of objects, $\mathcal{I}$\ is a finite set of items (or attributes) and
$\mathcal{R}$ is a binary \textsc{(}incidence\textsc{)} relation
\textsc{(}\textit{i.e.}, $\mathcal{R}$ $\subseteq$ $ \mathcal{O}$
$\times$ $\mathcal{I}$\textsc{)}. Each couple \textsc{(}$o$,
$i$\textsc{)} $\in$ $\mathcal{R}$ expresses that the object $o$
$\in$ $\mathcal{O}$ contains the item $i \in \mathcal{I}$.
\end{definition}

\begin{example}
As on the running example, we will consider the formal context depicted by Table~\ref{table-con} with $\mathcal{O} =
\{1,2,3,4,5,6,7,8,9\}$ and $\mathcal{I}$ = $\{\texttt{a}, \texttt{b}, \texttt{c}, \texttt{d},  \texttt{e},  \texttt{f},  \texttt{g}, \texttt{h}\}$.
\begin{table}[htbp]
\begin{center}
\begin{tabular}{ccccccccc}
   & $ \texttt{a}$ & $ \texttt{b}$ & $ \texttt{c}$ & $ \texttt{d}$ & $ \texttt{e}$ & $ \texttt{f}$ & $ \texttt{g}$ & $ \texttt{h}$ \\
   \cline{2-9}

   $1$ & 0 & 1 & 0 & 0 & 0 & 0 & 1 & 0\\

   $2$ &1& 0 & 1 & 0 & 0 & 0 & 1 & 0 \\

   $3$ & 0 & 0 & 0 & 1 & 1 & 1 & 1 & 1 \\

   $4$ & 1 & 0 & 0 & 1& 1 & 1 &  1 & 1\\

   $5$ & 1& 0 & 0 & 0 & 1 & 1 &  1 & 1\\

   $6$ & 0 & 1 & 0 & 0 & 0 & 1 & 1 & 0\\

   $7$ & 1 &0 & 0 &0  & 0 & 1 & 1 &0 \\

   $8$ & 1 & 0 & 0 & 0 & 1 & 0 & 1 &0 \\

   $9$ &1 & 1 & 1 & 1 & 1 & 1 & 1 & 1 \\
   \cline{2-9}

   \end{tabular}
\end{center}
  \caption{Example of a formal context.}
  \label{table-con}
\end{table}
\end{example}

An \textit{Itemset} is a set of items, \textit{e.g.}, \{\texttt{c},  \texttt{d},  \texttt{e}\}~\footnote{In the remainder, we use a separator-free abbreviated form for the sets, e.g., $\{\texttt{cde}\}$ stands for the set of items $\{\texttt{c}, \texttt{d}, \texttt{e}$\}.} is an itemset composed by the items  \texttt{c},  \texttt{d} and  \texttt{e}. An \textit{Objset} is a set of objects, \textit{e.g.}, \{$1, {5}, {7}$\} is an objset composed of the objects ${1}$, ${5}$ and ${7}$.


Worth mentioning is the link between the power-sets $\mathcal{P}\textsc{(}\mathcal{I}\textsc{)}$ and
$\mathcal{P}\textsc{(}\mathcal{O}\textsc{)}$ associated respectively with the set of items $\mathcal{I}$ and the set of objects $\mathcal{O}$ defined as follows:

\begin{definition}\textsc{\textbf{\textsc{(}Galois connection\textsc{)}}}\mbox{}\\Let $\mathds{K}$ = \textsc{(}$\mathcal{O}$, $\mathcal{I}$, $\mathcal{R}$\textsc{)} be a formal context. The application $\psi$ is defined from the power-set of objects \textsc{(}\textit{i.e.}, ${\cal P}$\textsc{(}$\mathcal{O}$\textsc{)}\textsc{)} to the power-set of items \textsc{(}\textit{i.e.}, ${\cal P}$\textsc{(}$\mathcal{I}$\textsc{)}\textsc{)}. It associates with a set of objects $O$ the set of items $i \in \mathcal{I}$ common to all objects $o \in O$:
\begin{center}
$\psi: \mathcal{P}\textsc{(}\mathcal{O}\textsc{)}{} \rightarrow \mathcal{P}\textsc{(}\mathcal{I}\textsc{)}{}$\\$O$\mbox{ }$\mapsto$ $\psi$\textsc{(}$O$\textsc{)}\mbox{ }=\mbox{ }$\{i$ $\in$ $\mathcal{I}$ $| \forall  o \in O$, \textsc{(}$o$, $i$\textsc{)} $\in$ $\mathcal{R}\}$
\end{center}
Dually, the application $\phi$ is defined from the
power-set of items \textsc{(}\textit{i.e.}, ${\cal P}$\textsc{(}$\mathcal{I}$\textsc{)}\textsc{)} to
the power-set of objects \textsc{(}\textit{i.e.}, ${\cal
P}$\textsc{(}$\mathcal{O}$\textsc{)}\textsc{)}. It associates to a set of items $I$ the set
of objects $o \in  \mathcal{O}$ that contains all items $i$
$\in$ $I$:
\begin{center}
$\phi:\mathcal{P}\textsc{(}\mathcal{I}\textsc{)}{} \rightarrow \mathcal{P}\textsc{(}\mathcal{O}\textsc{)}{}$\\
$I$\mbox{ }$\mapsto$ $\phi$\textsc{(}$I$\textsc{)}\mbox{ }=\mbox{ }$\{o$ $\in$ $\mathcal{O}$
$|$ $\forall$ $i \in$ $I$, \textsc{(}$o$, $i$\textsc{)} $\in \mathcal{R}\}$
\end{center}
The coupled applications \textsc{(}$\psi$, $\phi$\textsc{)} form a Galois connection between the power-set of $\mathcal{O}$ and that of $\mathcal{I}$~\cite{barbut70,Ganter99}.
\end{definition}

This leads us to the definition of a formal concept.
\begin{definition}\label{formal concept}\textsc{\textbf{\textsc{(}Formal concept\textsc{)}}}\mbox{}\\A pair $\langle A, B \rangle$ $\in$
$\mathcal{O}$ $\times$ $\mathcal{I}$, of mutually corresponding
subsets, \textit{i.e.}, $\psi\textsc{(}A\textsc{)}$=$B$  and $\phi\textsc{(}B\textsc{)}$=$A$, is called a formal concept, where $A$ is called \textit{extent} and $B$ is called \textit{intent}.\\
The set of formal concepts extracted from a formal context $\mathds{K}$ = \textsc{(}$\mathcal{O}$, $\mathcal{I}$, $\mathcal{R}$\textsc{)} is denoted in the sequel  $\mathcal{C}_{\mathds{K}}$.
\end{definition}

Proposition~\ref{proposition_partial_order_formal_concepts} presents the partial order on formal concepts \textit{w.r.t.} set inclusion~\cite{Ganter99}.
\begin{proposition}\label{proposition_partial_order_formal_concepts} A partial order on formal concepts is defined as: $\forall$ $C_{1}$ = $\langle A_1, B_1 \rangle $ and $C_{2}$ = $\langle A_2, B_2 \rangle $ two formal concepts of $\mathcal{C}_{\mathds{K}}$, $C_{1}$ $\leq$
$C_{2}$ if $B_2$ $\subseteq$ $B_1$, and equivalently
$A_1$ $\subseteq$ $A_2$.
\end{proposition}

When two formal concepts fulfill the condition of Proposition~\ref{proposition_partial_order_formal_concepts}, they are said to be \textit{comparable}. Otherwise, they are said to be \textit{incomparable}. When partially sorted with set inclusion, formal concepts form a structure called the \textit{Galois \textsc{(}concept\textsc{)} lattice}, defined as follows:

\begin{definition}\textsc{\textbf{\textsc{(}Galois \textsc{(}concept\textsc{)} lattice\textsc{)}}}\mbox{}\\Given a formal context $\mathds{K}$, the
set of formal concepts $\mathcal{C}_{\mathds{K}}$  is a complete
lattice $\mathcal{L}_{\mathcal{C}_{\mathds{K}}}$, called the \textit{Galois \textsc{(}concept\textsc{)} lattice}, where $\mathcal{C}_{\mathds{K}}$ is considered with set inclusion between concepts' intents \textsc{(}or extents\textsc{)}~\cite{barbut70,Ganter99}.
\end{definition}

\subsection{Basic mathematical structures behind FCA}
\begin{definition}\label{Opérateurs de fermeture_ouverture}\textsc{\textbf{\textsc{(}Closure, Kernel operator\textsc{)}}}\mbox{}\\
Let \textsc{(}$S$,$\subseteq$\textsc{)} be a partially ordered set and $x$, $y$ be two elements of $S$. An operator $h$ defined from
\textsc{(}$S$,$\subseteq$\textsc{)} to
\textsc{(}$S$,$\subseteq$\textsc{)} is called a \textit{closure
operator} if it is:\\\textsc{(}$i$\textsc{)} \textit{Extensive};
\textit{i.e.}, $x\subseteq
h\textsc{(}x\textsc{)}$;\\\textsc{(}$ii$\textsc{)}
\textit{Isotone}; \textit{i.e.}, $x\subseteq y \Rightarrow
h\textsc{(}x\textsc{)} \subseteq h\textsc{(}y\textsc{)}$;
and \\\textsc{(}$iii$\textsc{)} \textit{Idempotent};
\textit{i.e.}, $h\textsc{(}h\textsc{(}x\textsc{)}\textsc{)}=h\textsc{(}x\textsc{)}$.\\Given the closure operator $h$ applied on the partially ordered set \textsc{(}$S$,$\subseteq$\textsc{)}, an element $x$ $\in$ $S$ is said to be \textit{closed} if its image by $h$ is equal to itself; \textit{i.e.}, $h$\textsc{(}$x$\textsc{)} = $x$.

If an operator $h'$, defined from
\textsc{(}$S$,$\subseteq$\textsc{)} to
\textsc{(}$S$,$\subseteq$\textsc{)}, is such that $h'\textsc{(}x\textsc{)}$ $\subseteq$ $x$, then $h'$ has the property to be \textit{contractive}. If it is also isotonic and idempotent, then $h'$ is said to be a \textit{kernel operator}.
\end{definition}

The following definition introduces the closure operators associated with a Galois connection.
\begin{definition}\label{Fermeture de la correspondance de Galois}\textsc{\textbf{\textsc{(}Galois closure operators\textsc{)}}}\mbox{}\\
Let us consider the power-sets $\mathcal{P}\textsc{(}\mathcal{I}\textsc{)}$ and $\mathcal{P}\textsc{(}\mathcal{O}\textsc{)}$, with the inclusion relation $\subseteq$, \textit{i.e.} the partially ordered sets $\textsc{(}\mathcal{P}\textsc{(}\mathcal{I}\textsc{)},\subseteq$\textsc{)} and $\textsc{(}\mathcal{P}\textsc{(}\mathcal{O}\textsc{)},\subseteq$\textsc{)}. The operators $\gamma$ = $\phi \circ \psi$ from $\textsc{(}\mathcal{P}\textsc{(}\mathcal{I}\textsc{)},\subseteq$\textsc{)} to $\textsc{(}\mathcal{P}\textsc{(}\mathcal{I}\textsc{)},\subseteq$\textsc{)}, and $\omega$ = $\psi \circ \phi$ from $\textsc{(}\mathcal{P}\textsc{(}\mathcal{O}\textsc{)},\subseteq$\textsc{)} to $\textsc{(}\mathcal{P}\textsc{(}\mathcal{O}\textsc{)},\subseteq$\textsc{)} are closure operators of the Galois connection~\cite{barbut70,Ganter99}. They define closure systems on $\textsc{(}\mathcal{P}\textsc{(}\mathcal{I}\textsc{)},\subseteq$\textsc{)} and $\textsc{(}\mathcal{P}\textsc{(}\mathcal{O}\textsc{)},\subseteq$\textsc{)}, respectively. The operator $\gamma$ generates closed subsets of items, while $\omega$ generates closed subsets of objects.
\end{definition}

The notion of pattern $P$ refers to either an  itemset or an objset and  is characterized by a support. The latter is detailed in the following definition.

\begin{definition}\label{supp_all}\textsc{\textbf{\textsc{(}Support of a pattern\textsc{)}}}\mbox{}\\Let $\mathds{K}$ = \textsc{(}$\mathcal{O}$, $\mathcal{I}$, $\mathcal{R}$\textsc{)} be a formal context. We define  the support  associated with a non-empty pattern $P$ as follows :

If $P$ is an itemset, then
 \begin{equation}
  \textit{Supp} \textsc{(} P\textsc{)} = |\{o\in\mathcal{O}|
\textsc{(}\forall\mbox{ } i \in P,\textsc{(}o,i\textsc{)}\in\mathcal{R}\textsc{)}\}|.
 \end{equation}

If $P$ is an objset, then
 \begin{equation}
   \textit{Supp }\textsc{(}P\textsc{)} = |\{i\in\mathcal{I}|
\textsc{(}\forall\mbox{ } o \in P,\textsc{(}o,i\textsc{)}\in\mathcal{R}\textsc{)}\}.
\end{equation}
\end{definition}

Once applied, the closure operator $\gamma$ (resp. $\omega$)  induces an equivalence relation on the power-set of items $\mathcal{P}\textsc{(}\mathcal{I}\textsc{)}$ (resp. on the power set of objects $\mathcal{P}\textsc{(}\mathcal{O}\textsc{)}$) splitting it into so-called  $\gamma$-\textit{equivalence classes} (resp. $\omega$-\textit{equivalence classes})~\cite{PASCAL00}. In each $\gamma$-equivalence class (resp. $\omega$-equivalence class), the largest itemset \textsc{(}\textit{w.r.t.} set inclusion\textsc{)} is called a \textit{closed itemset} (resp. \textit{closed objset}) while the minimal ones are called \textit{minimal generators}. The respective definitions of these particular patterns are given below.
\begin{definition}\textsc{\textbf{\textsc{(}Closed pattern\textsc{)}}}\mbox{}\\

The  itemset (resp. objset) $P$   $\subseteq \mathcal{I}$  (resp. $P$ $\subseteq \mathcal{O}$ ) is said to be \textit{closed} if $\gamma (P) =
P$ (resp. $\omega (P) =
P$)~\cite{pasquier99,Ayouni2010}

\end{definition}

\begin{definition}\label{definition_minimal_generator}\textsc{\textbf{\textsc{(}Minimal generator\textsc{)}}}\mbox{}\\

Given  a non-empty pattern $P$, two cases have to be distinguished :
\begin{itemize}
\item If $P$ is an itemset, i.e.~$ P \subseteq$ $\mathcal{I}$,  then
$ P$ is said to be a \textit{minimal
generator} of a closed itemset $I$ if $\gamma\textsc{(}P\textsc{)}$
= $I$ and $\forall$ $P_{1}$ $\subseteq \mathcal{I}$. If $P_{1}$
$\subseteq  P$ and $\gamma\textsc{(}P_1\textsc{)} = I$, then
$P$ = $P_{1}$~\cite{PASCAL00}.

\item If $P$ is an objset, i.e.~$ P \subseteq$ $\mathcal{O}$,  then
$ P$ is said to be a \textit{minimal
generator} of a closed objset $O$ when  $\omega\textsc{(}P\textsc{)}$
= $O$ and $\forall$ $P_{1}$ $\subseteq \mathcal{O}$. If $P_{1}$
$\subseteq  P$ and $\omega\textsc{(}P_1\textsc{)} = O$, then
$P$ = $P_{1}$~\cite{PASCAL00,Hamrouni2008a}.

\end{itemize}
\end{definition}

\section{Association Rule Mining}
\label{firstchapterARM}
In the following, we recall some basic definitions borrowed from Association Rule Mining.
\subsection{Association Rule framework}

As an important topic in data mining, ARM research~\cite{ceglar_06} has progressed in various directions since its inception. The formalization of the association rule extraction problem was initially introduced by Agrawal \textit{et al.}~\cite{Agra93}. The derivation of association rules is achieved starting from the set $\mathcal{FI}$ of frequent itemsets extracted from a formal context $\mathds{K}$, for a minimal support threshold \textit{minsupp}. The next definitions introduce the association rule framework.

\begin{definition}\textsc{\textbf{\textsc{(}Association rule\textsc{)}}}\mbox{}\\An association rule $R$ is a relation between itemsets and is of the form $R$: $X$ $\Rightarrow$ $\textsc{(}Y\backslash X\textsc{)}$, such that $X$ and $Y$ are
two itemsets, and $X$ $\subset$ $Y$. The itemsets $X$ and
$\textsc{(}Y\backslash X\textsc{)}$ are, respectively, called
the \textit{premise} \textsc{(}or \textit{antecedent}\textsc{)} and the \textit{conclusion} \textsc{(}or \textit{consequent}\textsc{)} of the
association rule $R$.
\end{definition}
\begin{definition}\textsc{\textbf{\textsc{(}Support, Confidence of an association rule\textsc{)}}}\mbox{}\\Let $R$: $X$ $\Rightarrow$
$\textsc{(}Y\backslash X\textsc{)}$ be an association rule. The support of $R$,
\textit{Supp}\textsc{(}$R$\textsc{)}, is equal to
\textit{Supp}\textsc{(}$Y$\textsc{)}, while its confidence is equal to \textit{Conf}\textsc{(}$R$\textsc{)} = $\displaystyle \frac{\textit{Supp}\textsc{(}Y\textsc{)}}{\textit{Supp}\textsc{(}X\textsc{)}}$.
\end{definition}

Note that the confidence of $R$ is always greater than or equal to
its frequency: \textit{Conf}\textsc{(}$R$\textsc{)} $\geq$
\textit{Freq}\textsc{(}$R$\textsc{)} = $\displaystyle
\frac{\textit{Supp}\textsc{(}R\textsc{)}}{|\mathcal{O}|}$. Indeed,
we have \textit{Supp}\textsc{(}$X$\textsc{)} $\leq$
$|\mathcal{O}|$.

\begin{definition}\textsc{\textbf{\textsc{(}Valid, Exact, Approximate association rule\textsc{)}}}\mbox{}\\An association rule $R$ is said to be
\textit{valid} \textsc{(}or \textit{strong}\textsc{)} if:
\begin{description}
  \item[$\bullet$] its support value \textit{Supp}\textsc{(}$R$\textsc{)} is greater than or equal to the user-specified threshold, \textit{minsupp}, and,
  \item[$\bullet$] its confidence value \textit{Conf}\textsc{(}$R$\textsc{)} is
greater than or equal to a user-specified threshold, denoted \textit{minconf}.
\end{description}
If \textit{Conf}\textsc{(}$R$\textsc{)} = 1, then $R$ is called an \textit{exact} association rule, otherwise it is called an \textit{approximate} association rule.
\end{definition}

Given user-specified minimum support and confidence, the problem of ARM can be split into two steps as follows~\cite{Agra93}:
\begin{itemize}
 \item Extract all frequent itemsets, \textit{i.e.} having the support value greater than or equal to \textit{minsupp}.
\item Generate valid association rules from frequent itemsets. This
generation is limited to rules having the confidence value greater than or
equal to \textit{minconf}.
\end{itemize}

The extraction of the association rules consists in determining the set of valid rules (whose support and confidence are at least equal, respectively, to a minimal threshold support and a minimal threshold of confidence predefined by the user).

The problem of extracting association rules suffers from the high number of generated rules from frequent itemset' set.
The huge number of association rules leads to a derivation to the principal objective, namely the discovery of reliable knowledge, with a manageable size. 

To palliate such a drawback, many techniques derived from FCA, have been proposed. These techniques have aimed to reduce, without information loss, the set of association rules \cite{Bouker2014,Ayouni2011,Gasmi2007}. The main idea is to determine a minimal set of association rules allowing the derivation of redundant association rules. This set is called the \textit{"Generic bases of association rules}".
\subsection{Extraction of Informative Association Rules}
The extraction of association rules is an important technique in data mining. The leading approach of generating association rules is based on the extraction of frequent patterns.

It has been proven that a large number of rules are redundant in the sense that they convey the same information as others \cite{Ashrafi2007,BenYahia2009}
\begin{definition}\textsc{\textbf{\textsc{(}Association rule redundancy\textsc{)}}}\mbox{}\\
Let $\mathcal{AR}$ be the set of valid association rules that can be drawn from a context $\mathds{K}$ for a minimum support threshold \textit{minsupp} and a minimum confidence threshold \textit{minconf}. An association rule $R_{1}$: $X_{1}$ $\Rightarrow$ $Y_{1}$ $\in$ $\mathcal{AR}$ is considered redundant with respect to a rule $R_{2}$: $X_{2}$ $\Rightarrow$ $Y_{2}$ $\in$ $\mathcal{AR}$ if:
\begin{enumerate}
\item \textit{Supp}\textsc{(}$R_{1}$\textsc{)}= \textit{Supp}\textsc{(}$R_{2}$\textsc{)} and \textit{Conf}\textsc{(}$R_{1}$\textsc{)}= \textit{Conf}\textsc{(}$R_{2}$\textsc{)}, and, 
\item $X_{2}$ $\subset$ $X_{1}$ and $Y_{1}$ $\subset$ $Y_{2}$.
\end{enumerate}
\end{definition}
The majority of the generic bases of association rules express implications between generators and closed frequent itemsets. In this thesis, we focus on the $\mathcal{IGB}$ generic base defined in what follows.
\begin{definition}\textsc{\textbf{\textsc{(}Informative generic basis $\mathcal{IGB}$\textsc{)}}}\mbox{}\\
 \label{defIGBbasis}
  Let $\mathcal{FCI}$ be the set of frequent closed itemsets extracted  from a context $\mathds{K}$ and $\mathcal{G}_{f}$ the set of its minimal generators.\\
   $\mathcal{IGB}$ = \{R : g$_s$ $\Rightarrow$ (I-g$_s$) $\mid$ I $\in$ $\mathcal{FCI}$ $\wedge$ I$\neq \emptyset$ $\wedge$ g$_{s}$ $\in$ $\mathcal{G}_{I'}$, I' $\in$ $\mathcal{FCI}$ $\wedge$ I' $\subseteq$ I $\wedge$ confidence(R) $\geq$ minconf $\wedge$ $\nexists$ g$'$ / g' $\subset$ g$_s$ $\wedge$ confidence(g$'$ $\Rightarrow$ I-g$'$)$\geq$ minconf \} \cite{Gasmi2005}.

\end{definition}
\begin{definition}\textsc{\textbf{\textsc{(}Generic basis properties\textsc{)}}}\mbox{}\\
 \label{IGBproprities}
 A generic basis $\mathcal{GB}$, is said to fulfil the ideal properties of an association rule representation if it is \cite{Hamrouni2009} : 
 \begin{enumerate}
 \item \textbf{lossless:} $\mathcal{GB}$must enable the derivation of all valid association rules,
 \item \textbf{sound:} $\mathcal{GB}$ must forbid the derivation of association rules that are not valid, and
 \item \textbf{informative:} $\mathcal{GB}$ must allow to exactly retrieving the \textit{support} and \textit{confidence} values of each derived association rule.
 \end{enumerate}
\end{definition}
Thus, the generic rules of the $\mathcal{IGB}$ generic base represent implications between minimal premises, according to the size or number of items and maximal conclusions.

In fact, in order to reduce the high number and improve the quality of obtained formal concepts or association rules, we opt for using some correlation measures. In the next sub-section, we focus on the definitions of these measures.
\section{Correlation measures}


In the following, we review some of the most frequently used correlation measures that are of use to assess the correlation  of these patterns. The latter notion heavily relies on the notion of support and we characterize its different kinds in the following :

\begin{definition}\label{supp_all}\textsc{\textbf{\textsc{(}Support of an itemset\textsc{)}}}\mbox{}\\Let $\mathds{K}$ = \textsc{(}$\mathcal{O}$, $\mathcal{I}$, $\mathcal{R}$\textsc{)} be a formal context. We distinguish two kinds of support associated with a non-empty itemset $I$:
\begin{description}
 \item[\textit{- Conjunctive support:}] \textit{Supp}\textsc{(}$\wedge I$\textsc{)} = $|\{o\in\mathcal{O}|
\textsc{(}\forall\mbox{ } i \in I,\textsc{(}o,i\textsc{)}\in\mathcal{R}\textsc{)}\}|$.  \textit{Supp}\textsc{(}$\wedge I$\textsc{)},  seen
as a conjunction of items \textsc{(}\textit{i.e.}, $i_1$ $\wedge$ $i_2$ $\wedge$
 \ldots $\wedge$ $i_n$\textsc{)}, is the number of
objects containing all items of $I$.

 \item[\textit{- Disjunctive support:}] \textit{Supp}\textsc{(}$\vee I$\textsc{)} =
$|\{o\in\mathcal{O}|
\textsc{(}\exists\mbox{ } i \in I,\textsc{(}o,i\textsc{)}\in\mathcal{R}\textsc{)}\}|$. \textit{Supp}\textsc{(}$\vee I$\textsc{)}, seen as a disjunction of items \textsc{(}\textit{i.e.}, $i_1$ $\vee$ $i_2$
$\vee$ \ldots $\vee$ $i_n$\textsc{)},  is the number
of transactions containing at least one item of $I$.

\end{description}
\end{definition}

Therefore, when using conjunctive support, the level of correlation gets back to simply computing the fraction of times that the items co-occur. As a metric, the conjunctive  support and its easy calculation and interpretability means that it is the go-to-measure of association in an overwhelming number of applications. In addition, the conjunctive  support fulfills the downward closure property\footnote{A property $ \rho$ is downward-closed if for every set with property $ \rho$, all its subsets also have the property $\rho$.}. However, the conjunctive  support screens out a poor correlation measure, since it  only satisfies  two properties out of the six that any correlation measure has to fulfill~\cite{Duan12}.

In the following, we present the correlation measures that will be of use in the remainder :
\begin{definition}\label{defbond}\textsc{\textbf{\textsc{(}\textit{Bond} correlation measure\textsc{)}}}\mbox{}\\
The \textit{bond} correlation measure~\cite{Omie03} (\emph{aka} \textit{Coherence}~\cite{comine_Lee},
\textit{Tanimoto coefficient}~\cite{Tanimoto1958} and \textit{Jaccard}~\cite{jaccard22}), computes the ratio between the conjunctive support and the disjunctive one.
Thus, the \textit{bond} measure  of a non-empty pattern $I \subseteq \mathcal{I}$ is defined as follows:
\begin{center}
 \begin{equation}\label{bondequation}
\textit{Bond}\textsc{(}\textit{I}\textsc{)} = \frac{\displaystyle \textit{Supp}\textsc{(}\wedge\textit{I}\textsc{)}}{\displaystyle
\textit{Supp}\textsc{(}\vee\textit{I}\textsc{)}}.
\end{equation}
\end{center}
\end{definition}
\begin{example}
With respect to the formal context shown by Table~\ref{table-con}, we obtain the following values for the itemset $\texttt{ab}$:

\begin{itemize}
\item $\textit{Supp}(\wedge \texttt{ab})=1 $;
\item $ \textit{Supp}(\vee \texttt{ab})=8 $; 
\item $\textit{Bond}(\texttt{ab})=\frac{1}{8}$.
\end{itemize}

\end{example}

\begin{definition}\label{defstability}\textsc{\textbf{\textsc{(}Stability measure \textsc{)}}}\mbox{}\\
The intentional stability measure  for a given formal concept highlights the proportion of the subsets of its objects whose closure is equal to the intent of this formal concept. This metric reflects the dependency of the intent on specific objects of the extent~\cite{Kuznetsov2007}. Intentional stability has been shown to be particularly present when investigating taxonomies of epistemic communities, i.e. groups of agents jointly interested in identical topics, sharing the same notions, etc~\cite{stab2207}.
        The intentional stability metric, $\sigma$, of the formal concept $\langle A,B\rangle$ is defined as follows:
         \begin{equation}\label{stability}
         \sigma(\langle A,B \rangle)=\frac{|\{C \subseteq A \mid \psi(C)=B\}|}{2^{|A|}}.
      \end{equation}
       \textsc{Klimushkin} et al. highlighted that a concept that would cover fewer objects was normally less stable than was a concept covering a larger number of objects~\cite{KlimushkinOR10} .
      The stability of the whole set of formal concepts of the coverage $\mathcal{F}_{\mathds{K}}$ is equal to :
         \begin{equation} \label{allstability2}
       \sigma(\mathcal{F}_{\mathds{K}}) = \sum_{i=1}^{n} (\sigma(\langle A_{i},B_{i} \rangle)).
       \end{equation}
\end{definition}
\begin{example}
Let us consider the formal context given by Table \ref{table-con}, from which we  extract the following  formal concept $\langle\{345679\}, \{\texttt{fg}\} \rangle$. The stability value is 0.593.
\end{example}
\section{Conclusion}
\label{firstchapterconclusion}
In this Chapter, we have presented the basic notions used in the remainder of this thesis.

In fact, a bicluster can be considered as a formal concept that reflects the relationship between objects and attributes. To better explain our study, we recall the definition of a bicluster in binary data given by \cite{Prelic2006}: “An inclusion maximal bicluster is the maximal set of objects related to a maximal set of attributes”. This definition perfectly matches with that of a formal concept in the FCA theory. Similarly for association rules, ARM can be used to compose biclusters by finding all association rules that represent biclusters' samples/genes, then extracting the supporting transactions of these items.
 
In the next chapter, we present an overview of biclustering gene expression data and we scrutinize pioneering work that has addressed the extraction of biclusters. Worthy of mention, finding the optimal set of biclusters has been shown to be an NP-hard problem \cite{C.Madeira2004}.

\chapter{Overview of biclustering gene expression data}
\minitoc
\section{Introduction}
DNA microarray technologies help to measure the expression levels of thousands of genes under experimental conditions \cite{C.Madeira2004}. The presence of local patterns in biological data has motivated the wide study to deal with them using pattern-mining-based searches. The use of biclustering in biological data is widespread thanks to its capability to unveil hidden patterns within them. In particular, biclustering is very relevant in the field analysis of gene expression data. In fact, its main thrust stands in its ability to identify groups of genes that behave in the same way under a subset of samples (conditions). However, The pioneering algorithms of the literature has shown some limits in terms of the quality of unveiled biclusters. 

In this chapter, first, we briefly describe how gene expression data is constructed to understand the data used in this thesis. Next, we detail the  biclustering problem, the microarray data used in the experimental phase to evaluate our biclustering algorithms, and the description of the considered tests. In addition, Section ~\ref{relatedworksection}, we scrutinize pioneering works that addressed the extraction of biclusters. Sections~\ref{bicsoftware} and ~\ref{bicvalidation} present some web tools of biclustering algorithms and present some interesting statistical and biological validation. Finally, this chapter is concluded by Section~\ref{secondchapterconclusion}.

\section{Gene expression data construction}

A biological network is a linked collection of biological entities, e.g. genes, proteins, metabolistes, etc. \cite{Henriques2016}. Analyzing information and extracting biologically relevant knowledge, from these entities, is one of the key issues of bioinformatics.

\textit{Gene expression} is the mechanism allowing the production of a protein from a gene. This process happens in two main steps: transcription and translation. While transcription concerns the production of messenger RNA (mRNA) by the enzyme RNA polymerase, and the processing of the resulting mRNA molecule, the translation step pretains to the use of mRNA to direct protein synthesis and the subsequent post-translational processing of the protein molecule.

\begin{figure}[H]
       \begin{center}
          
        \fbox{\includegraphics[width=0.9\textwidth]{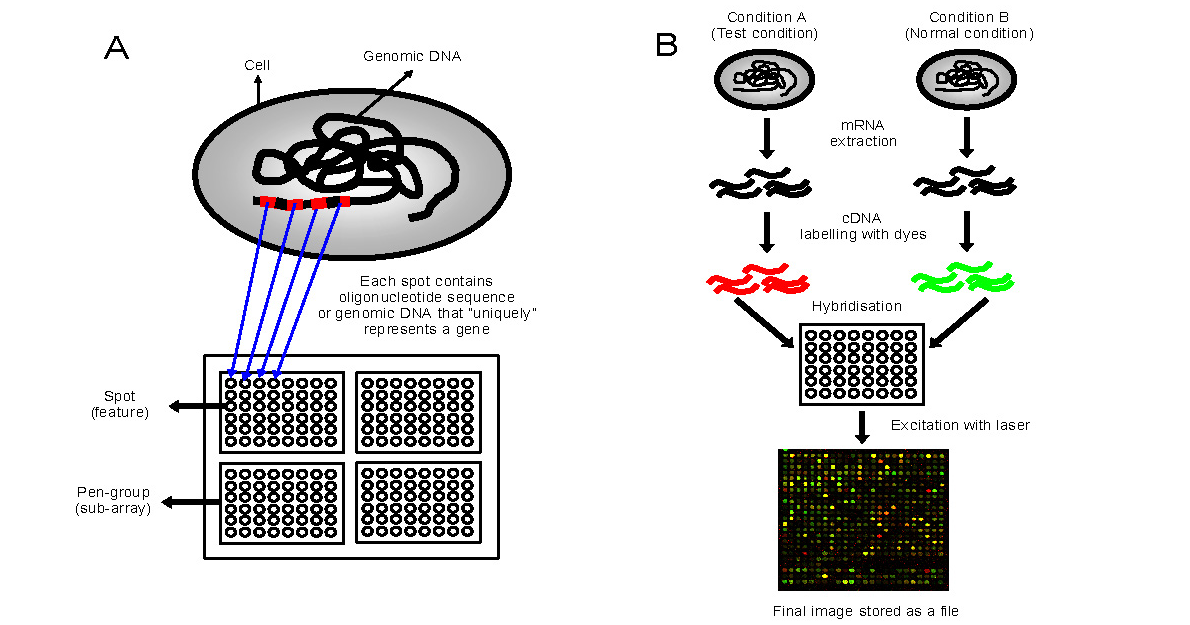}} 
       \end{center}
       \caption{ (A) A microarray may contain thousands of ʻspotsʼ. (B) Schematic of the experimental protocol to study the differential expression of genes. Figure at:http://www.mrc-lmb.cam.ac.uk/genomes/madanm/microarray/}.
       \label{dnasteps}
       \end{figure}

The concentration of mRNA is measured using DNA microarray technologies into numerical values, namely \textit{gene expression data}. These technologies, aka \textit{DNA microarray technologies},  enable the  assessment of the expression levels of thousands of genes under a number of different experimental conditions \cite{C.Madeira2004}. In fact, these technologies have become indispensable tools for a great number of biologists. This is since they are used to monitor genome wide expression levels of genes in a given organism. A microarray is typically a glass slide onto which DNA molecules are fixed in an orderly manner at specific locations called spots (or features). A microarray may contain thousands of spots and each spot may contain a few million copies of identical DNA molecules that uniquely correspond to a gene (Figure \ref{dnasteps}A). The DNA in a spot may either be genomic DNA or a short stretch of oligo-nucleotide strands that correspond to a gene. The spots are printed onto the glass slide by a robot or are synthesised by the process of photolithography. Figure \ref{dnasteps}B gives a general picture of the experimental steps involved. First, RNA is extracted from the cells. Next, RNA molecules in the extract are reversely transcribed into cDNA by using an enzyme reverse transcriptase and nucleotides labelled with different fluorescent dyes.
 


DNA microarrays have been used successfully in various research areas such as gene discovery \cite{Hughes2000}, disease diagnosis \cite{Rosenwald2002} and drug discovery \cite{Gmuender2002}. The functions of the genes and mechanisms underlying diseases can be identified using microarrays.

To do so, gene expression data is arranged in a data matrix (see Table \ref{tableGED}). In the latter, rows represent genes, columns represent samples (experimental conditions), and each cell of the matrix denotes the expression level of a gene under a certain experimental condition. In this respect, the discovery of transcriptional modules of genes that are co-regulated in a set of experiments is of paramount importance \cite{C.Madeira2004}.

\begin{center}
\begin{table}[H]\centering
\begin{tabularx}{\linewidth}{|X|X|X|X|X|X|}
\hline            & \textit{Condition$_{1}$} & $\ldots$   & \textit{Condition$_{j}$}& $\ldots$ & \textit{Condition$_{m}$} \\ 
\hline \textit{Gene$_{1}$} & m$_{11}$    & $\ldots$   & m$_{1j}$    & $\ldots$ &     m$_{1m}$      \\ 
\hline $\vdots$   & $\vdots$    &  $\ldots$  & $\vdots$    & $\ldots$ &          $\vdots$     \\
\hline \textit{Gene$_{i}$} & m$_{i1}$    &  $\ldots$  & m$_{ij}$    & $\ldots$ & m$_{im}$\\
\hline $\vdots$   & $\vdots$    &  $\ldots$  & $\vdots$    & $\ldots$ &  $\vdots$\\
\hline \textit{Gene$_{n}$} & m$_{n1}$    &  $\ldots$  & m$_{nj}$    &  $\ldots$& m$_{nm}$\\
\hline
 \end{tabularx}
\captionof{table}{Gene expression data matrix.}
\label{tableGED}
\end{table}
\end{center}

\section{Biclustering problem}
\label{biclusteringproblem}
The discovery of transcriptional modules of genes that are co-regulated in a set of experiments is of paramount importance \cite{C.Madeira2004}.
 
 Interestingly enough, the clustering technique has been shown to be of benefit in many challenges in bioinformatics. In fact, it allows researchers to gather information such as cancer occurrences, specific tumor subtypes and cancer survival rates \cite{Wei2010}. Although encouraging results have been produced using clustering algorithms. The use of clustering algorithms has two major drawbacks: \begin{enumerate}
 \item They consider the whole set of samples. This is despite the fact that genes may not be relevant to every sample. Instead, they can be relevant to only a subset of samples, which is a fundamental aspect for numerous problems in the biomedicine field \cite{Wang2002}. Thus, clustering should be performed simultaneously on both genes and conditions.
 \item Each gene can only be clustered into one group. Nevertheless, many genes can belong to several clusters depending on their influence in different biological processes \cite{Gasch2002}.
 \end{enumerate}

 In this respect, \textit{biclustering}, which is a particular clustering type, has been palliating these drawbacks. Hence, biclustering aims to identify maximal sub-matrices (\textit{aka biclusters}) where a subset of genes expresses highly correlated behaviors over a range of conditions \cite{C.Madeira2004}. Nevertheless, biclustering task is a highly combinatorial problem and is known to be an NP-Hard one \cite{DBLP:conf/ismb/ChengC00}.

 As it could be witnessed in the dedicated literature, the biclustering usage is widespread in gene expression data analysis. It was first introduced by the pioneering work of \cite{DBLP:conf/ismb/ChengC00}.
 
In the following, we recall some basic definitions borrowed from the biclustering field.
\begin{definition}\textsc{\textbf{\textsc{(}Bicluster\textsc{)}}}\mbox{} \\
 A bicluster is a subset of objects (genes) associated with a subset of attributes (conditions) in which rows are co-expressed.\\
  The bicluster associated with the matrix $M=$($I$,$J$) is a couple ($A$,$B$), such that $A$ $\subseteq$ $I$ and $B$ $\subseteq$ $J$, and ($A$,$B$) is maximal if there does not exist a bicluster ($C$,$D$) with $A$ $\subseteq$ $C$ or $B$ $\subseteq$ $D$. 
 \end{definition} 
 This leads us to the definition of biclustering.
 \begin{definition}\textsc{\textbf{\textsc{(}Biclustering\textsc{)}}}\mbox{}\\ 
The biclustering problem focuses on the identification of the best biclusters of a given dataset. The best bicluster must fulfill a number of specific homogeneity and significance criteria (guaranteed through the use of a function to guide the search) \cite{Orzechowski2013}. 

 \end{definition}

\begin{sidewaysfigure}[p] 
 \fbox{
            \scalebox{0.95}{
\begin{tikzpicture} [grow=right,
  every node/.style = {shape=rectangle, rounded corners,
    draw, align=center,
    top color=white, bottom color=blue!20}]] 
\tikzstyle{level 1}=[level distance=60mm,sibling distance=40mm] 
\tikzstyle{level 2}=[level distance=90mm,sibling distance=15mm] 
 \node{Biclustering algorithms} 
   child{node{Pattern-mining-based algorithms}
       child{node{FCA-based approaches}} 
       child{node{ARM-based approaches}} 
       child{node{SPM-based approaches
       }}
   }
    child{node{ Stochastic search-based algorithms} 
        child{node{Hybrid-based approaches} }
        child{node{Evolutionary computation-based approaches}}
        child{node{Neighborhood search-based approaches}}
     } 
  child{node{Systematic search-based algorithms} 
  child{node{ Biclusters enumeration-based approaches }}
      child{node{ Greedy iterative search-based approaches }}
      child{node{ Divide-and-conquer-based approaches }}
      }
 
  ; 
 \end{tikzpicture} }}
 
\caption{Structured view on existing biclustering algorithms.}
\label{view_algorithm}
\end{sidewaysfigure}
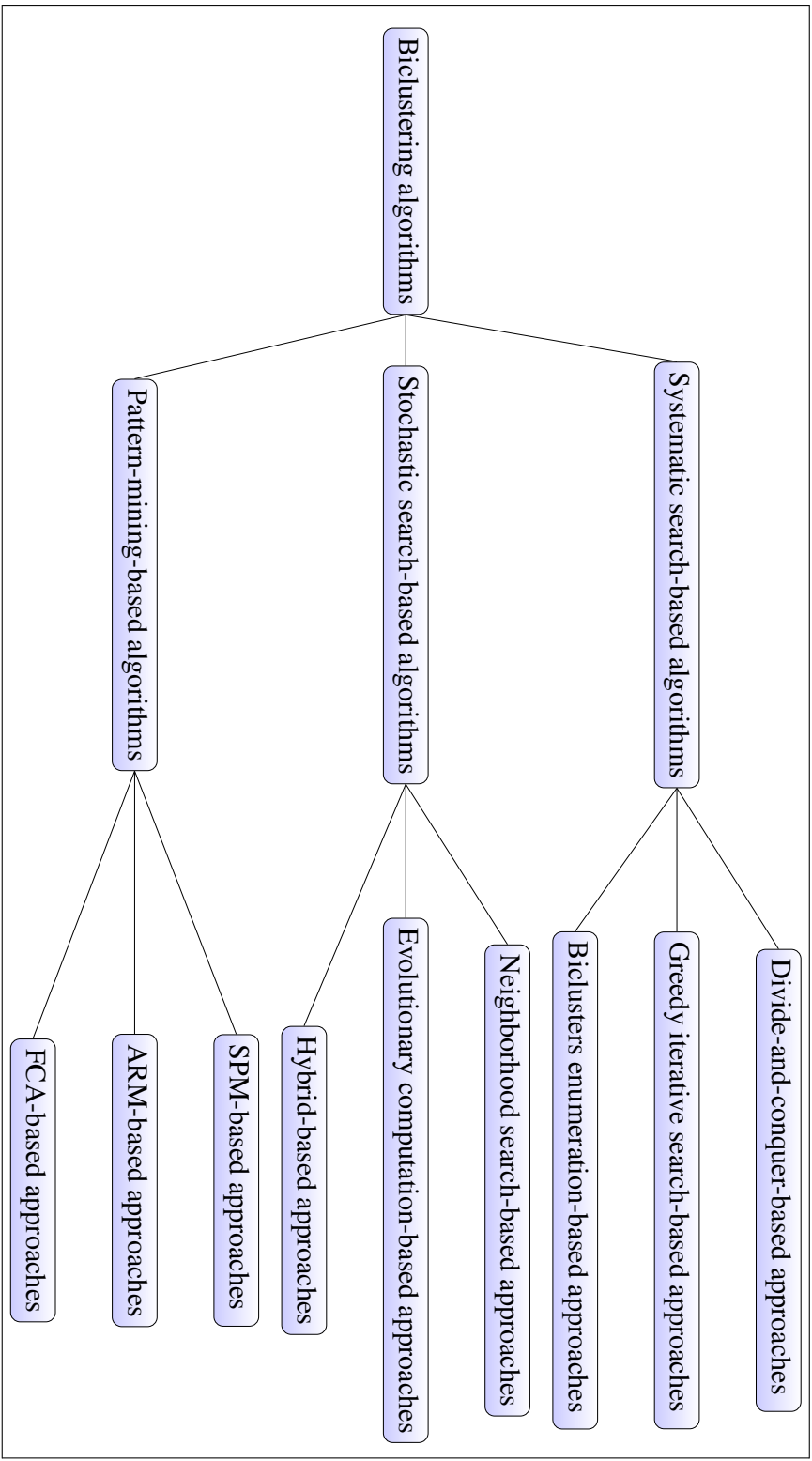 

\section{Biclustering gene expression data: Literature review}
In this section, we focus on presenting an overview of the literature approaches, which are related to our topic of mining biclusters from gene expression data.
\subsection{Structured view on biclustering algorithms}
\label{relatedworksection}
The costly computation complexity of extracting maximal sub-matrices of genes and conditions such that the genes express highly correlated behaviors over a range of conditions has been a main impediment to the wide-scale use of gene expression analysis community. A review of various biclustering algorithms for gene expression data was provided in \cite{Freitas20013}, where existing biclustering algorithms were grouped into two main streams to which the third stream would be added. At a glance, as depicted by Figure \ref{view_algorithm}, the dedicated literature has witnessed three main streams for addressing the biclustering task. These streams are detailed in the following.

\subsubsection{Systematic search-based biclustering}

 The systematic search-based stream includes the following approaches:
 
 \begin{enumerate}
  \item  Divide-and-conquer-based approach: Generally, this approach repeatedly splits the problem into smaller ones with similar structures to the original problem, until these sub-problems become smaller enough to be straightforwardly solved. The solutions to the sub-problems are then combined to create a solution to the original problem \cite{Freitas20013}. \\Algorithms adopting this approach were given in \cite{Prelic2006} and \cite{Teng2008}.
 
  \item  Greedy-iterative-search-based approach: In this approach, a solution is constructed in a step-by-step way using a given quality criterion. Decisions made at each step are based on information at hand without worrying about the impact of these decisions in the future. Moreover, once a decision is made, it becomes irreversible and is never reconsidered \cite{Freitas20013}. \\Algorithms adopting this approach were given in \cite{DBLP:journals/jcb/Ben-DorCKY03,Cheng2008}  and \cite{Zhang2005}.
  \item Bicluster-enumeration-based approach: As indicated by its name, an enumeration algorithm enumerates all the solutions for the original problem. The enumeration process is generally represented by a search tree \cite{Freitas20013}. \\Algorithms adopting this approach were given in \cite{Ayadi2009,Ayadi2012a,Ihmels2004} and \cite{Tanay2002}.                                                                                                                                                                                                                                   \end{enumerate}
  
  \subsubsection{Stochastic search-based biclustering}
 The stochastic search-based stream includes the following approaches: 
  \begin{enumerate}
  \item Neighborhood-search-based approach: It starts with an initial solution and then moves iteratively to a neighboring solution thanks to the neighborhood exploitation strategy.\\
  Algorithms adopting this approach were given in \cite{Ayadi2010} and \cite{Das2010}.
  \item Evolutionary-computation-based approach: This approach is based on the natural evolutionary process such as population, reproduction, mutation, recombination, and selection.\\
  Algorithms adopting this approach were given in \cite{Divina2007} and \cite{Divina2006}.
  \item Hybrid-based approach: The latter tries to combine the neighborhood search and evolutionary approaches.\\ 
  Algorithms adopting this approach were given in \cite{Gallo2009} and \cite{Mitra2006}.                                                                           
  \end{enumerate}  
  
  \subsubsection{Pattern-mining-based biclustering} 
   
The Pattern-mining-based stream includes:
  \begin{enumerate}
    \item Sequential-Pattern-Mining (SPM)-based approaches: SPM is used in order to extract order-preserving biclusters. A bicluster is order-preserving if there is a permutation of its columns under which the sequence of values in every row increases. In this context, SPM is applied, and the biclusters are extracted from the frequent sequences as well as their supporting transactions. \\Algorithms adopting this approach were given in \cite{Henriques2013} and \cite{Henriques2014a}.
    \item Association Rules Mining (ARM)-based approaches: ARM can be used to compose biclusters. To perform this task they divide the problem into two sub-problems: 
    \begin{enumerate}
    \item Finding all association rules that represent biclusters' samples/genes. In fact, they consider items of both the premise and conclusion of an association rule.
    \item Extracting the supporting transactions of these items. 
    \end{enumerate} 
    The authors in \cite{Mondal2014} provide a review of various biological applications of association rules mining.
    \item Formal Concept Analysis (FCA)-based approaches: FCA can be viewed as a kind of biclustering for binary data. It provides pattern (\textit{bicluster}) extraction from a binary relation, namely a \textit{formal concept}. In its gene expression data applications, the concept's extent represents maximal sets of genes related to a maximal set of samples (concept's intent). \\Algorithms adopting this approach were given in \cite{Kaytoue2014} and \cite{Kaytoue2011a}. 
    \end{enumerate}

   To preserve conciseness, we do not further expand this overview and redirect the reader to the already extensive number of recent surveys on biclustering \cite{C.Madeira2004,Charrad2011,Eren2013,Freitas20013,Sim2013,Padilha2017} for more details. In this thesis, we are particularly interested in the pattern-based biclustering algorithms especially ARM and FCA-based approaches.

\subsection{On the relevance of pattern-mining-based biclustering}
\label{relevance}   
 Patten-mining searches and its integration with biclustering, referred to as pattern-mining-based biclustering, defines a new promising direction. Section \ref{relevance} covers the benefits of these approaches.
\subsubsection{Potentialities of pattern-based biclustering}

Contributions of pattern-mining-based approaches for biclustering include:
\begin{itemize}
\item Efficient exhaustive searches. Pattern mining algorithms allow for the efficient analysis of large matrices (over 10.000 $\times $ 400 elements). Additional pattern mining principles can be used to foster scalability, including searches in distributed/partitioned data settings or targeting approximate patterns \cite{Han2007,DBLP:journals/bmcbi/GuptaRK11};
\item Biclusters from patterns with parametrizable coherency strength (multiple ranges of values) \cite{Okada2007,Pandey2009}, contrasting with peer approaches to find biclusters with differential values or fixed coherency strength \cite{Tanay2002};
\item Flexible structures of biclusters: Arbitrary positioning of biclusters without the need to fix the number of biclusters apriori \cite{Okada2007,Serin2011};
\item Inherent orientation to learn constant values on columns, already an improvement over approaches requiring constant values on both columns and rows \cite{Henriques2015a,Henriques2014,Henriques2014a};
\item Easy extension for labelled data using discriminative pattern mining or classification rules \cite{Fang2010,Odibat2014}.
\end{itemize}

Biclustering has been applied for several domains. In this thesis, we focus on the application of biclustering on biological domains, where the discovery of biclusters is applied over expression data to identify co-regulated genes \cite{C.Madeira2004,Freitas20013,Eren2013,Henriques2014,Henriques2015a}. 

\begin{figure}
      \begin{center}
         
       \fbox{\includegraphics[width=0.9\textwidth]{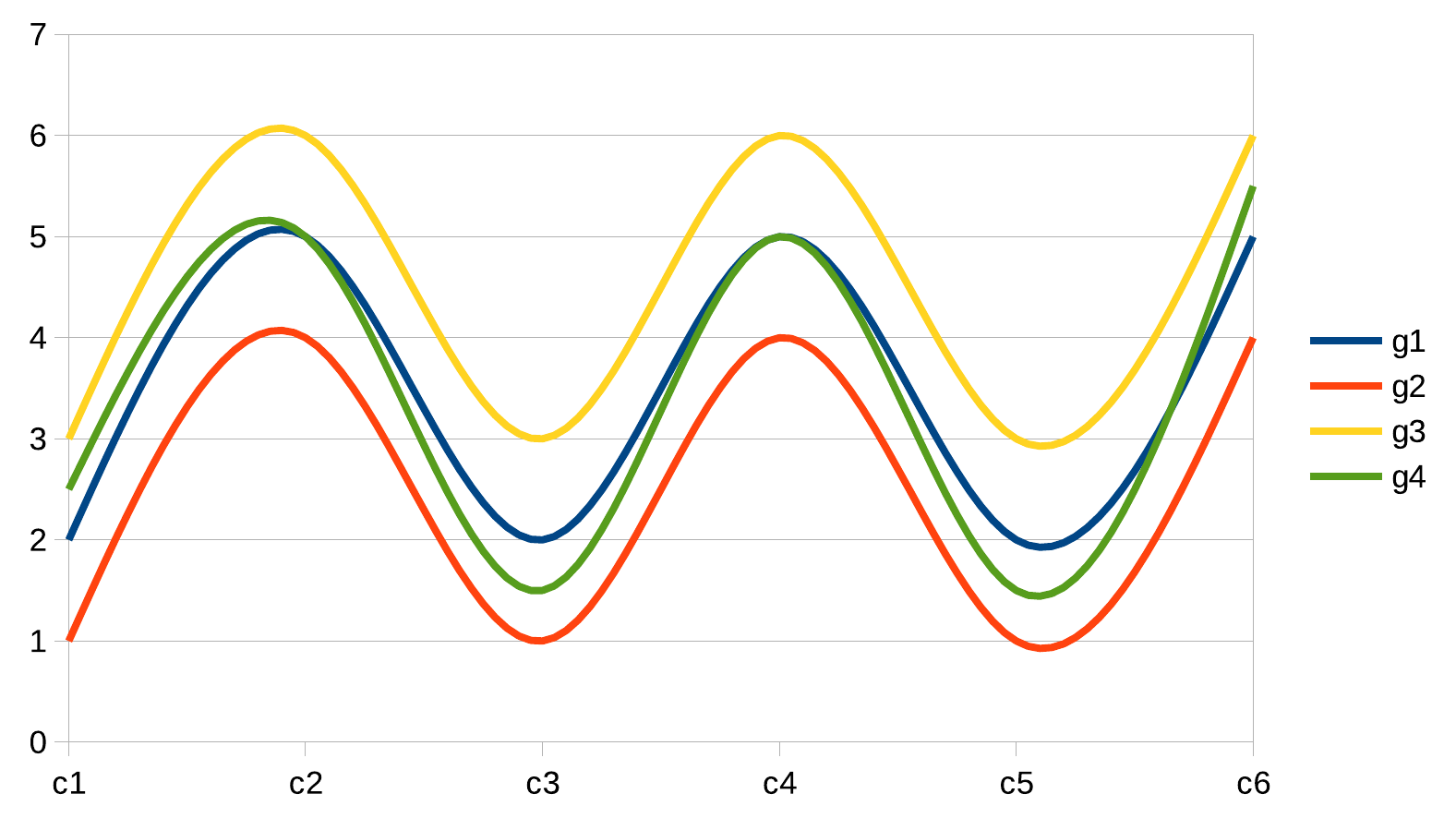}} 
      \end{center}
      \caption{Examples of positive correlations.}
      \label{positivecorrelations}
      \end{figure}

\begin{figure}
      \begin{center}
         
       \fbox{\includegraphics[width=0.9\textwidth]{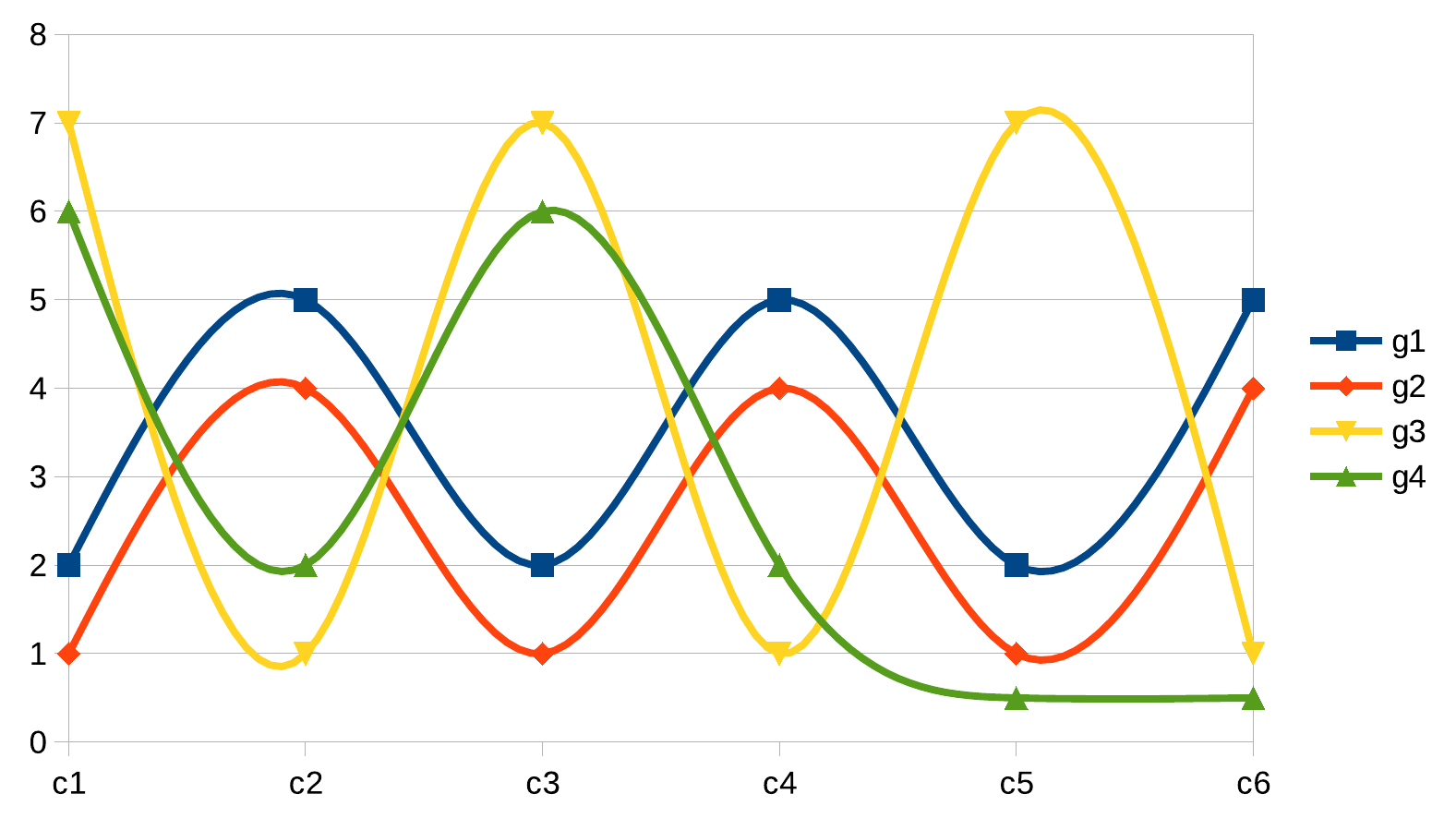}} 
      \end{center}
      \caption{Examples of negative correlations.}
      \label{negativecorrelations}
      \end{figure}

In this context, most of the existing biclustering algorithms only identify positive correlation genes. Recently, biological studies have turned to a trend focusing on the notion of negative correlations. These biological interests give reason to develop and investigate the problem of discovering negative correlations of statistical and biological significance from gene expression data. As a result, biclusters can be of positive or negative correlations. Figures \ref{positivecorrelations} and \ref{negativecorrelations} depict an example of these correlations. 
We present in the following some work that extracts positively (negatively)-correlated biclusters from gene expression data.

We present, in what follows, the state of the art approaches dealing with biclustering gene expression data using FCA and ARM. We precisely start with extracting biclusters of positive correlations. 
\subsection{Extracting biclusters of positive correlations}
The guiding idea is to extract positively-correlated biclusters, since according to \cite{DBLP:journals/bioinformatics/LuanL03} and \cite{SD2003}, in DNA microarray data analysis, we add genes into a bicluster whenever their trajectory patterns of expression levels are similar across a set of samples. In this respect, we can cite the following approaches:
\subsubsection{ARM-based approaches}
ARM can alternatively be used to compose biclusters \cite{Henriques2015a}. Its core task is the support- and confidence-guided discovery of association rules between itemsets \cite{Han2007}. In this context, when using association rules to compose biclusters, the items of the antecedent and consequent of a rule, as well as the supporting transactions from both sides, are considered to extract biclusters.

Many approaches have paid attention to the extraction of biclusters using ARM. One of the algorithms belonging to this context is given by Carmona et al. \cite{Carmona-Saez2006}. The latter extended simple rules by integrating annotations from semantic sources, knowledge bases
and bibliographic databases. Annotations are labels associated with groups of rows or groups of columns.

In line with this work, GenMiner \cite{Martinez2007,Martinez2008} integrated the input data with annotations. In particular, it focused primarily on rules over expression data of the type\textit{ annotations $\Longrightarrow$ expression profiles}. profiles. Authors also extended the ARM towards two additional types of associations: 1) \textit{expression profiles $\Longrightarrow$ gene annotations}, meaning that a group of genes with an expression pattern across a set of conditions is likely to have a set of corresponding annotations; and 2) relations among gene annotations. Illustrative rules include: \textit{annotation$_{1}$ $\Longrightarrow$\{ c$_{1}$$\downarrow$, c$_{2}$$\uparrow$\}}, meaning that a group of genes (with the same annotation) is likely to be under-expressed in condition c$_{1}$ and over-expressed in condition c$_{2}$. An alternative rule is \textit{\{c$_{1}$$\downarrow$, c$_{2}$$\uparrow$\} $\Longrightarrow$ annotation$_{1}$}; i.e., a group of genes with the expression profile given by $c_{1}$and $c_{2}$ is likely to have specific annotations. However, in the absence of background knowledge, annotations can be retrieved from the input matrix based on clusters of rows and columns.

In \cite{Mondal2012}, the authors proposed a new approach, called FIST, for extracting bases of extended association rules and conceptual biclusters, using frequent closed itemsets \cite{Pasquier1999a}. Nevertheless, they failed to detail their discretization method and treated the matrix as though it was already binary. This was done despite the fact that microarray data is not initially coded in a binary format. Furthermore, their approach did not entail any biological validation of the extracted biclusters.

A lot of other work has also emerged. We cite, for example, the Debi algorithm \cite{Serin2011} which which is based on $0/1$ discretization. It generates association rules and uses also maximal itemsets for biclustering, which are associated with biclusters having a maximized size of columns. Such flattened biclusters are only of interest when there is an extension step to be performed to include new rows. However, since both vertical and smaller biclusters are lost, this representation leads to incomplete solutions as they are just a subset of all biclusters.

In \cite{Boutsinas2013}, the authors proposed an association rules-based biclustering algorithm that used the Apriori algorithm. However, it is still insufficient, due to high combinatorial possibilities associated with the exhaustive discovery of association rules (redundant rules).
\subsubsection{FCA-based approaches}
FCA can be applied to perform biclustering. A bicluster is a specific formal concept (Definition \ref{formal concept}) called a bi-set. A bi-set must satisfy a local constraint: The column set (or intent) is the maximal set of columns that are true for the supporting set of rows (or extent) \cite{Medina2008}. Bi-sets may, additionally, satisfy user-defined constraints.

 In this respect, many approaches have been devoted to the extraction of biclusters using FCA. The approach suggested in \cite{Pensa2004} relied on a single threshold, where expression values greater than this threshold were represented by 1, otherwise by 0. Most discretization techniques commonly applied to gene expression data used absolute expression values. However, the main drawback of this technique was how to find the best method to set the threshold value. 
 
 The approach proposed in \cite{Besson2005} stood also with this principle. This approach allowed mining concepts in gene expression data under monotonic constraints. Nevertheless, the D-Miner algorithm did not work on the biological validation of extracted concepts.

 In this same context, the authors in \cite{Kaytoue2011a} and \cite{Kaytoue2011b} used the inter-ordinal scaling of numerical data and interval pattern structures combined with FCA techniques, where they considered that formal concepts were the groups of genes whose expression values were in the same intervals for a subset of conditions.
 
 In the same vein, we mention the Trimax algorithm \cite{Kaytoue2014}. In fact, this latter was devoted to the extraction of biclusters of similar values. In this approach, Kaytoue el al. \cite{Kaytoue2014} referred to the algorithm presented in \cite{Kaytoue2011a}, using the Triadic Concept Analysis \cite{Lehmann1995,Trabelsi2012} in order to extract biclusters with similar values. Both of the latter approaches only paid attention to the extraction of one type of biclusters, i.e. biclusters with similar values (similar patterns). In addition, they did not offer any biological validation for the obtained biclusters.
   
   The above mentioned biclustering algorithms have had the tendency to either focus on one type of biclusters, extract overlapping ones or refrain from biological validation. Thus, in this thesis, we introduce new FCA-based approaches for the extraction of biclusters from gene expression data.

\bigskip    
Most of the existing biclustering algorithms identify only positive correlation genes. Yet, recent biological studies have turned to a trend focusing on the notion of negative correlations. In the following, we detail this aspect.
\subsection{Extracting biclusters of negative correlations}
These approaches have tackled the extraction process in a different manner. It is of paramount importance to extract negatively-correlated biclusters since most of the existing biclustering algorithms identify only positively-correlated genes despite the fact that recent biological studies have focused on the notion of negative correlations. These biological interests have given reason to develop and investigate the problem of discovering negative correlations of statistical and biological significance from gene expression data. The authors in \cite{Zhao2008} studied in depth the negatively-correlated pattern. Actually, the expression values of some genes tend to be the complete opposite of the other genes. In a straightforward case, given two genes G1 and G2, under the same condition C, if both G1 and G2 are affected by C, while G1 goes up and G2 goes down, we say that they have a negative correlation pattern. For example, the genes YLR367W and YKR057W of the Yeast microarray dataset \cite{Tavazoieand1999} have a similar disposition, but have a negative correlation pattern with the gene YML009C under 8 conditions \cite{Zhao2008}. Some suggest that these genes are a part of the protein translation and translocation processes and therefore should be grouped into the same cluster. Later on, several other algorithms were proposed \cite{Nepomuceno2015,Odibat2014,roy2013cobi,Zeng2010}.

In this section, we present the problem of extracting biclusters of negative correlations. In this
respect, we can cite the following approaches:
\subsubsection{General approaches}
Currently several algorithms have been proposed to identify negative correlation from microarray data.

The authors in \cite{Zeng2010,Madeira2009,Madeira2010} put forward AIE \cite{Zeng2010}, CCC \cite{Madeira2009}, and e-CCC \cite{Madeira2010}. These latter were based on efficient string processing techniques, such as suffix trees, which could unveil negative correlations. e-CCC could extract more negative correlation than CCC because it can tolerate a specified number of errors \cite{Madeira2010}.

The approach proposed in \cite{Li2009} stood also within this principle. This approach permitted extracting both positive and negative correlations. However, it could not discover all maximal negative correlations due to the use of parallel technologies.

In \cite{Nepomuceno2015a}, the authors introduced the idea of integrating biological information. Basically, some a priori biological information was introduced as an input and the search process had a bias to find better biclusters. This algorithm was based on the algorithm presented in \cite{Nepomuceno2015}. Although several procedures differed the most relevant contribution was the fitness function definition to integrate biological information. This function consisted of three parts: The first one was a term to control the size of biclusters. The second one was the correlation among genes to capture co-expressed genes. The third term was an additional term to integrate biological information.

Much other work has also emerged. We cite, for example, the algorithms proposed in \cite{Odibat2014} where authors presented a novel algorithm for discovering arbitrarily positioned co-clusters. They extended this algorithm to discover discriminative co-clusters by integrating the class label in the co-clustering discovery process. Both proposed algorithms were robust against noise, allow overlapping and capturing positive and negative correlations in the same co-cluster. 

Ayadi et al. \cite{Ayadi2014} suggested a memetic algorithm, called MBA, for discovering negative correlated genes of microarrays data. MBA operated on a set of candidate biclusters and used these biclusters to create new solutions by applying variation operators such as combinations and local improvements. 
\subsubsection{Pattern-based approaches}
These approaches have tackled the extraction process using pattern-mining searches.  In this respect, we can cite the following approaches:

The algorithm put forward in \cite{Li2010} was utilized for mining negative correlations. The proposed method transformed the data matrix to bipartite graph database bGD (mine frequent R-biclique subgraphs of bGD), then transformed bGD into transaction database TD (mine frequent itemsets) to extract negative correlation patterns including those with continuous time points and discretion time points. However, some results obtained from \cite{Li2010} can only contain positive correlations.

The approach suggested in \cite{Henriques2014} stood with this principle. The proposed BicPAM biclustering approach integrated existing state-of-the-art pattern-based approaches and put forward a new method which allowed mining non-constant types of biclusters, including additive and multiplicative coherencies in the presence or absence of symmetries; i.e., it extracted positive correlations and probably negative correlations (symmetries).

The approach proposed in \cite{Tu2016} also belonged to those dealing with FCA. In fact, it was based on the principle of finding size-balanced negative correlation expression patterns over a subset of time points or experimental conditions with the formal conecpt analysis technique. The main drawbacks in this approach were (1) the use of the concept of a latice, which is more costly in memory size. (2) In the discretization phase , it used a parameter to tune the desired derivation from the mean and standard derivation of rows. However, this parameter was set fixed to 0.5 (the main drawback is how to find the best threshold value). (3) It fixed a minimum number of genes in each of two subsets from every negative correlation expression pattern.

 Interestingly enough, to the best of our knowledge, there is no previous work that has dealt with both biclustering and $\mathcal{IGB}$ representation (Definition \ref{defIGBbasis}) to extract biclusters of negative correlations from gene expression data.

\section{Microarray datasets }
We present in this section the microarray datasets used by the biclustering community in order to evaluate biclustering algorithms. These datasets are grouped in Table \ref{tabledataset}.

Note that the most used microarray data are:  Yeast Cell Cycle and Saccharomyces Cerevisiae \cite{Ayadi2009,ayadi201414,Prelic2006,Prelic2006a}. Indeed, these microaarays are the easiest to validate and interpret statistically
and biologically.
\section{Biclustering software}
\label{bicsoftware}
We present in this section some web tools of biclustering algorithms. These web tools are sketched in Table \ref{biclusteringtools}.

For our comparative study, we opt for BicAT. This latter is the most-used tool in the biclustering community. This is because it contains the most known algorithms in the biclustering domain.

\begin{center}
\begin{table}[]\centering
\begin{tabularx}{\linewidth}{|X|l|l|X|}
\hline \textbf{Microarray data} & \textbf{Number of genes} & \textbf{Number of conditions}  & \textbf{Website}\\ 
\hline  Arabidopsis Thaliana &  334 &  69 & \url{ http://www.tik.ethz.ch/
sop/bimax/}\\
\hline  Alzheimer & 1663  &  33 & \textendash \\
\hline  Colon Rectal Cancer & 2000  & 62  & \url{http ://microarray.prin ceton.edu /oncology/affydata/index.html}\\
\hline  Human B-cell Lymphoma & 4026  & 96   & \url{http ://arep.med.harvard.edu/
biclustering/}\\
\hline  Leukemia&  7129 & 72  & \url{http ://sdmc.lit.org.sg/GEDatasets/Datasets.html}\\
\hline  Lung Cancer & 12533  &  181 & \url{http ://sdmc.lit.org.sg/
GEDatasets/Datasets.html}\\
\hline  Prostate Cancer&  12600 &  136  & \url{http ://sdmc.lit.org.sg/GEDatasets/Datasets.html}\\
\hline  Saccharomyces Cerevisiae& 2993  & 173  & \url{http ://www.tik.ethz.ch/
sop/bimax/}\\
\hline  Yeast Cell Cycle& 2884  & 17  & \url{http ://arep.med.harvard.edu/
biclustering/}\\ 
\hline
 \end{tabularx}
\captionof{table}{Microarray data sets used for evaluation of biclustering algorithms.}
\label{tabledataset}
\end{table}
\end{center}

\begin{center}
\begin{table}[]\centering
\scalebox{1.0}{
\begin{tabularx}{\linewidth}{|l|X|X|}
\hline \textbf{Web tool} & \textbf{Algorithms referred} & \textbf{Observations} \\ 
\hline  BiCAT & CC, ISA, xMotifs, OPSM, BiMax &  \url{http://www.tik.ethz.
ch/sop/bicat/ } \\
\hline  BiCAT-plus & Extension of BiCAT  & \url{http://home.kspace.org/
FADL/Downloads/PhD/Bicat-Plus
paper/}    \\
\hline  BiGGeSTS & CCCBin and eCCCBi[45]  & Expression data with continuous / temporal conditions. \url      {http://kdbio.inesc-id.pt/
software/biggests/}  \\
\hline  BiBench & CC, Plaid, OPSM, ISA, Spectral, xMOTIFs, Bayesian Biclustering, COALESCE, CPB, QUBIC, FABIA  &  \url{http://bmi.
osu.edu/hpc/software/bibench/} \\
\hline  MTBA & CC, Bipartite Spectral Graph Partitioning, OPSM, ISA, Spectral Biclustering, Information Theoretic Learning (ITL), xMOTIF, Plaid, FLOC, BiMax, Bayesian Biclustering, LAS, Qubic, Fabia   &  \url{http://iitk.ac.in/iil/mtba/} \\
\hline  biclust (R) &  CC, Spectral, Plaid Model, xMotifs, Bimax &  Package R \\
\hline  BicPAMS & BicPAM, BicNET, BicSPAM, BiC2PAM, BiP, DeBi and BiModule &  \url{https://web.ist.utl.pt/rmch/bicpams} \\
\hline
 \end{tabularx} }
\captionof{table}{Web tools for use of biclustering algorithms.}
\label{biclusteringtools}
\end{table}
\end{center}

\section{Biclusters validation}
\label{bicvalidation}
The application of biclustering algorithms on the microarray datasets generates big groups or partitions containing tens or hundreds of genes. Thus, it is essential to use sophisticated tools to validate them.

The bicluster validation can be statistical based on the properties of obtained biclusters or biological based on the genes annotation of different biclusters.

In the following, we respectively describe statistical and biological validation.
\subsection{Statistical validation}
\label{Statistical_validation_description}
The statistical validation presents a key step for the biological validation and interpretation of obtained biclusters. To evaluate the statistical relevance of our algorithms, we heavily rely on the following criteria.
\begin{itemize}
\item \textbf{Coverage} \cite{Bleuler2004,Liu2009,Mitra2006}:  It represents the total number of cells in a microarray data matrix covered by the obtained biclusters. In the biclustering domain, validation using coverage is considered interesting since large coverage of a dataset is very important in several applications that rely on biclusters \cite{Freitas20013}. In fact, the higher the number of highlighted correlations, the greater the amount of extracted information. Consequently, the higher the coverage, the lower the overlapping in biclusters.

 \item \textit{\textbf{p-value}}:  We compute the percentage of biclusters having an adjusted \textit{p-value}, \textit{i.e.} the proportion between the number of biclusters having an adjusted \textit{p-value} and the total number of obtained bicluters. We compute the adjusted \textit{p-value} \cite{Prelic2006}, \textit{i.e.} based on the exact value of Fisher test \cite{Fisher1922}, to measure the quality of the obtained biclusters. In fact, the biclusters having a p-value lower than 5\% are considered as over-represented; in other words, the majority of genes of a bicluster have common biological characteristics.
  The best biclusters have an adjusted p-value less than 0.001.
  This measure is computed thanks to the web tool \textit{\textbf{FuncAssociate}}\footnote{Available at  \url{http://llama.mshri.on.ca/funcassociate/}}\cite{Berriz2003}.

\end{itemize}

\subsection{Biological validation}
The Gene Ontology (GO) project\footnote{\url{http://geneontology.org/}} is a collaborative effort to address the need for consistent descriptions of gene products in different databases. The project began as a collaboration between three model organism databases, among them the Saccharomyces Genome Database (SGD). This latter concerns our datasets (Yeast cell cycle and Saccharomyces Cerevisiae).
The GO project provides controlled vocabularies of defined terms representing gene product properties. This covers three domains: \textit{(i) biological process}, \textit{(ii) molecular function} and \textit{(iii) cellular component}.

  Biological validation of biclusters of microarray data is one of the most important open issues. So far, there have been no general guidelines in the literature on how to biologically validate such biclusters. We briefly present some biclustering tools that are publicly available for microarray data analysis.
  \begin{enumerate}
  \item GOTermFinder (\url{http://db.yeastgenome.org/cgi-bin/GO/g TermFinder}) searches for significant shared GO terms, or parents of GO terms, used to annotate gene products in a given list.
  \item FuncAssociate (\url{http://llama.med.harvard.edu/cgi/func/funcassociate}) is a web-based tool that accepts as input a list of genes and returns a list of GO attributes that are over/under-represented among the genes of the input list. Only those over/under-represented genes are reported.
  \item GeneBrowser (\url{http://bioinformatics.ua.pt/genebrowser2/}) is a web tool that combines, for a given list of genes, data from several public databases with visualisation and analysis methods to help identify the most relevant and common biological characteristics. The provided functionalities include the following: a central point with the most relevant biological information for each inserted gene; a list of the most related papers in PubMed\footnote{\url{http://www.ncbi.nlm.nih.gov/pubmed}} and gene expression studies in ArrayExpress; and an extended approach to functional analysis applied to GO, homologies, gene chromosomal localisation and pathways.
  \item GENECODIS (\url{http://genecodis.dacya.ucm.es/}) is a web tool for the functional analysis of a list of genes. It integrates different sources of information to search for annotations that frequently co-occur in a list of genes and ranks them according to their statistical significance.
 
  \end{enumerate}
In order to biologically evaluate our biclusters, we make use of the \textbf{\textit{GoTermFinder}} web tool\footnote{Available at \url{http://db.yeastgenome.org/cgi-bin/GO/goTermFinder }} \cite{Boyle2004}. It searches for significant shared GO terms, used to describe the genes in a given list to help discover what the genes may have in common. In fact, the biological criterion permits measuring the quality of the resulting biclusters, by checking whether the genes of a bicluster have common biological characteristics.

\section{Conclusion}
\label{secondchapterconclusion}
In this chapter, we have presented some of the pioneering work that has tackled the issue of extracting biclusters from gene expression data. Scrutiny of the above mentioned work highlights the fact that the above mentioned approaches have the tendency to either focus on one type of biclusters, extract overlapping ones, refrain from biological validation or need a background knowledge. In addition, most approaches based on pattern-mining searches focus only on extracting biclusters of positive correlations. Consequently, in the following chapter, we introduce new pattern-mining-based approaches for the extraction of positively-correlated biclusters.

\part{{\Huge Contributions}}
\chapter{Identifying biclusters of positive correlations}
\label{chapter3}
 \minitoc
\section{Introduction}
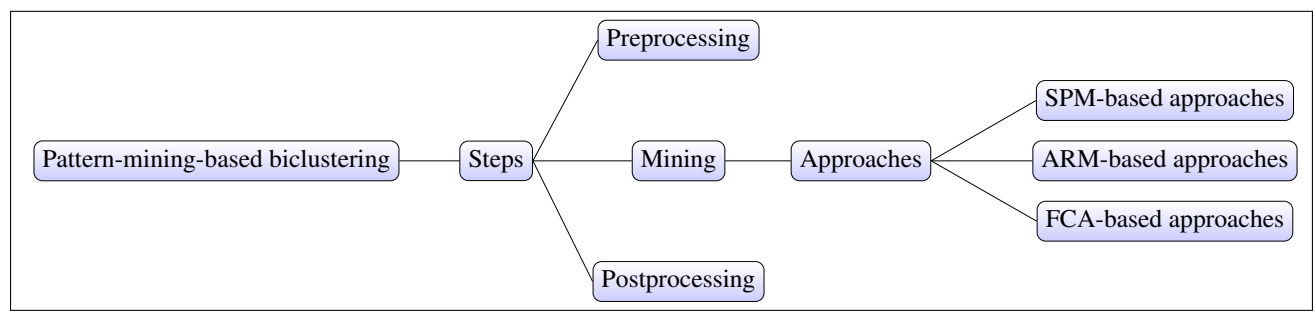
\begin{figure}[H] 
        \fbox{
                   \scalebox{0.8}{
       \begin{tikzpicture} [grow=right,
         every node/.style = {shape=rectangle, rounded corners,
           draw, align=center,
           top color=white, bottom color=blue!20}]] 
       \tikzstyle{level 1}=[level distance=46mm,sibling distance=40mm] 
       \tikzstyle{level 2}=[level distance=30mm,sibling distance=20mm]
       \tikzstyle{level 3}=[level distance=30mm,sibling distance=10mm] 
       \tikzstyle{level 4}=[level distance=50mm,sibling distance=10mm]
        
        \node{Pattern-mining-based biclustering} 
           child{node{ Steps}
               child{node{Postprocessing}}
               child{node{Mining}
               child{node{Approaches}
                child{node{FCA-based approaches}}
                child{node{ARM-based approaches}}
                child{node{SPM-based approaches}}
               }}
               child{node{Preprocessing}}
            } 
        
         ; 
        \end{tikzpicture} }}
        
       \caption{Pattern-mining-based biclustering: Steps and mining options.}
       \label{view_pattern_options}
       \end{figure}

Biclustering has been demonstrated to very relevant within the field analysis of gene expression data. In fact, its main thrust  stands in its ability to identify groups of genes that behave in the same way under a subset of samples (conditions). However, The pioneering algorithms of the literature have shown some limits in terms of quality of unveiled biclusters. Thus, we introduce new algorithms for biclustering microarray data. While traditional biclustering methods rely on flexible merit functions to guide the space exploration, pattern-based approaches require these functions to be defined in terms of support and, eventually, confidence or other interestingness metrics. The application of this function enables an efficient space search that produces an arbitrarily high number of coherent biclusters. Figure \ref{view_pattern_options} covers principles according to three major decision dimensions (preprocessing, mining, postprocessing).
       
In this chapter, we focus on the extraction of positively-correlated biclusters. We propose novel approaches to more improve the biclustering task . This is achieved through the use of Association Rules Mining (ARM) and  Formal Concept Analysis (FCA).     

The rest of this chapter is organized as follows: Section \ref{chapterBiARM} deals with the presentation of our proposed algorithm BiARM. Our algorithm, called BiFCA+, is explained in detail in Section \ref{chapterBiFCA+}. In Section \ref{chapterBiFCA} is dedicated to the description of the BiFCA algorithm. Finally, in Section \ref{chapteexppositive}, we provide the results of the application of our algorithms on real-life microarray datasets.
\section{BiARM: Mining low overlapping bicluters using ARM }
\label{chapterBiARM}
Biclustering is a thriving task and of paramount importance in a lot of biomedical applications. Indeed, biclusters aim, among-others, to discover unveiling principles of cellular organizations and functions, to cite but a few.

In this section, we introduce a new algorithm called, \textit{\textbf{BiARM}}, which aims to efficiently extract the most meaningful, low overlapping biclusters. The main originality of our algorithm stands in the fact that it relies on the extraction of generic association rules. The reduced set of association rules faithfully mimics relationships between sets of genes, proteins, or other cell members and gives important information for the analysis of diseases. 
\subsection{BiARM algorithm}
The \textit{BiARM} \cite{Houari2015} biclustering algorithm is an ARM-based algorithm that identifies biclusters from gene expression data.
\textit{BiARM} operates in four main phases. The first one is the \textit{discretization phase} which consists in the binarization of the elements of the input data matrix. In this phase, we start by discretizing the initial numerical data matrix into a -101 data matrix which represents the relation between all conditions for the gene set in the gene expression matrix. This preprocessing step aims to highlight the trajectory patterns of genes. Then we discretize the $-101$ data matrix into a binary one in order to extract association rules from a binary context.
The second phase is the \textit{mining phase} where we extract the generic ARs that represent the bicluster's conditions. The third phase is the \textit{closing phase}; this one corresponds to the discovery of the supporting transactions (genes) from each rule in this base . Finally, we have the \textit{filtering phase} in which we compute the similarity measure. This latter is defined as the ratio between the conjunctive support of two biclusters and their disjunctive support where we consider only those having the Jaccard measure not exceeding a given threshold $minjaccard$. This is done in order to remove the biclusters that have a high overlap.

The pseudo-code description of \textit{BiARM} is shown in Algorithm \ref{algobiarm}.
\begin{algorithm}[!h] 
\caption{The \textsc{BiARM} Algorithm} \label{algobiarm}
\begin{algorithmic}[1]
\STATE Input: A gene expression matrix $M_{1}$,   \textit{minsupp}, \textit{minconf} and  \textit{minjaccard};\\
\STATE The set of biclusters $\mathcal{\beta}$;\\
\STATE \textbf{Begin}
\STATE $\mathcal{\beta}$ := $\emptyset$ ;\\
/* \textbf{First phase} */\\
\STATE
 Discretize $M_{1}$ using Equation \ref{disc1biarm} to obtain $M_{2}$ \label{step1} \;\\
/* \textbf{Second phase} */\\

\STATE Discretize $M_{2}$ using Equation \ref{discr2biarm} to obtain $M_{3}$ \label{step2} \;\\
/* \textbf{Third phase} */\\
\STATE
Extract all generic ARs using \textit{minsupp} and \textit{minconf}\label{step3_1}\;
\STATE
Extract genes that support the frequent items (the supporting transactions) // {\scriptsize obtained from line \ref{step3_1}}
\label{step3_2}\;\\
/* \textbf{Fourth phase} */\\

\STATE \textbf{For} each two biclusters B$_{i}$,B$_{j}$ obtained from the previous phase \textbf{do} 
\STATE \textbf{If} jaccard (B$_{i}$,B$_{j}$) $<$ \textit{minjaccard} \textbf{then}  
        \STATE \hspace{0.24cm} $\beta$ = $\beta$ $\bigcup$ $\{B_{i}   and   B_{j}\}$ \label{step4_1};
\STATE \textbf{Else}
         \STATE \hspace{0.24cm} $\beta$ = $\beta$ $\bigcup$ $\{B_{i} or B_{j}\}$\label{step4_2};
\\ \STATE  \textbf{Endfor}  \\       
\STATE \textbf{Return }\textbf{$\mathcal{\beta}$}\; \label{output1}\\
\STATE\textbf{End}
\end{algorithmic}
\end{algorithm}
\subsubsection{Phase 1: From numerical data to $-101$ data matrix}
Our method, at first, applies a preprocessing phase to transform the original data matrix $M_{1}$ into a $-101$ data matrix $M_{2}$. This phase aims to highlight the trajectory  patterns of genes. According to both \cite{DBLP:journals/bioinformatics/LuanL03} and \cite{SD2003}, in microarray data analysis, we add genes into a bicluster (cluster) whenever their trajectory patterns of expression levels are similar across a set of conditions. 

Interestingly enough our proposed discretization phase keeps track of the profile shape \footnote{Which may be either monotone increasing, monotone decreasing, up-down or down-up, etc.} over conditions and preserves the similarity information of trajectory patterns of the expression levels.

Before applying the ARM algorithm, we must first discretize the initial data matrix (Line \ref{step1}). The discretization process outputs the $-101$ data matrix. It consists in combining in pairs, for each gene, all the conditions between them. Indeed, the -101 data matrix gives an idea about the profile. Furthermore, one can have a global view of the profile of all conditions between them.

In our case, each column of the $-101$ data matrix represents the meaning of the variation in genes between a pair of conditions of $M_{1}$. The $-101$ data matrix offers useful information for the identification of biclusters,  i.e. up (1), down (-1) and no change (0).

Formally the matrix $M_{2}$ ($-101$ data matrix) is defined as follows :\\
\begin{align}
\label{disc1biarm}
M_{2}=
\begin{cases}
1  & \text{if   } \text{ } \text{ }     M_{1}[i,l]<M_{1}[i,l2]     \\
-1 & \text{if    }  \text{ } \text{ }      M_{1}[i,l]>M_{1}[i,l2] \\
0  & \text{if    }   \text{ }  \text{ }     M_{1}[i,l]=M_{1}[i,l2]
\end{cases}
\end{align}\\
\begin{list}{}{with:}
\item  $i$ $\in$ $\mathcal [1 \ldots n];$ $l$ $\in$ $\mathcal [1 \ldots m-1]$ ; $l2$ $\in$ $\mathcal [i+1 \ldots m];$ where  $n$ is the number of genes and $m$ is the number of conditions. 
\end{list}
 \medskip

 \subsubsection{Phase 2: From -101 data to binary data matrix}
 Let $M_{2}$ be a $-101$ data matrix. In order to build the binary data matrix, we compute the average number of repetitions for each column in the matrix $M_{2}$. It is better to choose the mean value since the maximum will produce a huge number of high overlapping biclusters, whereas the minimum value generates biologically none-valid biclusters.

After that, we define the binary matrix $M_{3}$ as follows:\\
\begin{align}
\label{discr2biarm}
M_{3}=
\begin{cases}
1  & \text{if   } \text{ } \text{ }     x_{1}= average\text{ }value     \\
0  & \text{ otherwise    }    
\end{cases}
\end{align} \\
          
\subsubsection{Phase 3: Extracting biclusters}
ARM can be viewed as a kind of biclustering for binary data. It provides extracting patterns (\textit{biclusters}) from a binary context. To perform this task they divide the problem into two sub-problems: \begin{enumerate}
\item Finding all association rules that represent biclusters' samples/genes. In fact, they consider items of both the premise and conclusion of an association rule.
\item Extracting the supporting transactions of these items.
\end{enumerate}
In this work, we use the $\mathcal{IGB}$ representation of the set of valid ARs defined by \cite{Gasmi2005}. Our choice of this base is justified by the theoretical framework presented in \cite{Gasmi2005}. We extract the generic ARs from the transactional representation, which represent the biclusters' conditions with respect to \textit{minconf} and \textit{minsupp} measures. The confidence is used as the homogeneity criteria. In other words, we try to increase the confidence and decrease the value of the support. After extracting the $\mathcal{IGB}$ base, we move to extract the supporting transactions (genes) from each rule in this base.

In this respect, after preparing the binary data matrix, we move to extract association rules (biclusters' conditions) from it. This task is divided into two sub-problems:

\begin{enumerate}
\item Finding all the generic ARs that represent bicluster's conditions (From $M_{3}$).
\item Extracting the genes for each obtained generic AR from $M_{3}$.\\ 
\end{enumerate}

\subsubsection{Phase 4: Similarity measure}
The \textit{BiARM} algorithm has already been able to identify overlapping biclusters. In order to compute the similarity between two biclusters $B1$ and $ B2$, we use the Jaccard measure. This latter measures the overlapping between two biclusters. This score is used to measure the overlap between two \textit{BiARM} biclusters in terms of both genes and conditions.
                      
In fact, for the filtering process, we consider only biclusters with a low overlap (if two biclusters have a high overlap, then they have the same biological signification).

The correlation measure achieves its minimum of 0 when biclusters do not overlap at all and attains its maximum of 1 when they are identical.

\subsection{Illustrative example}

Let us consider the gene expression data matrix $M_{1}$ given by Table \ref{exemplebiarm}.
\begin{itemize}
 \item \textbf{Phase 1:}
Using equation \ref{disc1biarm}, we represent the $-101$ data matrix ($M_{2}$) as shown in Table \ref{-101biarm}. 

  \begin{center}
                                                                                                                                                                                                       \begin{table}[!t]
 \centering
                                                                                                                        \begin{tabularx}{\linewidth}{|X|X|X|X|X|X|X|}
\hline  &$c_{1}$  & $c_{2}$ &$c_{3}$  &$c_{4}$  &$c_{5}$  &$c_{6}$  \\ 
\hline  $g_{1}$& 10 & 20  & 5  & 15  & 0  & 18  \\ 
 \hline  $g_{2}$& 20 & 30  & 15  & 25  & 26  & 25  \\ 
 \hline  $g_{3}$&  23& 12 & 8 &  15&  20 & 50  \\ 
  \hline  $g_{4}$&  30& 40 & 25 & 35 & 35 & 15 \\ 
  \hline  $g_{5}$& 13 & 13  & 18  & 25 & 30 & 55 \\ 
 \hline  $g_{6}$& 20 & 20  & 15  & 8 & 12 & 23 \\
 \hline 
 \end{tabularx}
  \captionof{table}{Example of gene expression matrix ($M_{1}$).}   
   \label{exemplebiarm}
   \end{table}
\end{center} 

\begin{center}
\begin{table}[!t]

\centering

\begin{tabularx}{\linewidth}{|X|X | X | X| X| X |X | X | X | X | X |X |X | X| X | X |}
\hline
& C$_{1}$ & C$_{2}$ & C$_{3}$ & C$_{4}$ & C$_{5}$ & C$_{6}$  & C$_{7}$ & C$_{8}$ & C$_{9}$ & C$_{10}$ & C$_{11}$ & C$_{12}$ & C$_{13}$ & C$_{14}$ & C$_{15}$ \\ 
\hline
 $g_{1}$  & 1 & -1 & 1 & -1 & 1 & -1 & -1 & -1 & -1 & 1 & -1 & 1 & -1 & 1 & 1  \\
  
 \hline   $g_{2}$ & 1 & -1 & 1 & 1 & 1 & -1 & -1 & -1 & -1 & 1 & 1 & 1 & 1 & 0 & -1 \\ 
 \hline   $g_{3}$& -1 & -1 & -1 & -1 & 1 & -1 & 1 & 1 &  1 & 1 & 1 & 1 & 1 & 1 & 1 \\ 
 \hline   $g_{4}$& 1 & -1 & 1 & 1 & -1 & -1 & -1 & -1 & -1 & 1 & 1 & -1 & 0 & -1 & -1 \\ 
 \hline   $g_{5}$& 0 & 1 & 1 & 1 & 1 & 1 & 1 & 1 & 1 & 1 & 1 & 1 & 1 & 1 & 1 \\ 
 \hline   $g_{6}$& 0 & -1 & -1 & -1 & 1 & -1 & -1 & -1 & 1 & -1 & -1 & 1 & 1 & 1 & 1 \\
 \hline 
 \end{tabularx} 
  \captionof{table}{$-101$ data matrix ($M_{2}$)}
\label{-101biarm}
\end{table}
 \end{center}
 Using Equation \ref{discr2biarm} we obtain the binary matrix sketched in Table \ref{binairebiarm}.
 
  \begin{center}
 \begin{table*}[!t]
 
 \centering
  \begin{tabularx}{\linewidth}{|X|X | X | X| X| X |X | X | X | X | X |X |X | X| X | X |}
   \hline   & C$_{1}$ & C$_{2}$ & C$_{3}$ & C$_{4}$ & C$_{5}$ & C$_{6}$  & C$_{7}$ & C$_{8}$ & C$_{9}$ & C$_{10}$ & C$_{11}$ & C$_{12}$ & C$_{13}$ & C$_{14}$ & C$_{15}$ \\ 
   \midrule $g_{1}$ & 0 & 0 & 0 & 1 & 0 & 0 & 0 & 0 & 1 & 0 & 1 & 0 & 1 & 0 & 0 \\ 
   \hline $g_{2}$& 0 & 0 & 0 & 0 & 0 & 0 & 0 & 0 & 1 & 0 & 0 & 0 & 0 & 0 & 1 \\ 
   \hline $g_{3}$ & 0 & 0 & 1 & 1 & 0 & 0 & 1 & 1 & 0 & 0 & 0 & 0 & 0 & 0 & 0 \\ 
   \hline $g_{4}$ & 0 & 0 & 0 & 0 & 1 & 0 & 0 & 0 & 1 & 0 & 0 & 1 & 0 & 1 & 1 \\ 
   \hline $g_{5}$ & 1 & 1 & 0 & 0 & 0 & 1 & 1 & 1 & 0  & 0 & 0 & 0 & 0 & 0 & 0 \\ 
   \hline $g_{6}$ & 1 & 0 & 1 & 1 & 0 & 0 & 0 & 0 & 0  & 1 & 1 & 0 & 0 & 0 & 0 \\                                                                                                                       \midrule   
 \end{tabularx}
  \captionof{table}{Binary data matrix ($M_{2}$)}
 \label{binairebiarm}
 \end{table*}
 \end{center}                                                                      
  \item \textbf{Phase 2:}              
 After preparing the binary data matrix, we move to extract biclusters from matrix $M_{3}$ (Table \ref{binairebiarm}).
  By using the previous example, we obtain as a result the association rules presented in Table \ref{IGBbasisbiarm}. Taking the example of rule $R1$, the obtained bicluster is $B1=<(g3,g4),(C_{3},C_{4})>$.
\end{itemize}

\begin{minipage}{.5\textwidth}\centering

  \begin{tabular}{|l|l|}
   \hline  \textbf{Transactions} & \textbf{Items} \\ 
   \hline $1$ & 4 9 11 13 \\ 
    \hline $2$& 9 15  \\ 
     \hline $3$ & 3 4 7 8  \\ 
      \hline $4$ & 5 9 12 14 15  \\ 
       \hline $5$ & 1 2 6 7 8 \\ 
       \hline $6$ & 1 3 4 10 11 \\                                                                                                                       \hline 
  \end{tabular}
  
  \captionof{table}{Transactional representation of binary data set given in Table \ref{binairebiarm}.}
   \label{basedetransaction}

  \end{minipage}
  \begin{minipage}{.5\textwidth}\centering

    \centering
    \begin{tabularx}{\linewidth}{| l| l | l |}
     \hline \textbf{Association rule}       & \textbf{Support}      & \textbf{Confidence} \\ 
    \hline $R1:  3 \Longrightarrow 4   $ & 0.33 & 1   \\ 
    \hline $R2:  7  \Longrightarrow 8$& 0.33 & 1  \\ 
    \hline $R3:  11  \Longrightarrow 4$ & 0.33 & 1  \\ 
    \hline $R4:  8  \Longrightarrow 7$ & 0.33 & 1   \\ 
    \hline $R5:  15  \Longrightarrow 9$ & 0.33 & 1   \\ 
    \hline   
   \end{tabularx}
   \captionof{table}{\textit{IGB} basis association rules extracted from Table \ref{basedetransaction} ($minsupp=40\%, minconf=80\%$).}
   \label{IGBbasisbiarm}

  \end{minipage}

  \subsection{Discussion} 
 We have presented our ARM-based biclustering algorithm (\textit{\textbf{BiARM}}) as a new biclustering algorithm for gene expression data. Our algorithm relies on the extraction of association rules from the dataset by discretizing this latter into a binary data matrix. However, a significant number of redundant rules has been found. Thus, and to remedy this problem, we have introduced the second contribution. This latter consists in proposing a new solution based on FCA.
\section{BiFCA+: Towards biclustering gene expression data with FCA}
\label{chapterBiFCA+}
We introduce a new algorithm, called BiFCA+, for biclustering microarray data. BiFCA+ heavily relies on the mathematical  background of FCA, in order to extract the biclusters' set. In addition, the \textit{Bond} correlation measure is of use to filter out the overlapping biclusters. 
\subsection{BiFCA+ algorithm}
The \textit{BiFCA+} \cite{Houari2018} biclustering algorithm is an FCA-based algorithm that identifies biclusters from gene expression data. It operates in three main phases. The first one is the \textit{discretization phase}. Starting from a numerical dataset, the basic idea is to build a formal context where  genes stand for objects and conditions for attributes.
Subsequently, it starts the \textit{mining phase}. This latter allows extracting formal concepts that represent correlated biclusters. Finally, we have to perform the \textit{filtering phase}. Given the biclusters obtained from the previous phase, we compute the similarity measure between each pair of biclusters. This latter is defined as the ratio between the conjunctive support of two biclusters and their disjunctive support. We only retain the biclusters having the \textit{Bond} correlation measure not exceeding a given threshold $minbond$.  This latter is performed in order to remove the biclusters that have a high overlap. The pseudo-code of \textit{BiFCA+} is shown in Algorithm \ref{algoBiFCA+}. \textit{BiFCA+} takes as an input a data matrix $M_{1}$ and a minimum correlation threshold $minbond$. BiFCA+ allows the determination, from the data matrix $M_{1}$, of the set of the obtained biclusters $\mathcal{\beta}$.

The  phases of BiFCA+ are thoroughly described in the following subsections.

 
 \begin{algorithm}[!h] 
 \caption{\textsc{BiFCA+} Algorithm} \label{algoBiFCA+}
 \begin{algorithmic}[1]
 \STATE Input: A gene expression data matrix $ \mathcal M_{1}$, \textit{minbond} ;\\
 
 \STATE Output: The set of biclusters $\mathcal{\beta}$ ;\\
\STATE \textbf{Begin}\\
 \STATE $\mathcal{\beta}$ := $\emptyset$;
  
   /* \textbf{First phase} */\\
   \STATE   Discretize $ \mathcal M_{1}$ using Equation \ref{discr1bifca+} to obtain $ \mathcal M_{2}$ ; // {\footnotesize The 3-state data matrix.} \label{step1} \\
    \STATE  Discretize $ \mathcal M_{2}$ using Equation \ref{discr2bifca+} to obtain $ \mathcal M_{3}$ \label{step2} ; // {\footnotesize The binary data matrix.} \\
     /* \textbf{Second phase} */\\
   \STATE  Extract the set of formal concepts FCs\label{step3}; // \\ 
     /* \textbf{The Third phase} */\\
     \FOR {each two biclusters FC$_{i}$= $\langle A_{i}, B_{i} \rangle $ and FC$_{j}$= $\langle A_{j}, B_{j} \rangle $ }
     \STATE \textbf{if} Bond($B$$_{i}$,$B$$_{j}$) $>$ \textit{minbond} \textbf{then}   \hspace{0.75cm} 
              \STATE \hspace{0.24cm}  $\beta$ = $\beta$ $\bigcup$ $\{FC_{i} or FC_{j}\};$ // {\footnotesize The bicluster with the highest number of samples.};
      \STATE \textbf{else}
               \STATE \hspace{0.24cm} $\beta$ = $\beta$ $\bigcup$ $\{FC_{i} and FC_{j}\};$ // {\footnotesize Consider FC$_{i}$ and FC$_{j}$ as biclusters.}\label{step4_2};
      
      \STATE \textbf{end if}

     \ENDFOR \\
    \STATE \textbf{Return} \textbf{$\mathcal{\beta}$}\ ; 
\STATE\textbf{End}

 \end{algorithmic}
 \end{algorithm}


\subsubsection{Phase 1: Pre-processing of gene expression data matrix}

\label{descritize1}
Our method applies a pre-processing phase to transform the original data matrix $M_{1}$ into a binary one. This phase is split into two steps:
\begin{enumerate}
\item First, we discretize the original data into a 3-state data matrix $M_{2}$. This step aims to unveil the trajectory  patterns of genes. According to  \cite{DBLP:journals/bioinformatics/LuanL03} and \cite{SD2003}, in DNA microarray data analysis, we add genes into a bicluster whenever their trajectory patterns of expression levels are similar across a set of samples.

Interestingly enough, our proposed discretization phase keeps track of the profile shape\footnote{Which may be either monotone increasing, monotone decreasing, up-down or down-up, etc.} over conditions and preserves the similarity information of trajectory patterns of the expression levels.

                                                                                                                           Before delving through the mining process, we must at first discretize the initial data matrix. The discretization process outputs the 3-state data matrix. It consists in combining in pairs, for each gene, all the adjacent conditions. Indeed, the 3-state data matrix gives an idea about the profile. Furthermore, it gives a global view of the profile of all conditions.

In our case, each column of the 3-state data matrix carries the meaning of the variation in genes between a pair of conditions of $M_{1}$. It offers useful information for the identification of biclusters, i.e. up (1), down (-1) and no change (0).

Formally, matrix $M_{2}$ (3-state data matrix) is defined as follows :\\
                  
                                                                                                                       \begin{align}
                     \label{discr1bifca+}
                                                                                                                                             M_{2}=
                                                                                                                                             \begin{cases}
                                                                                                                                             	1  & \text{if   } \text{ } \text{ }     M_{1}\mathcal [j,l]< M_{1}$$\mathcal [j,l+1];     \\
                                                                                                                                             	-1 & \text{if    }  \text{ } \text{ }       M_{1}\mathcal [j,l]> M_{1}\mathcal [j,l+1]; \\
                                                                                                                                            	0  & \text{if    }   \text{ }  \text{ }     M_{1}\mathcal [j,l]= M_{1}\mathcal [j,l+1];
                                                                                                                                              \end{cases}
                                                                                                                                              \end{align}
                                                                                                                                             with $j$ $\in$ $\mathcal [1 \ldots n];$ $l$ $\in$ $\mathcal [1 \ldots m-1]$                                                                                                                                 


\item For the second step of the pre-processing phase, we build the binary data matrix in order to extract formal concepts. In this respect, we compute the average number of repetitions for each column in matrix $M_{2}$ (3-state data matrix). In other words, we have:
 \begin{enumerate}
 \item $\rvert$\textit{maxrepeat}$\rvert$: This variable stands for the maximal number of occurrences by column.
 \item $\rvert$\textit{minrepeat}$\rvert$: This variable stands for minimal number of occurrences by column. 
 \item $\rvert$\textit{mediumrepeat}$\rvert$:  It stands for the medium number of occurrences by column.
 \end{enumerate}


   Formally, we define the binary matrix as follows:\\
                    \begin{align}
                     \label{discr2bifca+}
                     M_{3}=
                    \begin{cases}
                    	1  & \text{if   } \text{ } \text{ }     M_{2}$$\mathcal[j,l]= average\text{ } value     \\
                       	0  & \text{ otherwise    }    
                    \end{cases}
                    \end{align} 
                      with    $j$ $\in$ $\mathcal [1 \ldots n]$ and$ $ $l$ $\in$ $\mathcal [1 \ldots m-1]$   
                    
After the discretization phase, the dimensions of our data matrix ($M_{2}$) become equal to $n*(m-1)$.
\end{enumerate}
 \begin{example}
 Let us consider the data matrix $M_{1}$ given by Table \ref{exemple1}. For the first row, we have $M_{1_{1j}}$= (10, 20, 5, 15, 0, 18) with $j$ $\in$ $\mathcal [1 \ldots 6]$. In the first step of the pre-processing phase we obtain the discretized first row, i.e. $M_{2_{1j}}$= (1, -1, 1, -1, 1) with $j$ $\in$ $\mathcal [1 \ldots 5]$. In the second step, the first column becomes $M_{3_{i1}}$= (0, 0, 0, 0, 1, 1) with $i$ $\in$ $\mathcal [1 \ldots 6]$.
 \end{example}  
\subsubsection{Phase 2: Extracting formal concepts (biclusters)}
\label{extractingconcepts}
  
 FCA can be viewed as a kind of biclustering for binary data. It provides extracting patterns (\textit{biclusters}) from a binary context.

In this respect, after preparing the binary data matrix, we move to extract formal concepts (biclusters) from the binary matrix $M_{3}$. 

The extraction of the formal concepts is carried out through the invocation of a slightly modified version of the efficient LCM algorithm \cite{Uno2004}. The choice of this algorithm is argued by the fact that it has a linear complexity in the number of closed attributes and has been shown to be one of the best algorithms dedicated to such a task.
\subsubsection{Phase 3: Computation of the similarity measure (\textit{Bond})}
\label{bondmeasure}
The BiFCA+ algorithm has been already able to identify overlapping biclusters.                                                                                                                  
In fact, for the filtering process, we only consider biclusters having a low overlap. Indeed, for biclusters that have a high overlap, they have the same biological signification.
                                                                                                                                               The \textit{Bond} correlation measure achieves its minimum of 0 when the biclusters do not overlap at all and and attains maximum value 1 whenever they are identical.
                                                                                                                                               
                                                                                                                                               In order to compute the similarity between two biclusters (i.e, formal concepts) FC$_{1}$ and FC$_{2}$, FC$_{1}$= $\langle A_{1}, B_{1} \rangle $ and FC$_{2}$= $\langle A_{2}, B_{2} \rangle $, where A$_{i}$, $i=1,2$, represents the extent and B$_{i}$ represents the intent, we use the \textit{Bond} correlation measure. The latter assesses the overlap between two concept's intents (cf. Definition \ref{defbond}).
            
Finally, we only retain the obtained biclusters for which the \textit{Bond} correlation measure does not exceed a given threshold. The set of such biclusters represents a solution to our problem.

 In the following, we provide an illustrative example of the BiFCA+ approach.
\subsection{Illustrative example}

 Let us consider the data matrix given by Table \ref{exemplebifca+}. Each column represents all the gene expression levels from a single experiment, and each row represents the expression of a gene across all experiments. 
 \subsubsection{Pre-processing phase of data matrix}
 The pre-processing phase goes as follows:

 \begin{enumerate}
 \item First, we transform the numerical data into the 3-state data matrix. This is done using Equation \ref{discr1bifca+}. Table \ref{-101bifca+} represents the obtained results.
 \item Second, we create the binary matrix, using the 3-state data matrix. Let us consider the 3-state data matrix given by Table \ref{-101bifca+}. For the sake of building the binary data matrix, we compute the average number of repetitions for each column in the matrix $M_{2}$; e.g., for the column $ \grave C_{1} $ we have:
  \begin{enumerate}
  \item $\rvert$\textit{maxrepeat}$\rvert$: is set equal to 3 and  corresponds to the value $1$.
  \item $\rvert$\textit{minrepeat}$\rvert$: is set equal to 1 and corresponds to the value $-1$. 
  \item $\rvert$\textit{mediumrepeat}$\rvert$: Mediumrepeat is $2$ and corresponds to the value $0$.  So, the \textit{average value} is 0. \end{enumerate}

   Subsequently, and using Equation \ref{discr2bifca+}, we obtain the binary matrix sketched by Table \ref{binairebifca+}.
   \end{enumerate}
    \begin{center}
      \begin{table*}[!t]
      
      \centering
      \begin{tabularx}{\linewidth}{|X|X|X|X|X|X|X|}
       \hline  &$c1$  & $c2$ &$c3$  &$c4$  &$c5$  &$c6$  \\ 
       \hline  $g_{1}$& 10 & 20  & 5  & 15  & 0  & 18  \\ 
       \hline  $g_{2}$& 20 & 30  & 15  & 25  & 26  & 25  \\ 
       \hline  $g_{3}$&  23& 12 & 8 &  15&  20 & 50  \\ 
       \hline  $g_{4}$&  30& 40 & 25 & 35 & 35 & 15 \\ 
       \hline  $g_{5}$& 13 & 13  & 18  & 25 & 30 & 55 \\ 
       \hline  $g_{6}$& 20 & 20  & 15  & 8 & 12 & 23 \\
       \hline 
      \end{tabularx}
       \captionof{table}{Example of gene expression data matrix ($M_{1}$).}
      \label{exemplebifca+}
      \end{table*}  
   \end{center} 

\begin{center}
\begin{table*}[!t]

\centering
 \begin{tabularx}{\linewidth}{| X| X | X | X | X | X |}
 \hline & $\grave C_{1}$ & $\grave C_{2}$ & $\grave C_{3}$ & $\grave C_{4}$ & $\grave C_{5}$  \\ 
  \hline   $g_{1}$ & 1 & -1 & 1 & -1 & 1   \\ 
  \hline   $g_{2}$ & 1 & -1 & 1 & 1 & -1  \\ 
  \hline   $g_{3}$& -1 & -1 & 1 & 1 & 1  \\ 
  \hline   $g_{4}$& 1 & -1 & 1 & 0 & -1  \\ 
  \hline   $g_{5}$& 0 & 1 & 1 & 1 & 1  \\ 
   \hline   $g_{6}$& 0 & -1 & -1 & 1 & 1  \\
    \hline 
 \end{tabularx}
 \captionof{table}{3-state data matrix ($M_{2}$).} 
 \label{-101bifca+}
 \end{table*}
\end{center}

  \begin{center}
 \begin{table*}[!t]
 
  \centering
  \begin{tabularx}{\linewidth}{| X| X | X | X | X | X |}
   \hline & $\grave C_{1}$ & $\grave C_{2}$ & $\grave C_{3}$ & $\grave C_{4}$ & $\grave C_{5}$ \\ 
   \hline   $g_{1}$& 0 & 0 & 0 & 1 & 0   \\ 
   \hline   $g_{2}$& 0 & 0 & 0 & 0 & 1  \\ 
   \hline   $g_{3}$& 0 & 0 & 0 & 0 & 0  \\ 
   \hline   $g_{4}$& 0 & 0 & 0 & 0 & 1  \\ 
   \hline   $g_{5}$& 1 & 1 & 0 & 0 & 0  \\ 
   \hline   $g_{6}$& 1 & 0 & 1 & 0 & 0  \\
   \hline 
  \end{tabularx}
   \captionof{table}{Binary data matrix ($M_{3}$).}
 \label{binairebifca+}
 \end{table*}
 \end{center}

 \subsubsection{Formal concept extraction phase}
 After preparing the binary data matrix, we move to extract formal concepts, i.e. the candidate biclusters, from the matrix $M_{3}$.\\
                                                                                                                                                                                                                                                      By using the binary data matrix given in Table \ref{binairebifca+}, we obtain as a result the formal concepts shown in Table \ref{conceptbifca+}.
                                                                                                                                                                                                                                                 
                                                                                                                           \begin{table}[H]
\hspace{+4.4 cm} 
\begin{tabularx}{\linewidth}{| l| l | l |}
 \hline FCs & {\textit{Extent (genes)}} & {\textit{Intent (conditions)}} \\ 
\hline  FC$_{1}$ &  $g_{5}g_{6}$ & $\grave C_{1}$\\ 
\hline  FC$_{2}$ & $g_{2}g_{4}$& $\grave C_{5}$ \\ 
\hline  FC$_{3}$  &$g_{5}$ & $\grave C_{1}$$\grave C_{2}$ \\ 
\hline  FC$_{4}$  & $g_{6}$ & $\grave C_{1}$$\grave C_{3}$ \\ 
 \hline  FC$_{5}$  & $g_{1}$ & $\grave C_{4}$ \\ 
\hline 
\end{tabularx}
\captionof{table}{Formal concepts extracted from binary context given in Table \ref{binairebifca+} .}                                                                               \label{conceptbifca+}
\end{table}
                                                                                                                              
                      
\subsubsection{Filtering phase} 
 In this phase, we only retain biclusters having a low overlap. This overlap is assessed through the \textit{Bond} correlation measure. For example, with respect to the formal concepts shown in Table \ref{conceptbifca+}, if we consider FC$_{3}$ and FC$_{4}$, we compute the \textit{Bond} correlation measure:

                                                                                                                                                 $ \textit{Bond}$ ($B_{3}$,$B_{4}$) = $\frac{ | \{\grave C_{1}\grave C_{2}\} \bigcap  \{\grave C_{1}\grave C_{3}\} |}{|\{\grave C_{1}\grave C_{2}\} \bigcup  \{\grave C_{1}\grave C_{3}\}|}$\\
                                                                                                                                                  
                                                                                                                                                $\textit{Bond}$ ($B_{3}$, $B_{4}$) = $\frac{1}{3} $ = 0.33\\
                                                                                                                                                 \\
                                                                                                                                                The \textit{Bond} correlation measure threshold is set equal to 0.5. Therefore, we consider the formal concepts $FC_{3}$ and $FC_{4}$ as non overlapping biclusters.
                                                                                                                           Nevertheless, by lowering the threshold value to 0.3, we only consider one bicluster, i.e. the one having the highest number of conditions.

 \subsection{Discussion} 
 A new FCA-based biclustering method for gene expression data has been proposed. Our approach consists in extracting formal concepts from a dataset after a discretization into a 3-state data matrix then into a binary one. A 3-state data matrix allows observing the profile of each gene through all pairs of adjacent conditions in the gene expression matrix. However, a close look at existing studies proves that our results will be much improved if we extend the discretization of all columns and not only those which are adjacent.

 Thus, in the next section we present BiFCA, an optimized version of the BiFCA+ algorithm. Our improvement covers the first phase of BiFCA+ (discretization phase) where we combine in pairs, for each gene, all condition’s pairs. In fact, the 3-state data matrix gives an idea about the profile shape. Furthermore, we can have a global view of the profile shape of all conditions.

\section{BiFCA: Mining biclusters using FCA }
\label{chapterBiFCA}

 In this section, we introduce a new algorithm for extracting biclusters from microarray data. Our algorithm relies on FCA, which has been shown to be an efficient methodology for biclustering binary data.The performance of our algorithm is evaluated on real-life DNA microarray datasets.

\subsection{BiFCA algorithm}
The BiFCA \cite{Houari2015a} biclustering algorithm relies on FCA. BiFCA operates in four main phases: (1) We start by discritizing the initial numerical data matrix into a -101 data matrix \footnote{The -101 data matrix is the matrix obtained after the discritization of the original data matrix.} which represents the relation between all conditions for the gene set in the gene expression matrix. (2) Then we discritize the -101 data matrix into a binary one in order to (3) extract formal concepts (candidate biclusters). (4) Finally, we compute the bond measure which is defined as the ratio between a conjunctive support of a concept and its disjunctive support, and we consider only those having  the bond measure not exceeding a given threshold $\alpha$, done, in order to remove concepts that have high overlapping.\\
 The pseudo-code description of the BiFCA algorithm is shown in Algorithm \ref{algoBiFCA}.
 \begin{algorithm}[!h] 
 \caption{The \textsc{BiFCA} Algorithm} \label{algoBiFCA}
 \begin{algorithmic}[1]
 \STATE Input: A gene expression data matrix $M_{1}$, \textit{minbond}, \textit{mincondition} ;\\
 
 \STATE Output: The set of biclusters $\mathcal{\beta}$ ;\\
\STATE \textbf{Begin}\\
 \STATE $\mathcal{\beta}$ := $\emptyset$;
  
   /* \textbf{First phase} */\\
   \STATE   Discretize $ M_{1}$ using Equation \ref{disc1biarm} to obtain $M_{2}$ ; // {\footnotesize The -101 data matrix.} \label{step1} \\
    \STATE  Discretize $M_{2}$ using Equation \ref{discr2biarm} to obtain $M_{3}$ \label{step2} ; // {\footnotesize The binary data matrix.} \\
     /* \textbf{Second phase} */\\
   \STATE  Extract the set of formal concepts FCs\label{step3}; // \\ 
     /* \textbf{Third phase} */\\
     \FOR {each two biclusters FC$_{i}$= $\langle A_{i}, B_{i} \rangle $ and FC$_{j}$= $\langle A_{j}, B_{j} \rangle $ }
     \STATE \textbf{if} ((Bond($B$$_{i}$,$B$$_{j}$) $>$ \textit{minbond}) and (\textit{nbcondition}$>$\textit{mincondition} )) \textbf{then}   \hspace{0.75cm} 
              \STATE \hspace{0.24cm}  $\beta$ = $\beta$ $\bigcup$ $\{FC_{i} or FC_{j}\};$ // {\footnotesize The bicluster with the highest number of samples.};
      \STATE \textbf{else}
               \STATE \hspace{0.24cm} $\beta$ = $\beta$ $\bigcup$ $\{FC_{i} and FC_{j}\};$ // {\footnotesize Consider FC$_{i}$ and FC$_{j}$ as biclusters.}\label{step4_2};
      
      \STATE \textbf{end if}

     \ENDFOR \\
    \STATE \textbf{Return} \textbf{$\mathcal{\beta}$}\ ; 
\STATE\textbf{End}

 \end{algorithmic}
 \end{algorithm}

   \begin{center}
                                                                                                                                                                                                                                                 \begin{table}[H]
                                                                                                                                                                                                                                                    \centering
                                                                                                                                                                                                                                                    \begin{tabularx}{\linewidth}{|X|X|X|X|X|X|X|}
                                                                                                                                                                                                                                                      \hline  &$c1$  & $c2$ &$c3$  &$c4$  &$c5$  &$c6$  \\ 
                                                                                                                                                                                                                                                      \hline  $g_{1}$& 10 & 20  & 5  & 15  & 0  & 18  \\ 
                                                                                                                                                                                                                                                      \hline  $g_{2}$& 20 & 30  & 15  & 25  & 26  & 25  \\ 
                                                                                                                                                                                                                                                      \hline  $g_{3}$&  23& 12 & 8 &  15&  20 & 50  \\ 
                                                                                                                                                                                                                                                      \hline  $g_{4}$&  30& 40 & 25 & 35 & 35 & 15 \\ 
                                                                                                                                                                                                                                                      \hline  $g5$& 13 & 13  & 18  & 25 & 30 & 55 \\ 
                                                                                                                                                                                                                                                        \hline  $g6$& 20 & 20  & 15  & 8 & 12 & 23 \\
                                                                                                                                                                                                                                                      \hline 
                                                                                                                                                                                                                                                    \end{tabularx}
                                                                                                                                                                                                                                                   \captionof{table}{Example of gene expression matrix ($M_{1}$).}
                                                                                                                                                                                                                                                    \label{exempleBifca}
       \vspace{-0.8cm}                                                                                                                                                                                                                                           \end{table}
                                                                                                                         \vspace{-0.8cm}                                                                                                                         \end{center}

 \vspace{-0.10cm}                                                                                                                                                                                                                                     \subsubsection{Phase 1: From numerical data to -101 data matrix}
                                                                                                                                                                                                                                              
                                                                                                                                                                                                    Before starting the mining phase, our method first applies a preprocessing phase to transform the original data matrix $M_{1}$ into a -101 data matrix $M_{2}$ (matrix of behavior). To do this, we use Equation \ref{disc1biarm}.
                                                                                                                                                                                 \subsubsection{Phase 2: From -101 data to binary data matrix}
For the second phase of the pre-processing phase, we build the binary data matrix in order to extract formal concepts. In this respect, we compute the average number of repetitions for each column in matrix $M_{2}$. Afterwards, we define the binary data matrix using Equation \ref{discr2biarm}
 \subsubsection{Phase 3: Extracting formal concepts}

        After preparing the binary data matrix, we move to extract formal concepts (biclusters) from the binary matrix $M_{3}$. 
        
        As mentioned before, the extraction of formal concepts is carried out through the invocation of a slightly modified version of the efficient LCM algorithm.
\subsubsection{Phase 4: \textit{Bond} correlation measure}
 Before computing the \textit{Bond} correlation measure between two concepts' intents, we filter the obtained concepts by taking into consideration that the number of conditions is higher than a given threshold, due to the overwhelming number of generated formal concepts.

 The BiFCA algorithm is already able to identify overlapping biclusters. In order to compute the similarity between two biclusters (concepts) C1 and C2, C1= (I$_{1}$,E$_{1}$), C2= (I$_{2}$,E$_{2}$), where I$_{i}$, $i=1,2$, represents the intent and E represents the extent, we use the \textit{Bond} correlation measure. The latter measures the overlapping between two concepts \textit{i.e.} biclusters.
                                                                   
                                          In fact, for the filtering process, we consider only biclusters with a low overlap (if biclusters have a high overlapping, they have the same biological signification).
                                       
\subsection{Illustrative example}
Let us consider the dataset given by Table \ref{exempleBifca}. 
   \begin{itemize}
\item \textbf{Phase 1:} Using Equation \ref{disc1biarm} we represent the -101 data matrix (Table \ref{-101bifca}).                                                                                   \begin{center}
                                                                                                                                     \begin{table*}[!t]
                                                                                                                                                                                                                                                 \centering
                                                                                                                                                                                                                                                 \begin{tabularx}{\linewidth}{| X| X | X | X | X | X | X | X | X | X | X | X | X | X | X |X |}
\hline 
& C$_{1}$ & C$_{2}$ & C$_{3}$ & C$_{4}$ & C$_{5}$ & C$_{6}$  & C$_{7}$ & C$_{8}$ & C$_{9}$ & C$_{10}$ & C$_{11}$ & C$_{12}$ & C$_{13}$ & C$_{14}$ & C$_{15}$ \\ 
 
\hline
$g_{1}$        & 1 & -1 & 1 & -1 & 1 & -1 & -1 & -1 & -1 & 1 & -1 & 1 & -1 & 1 & 1  \\ 
\hline   $g_{2}$ & 1 & -1 & 1 & 1 & 1 & -1 & -1 & -1 & -1 & 1 & 1 & 1 & 1 & 0 & -1 \\ 
\hline   $g_{3}$& -1 & -1 & -1 & -1 & 1 & -1 & 1 & 1 &  1 & 1 & 1 & 1 & 1 & 1 & 1 \\ 
\hline   $g_{4}$& 1 & -1 & 1 & 1 & -1 & -1 & -1 & -1 & -1 & 1 & 1 & -1 & 0 & -1 & -1 \\ 
\hline   $g_{5}$& 0 & 1 & 1 & 1 & 1 & 1 & 1 & 1 & 1 & 1 & 1 & 1 & 1 & 1 & 1 \\ 
\hline   $g_{6}$& 0 & -1 & -1 & -1 & 1 & -1 & -1 & -1 & 1 & -1 & -1 & 1 & 1 & 1 & 1 \\
\hline 
 \end{tabularx}
\captionof{table}{-101 data matrix ($M_{2}$)}
\label{-101bifca}
\end{table*}
\end{center}

\begin{center}
\begin{table*}[!t]
\centering
\begin{tabularx}{\linewidth}{| X| X | X | X | X | X | X | X | X | X | X | X | X | X | X |X |}
\hline 
& C$_{1}$ & C$_{2}$ & C$_{3}$ & C$_{4}$ & C$_{5}$ & C$_{6}$  & C$_{7}$ & C$_{8}$ & C$_{9}$ & C$_{10}$ & C$_{11}$ & C$_{12}$ & C$_{13}$ & C$_{14}$ & C$_{15}$ \\ 

\hline $g_{1}$ & 0 & 0 & 0 & 1 & 0 & 0 & 0 & 0 & 1 & 0 & 1 & 0 & 1 & 0 & 0 \\ 
\hline $g_{2}$& 0 & 0 & 0 & 0 & 0 & 0 & 0 & 0 & 1 & 0 & 0 & 0 & 0 & 0 & 1 \\ 
\hline $g_{3}$ & 0 & 0 & 1 & 1 & 0 & 0 & 1 & 1 & 0 & 0 & 0 & 0 & 0 & 0 & 0 \\ 
\hline $g_{4}$ & 0 & 0 & 0 & 0 & 1 & 0 & 0 & 0 & 1 & 0 & 0 & 1 & 0 & 1 & 1 \\ 
\hline $g_{5}$ & 1 & 1 & 0 & 0 & 0 & 1 & 1 & 1 & 0  & 0 & 0 & 0 & 0 & 0 & 0 \\ 
\hline $g_{6}$ & 1 & 0 & 1 & 1 & 0 & 0 & 0 & 0 & 0  & 1 & 1 & 0 & 0 & 0 & 0 \\        
                                                                                  \hline   
\end{tabularx}
\captionof{table}{Binary data matrix ($M_{3}$)}
\label{binairebifca}
\end{table*}
\end{center}

\item \textbf{Phase 2:}                   
Let $M_{2}$ be a -101 data matrix (e.g Table \ref{-101bifca}). In order to build the binary data matrix, we compute the average number of repetitions for each column in matrix $M_{2}$; for example, for the column $C_{1}$ we have:\\
                                                                                                                                                                                                                                               maxrepeat \footnote{Maximum number of occurrences by column.} is 3 and  corresponds to the maxvalue 1.\\
                                                                                                                                                                                                                                               minrepeat \footnote{Minimum number of occurrences by column.} is 1 and corresponds to the minvalue -1. \\
                                                                                                                                                                                                                                               And mediamrepeat$=$2 and corresponds to the value 0. \\
  So, the \textit{average value} is 0. Passing to the binary matrix, column $C_{1}$ becomes \{0,0,0,0,1,1\}.

                                               Using Equation \ref{discr2biarm}, we obtain the binary matrix (Table \ref{binairebifca}).
                             
 \item \textbf{Phase 3:}
By using the previous example we obtain as a result the concepts in Table \ref{conceptbifca}. 
\begin{center}
\begin{table}[H]

\hspace{+4.5 cm}
\begin{tabularx}{\linewidth}{| l| l | l |}
\hline  & \textbf{Intents (conditions)} & \textbf{Extents (genes)} \\ 
\hline FC1 & $C_{4}$& $g_{1},g_{3},g_{6}$ \\ 
\hline  FC2 & $C_{9}$& $g_{1},g_{2},g_{4}$ \\ 
\hline  FC3& $C_{1}$&$g_{5},g_{6}$ \\ 
\hline FC4 & $C_{3}$,$C_{4}$ & $g_{3},g_{6}$ \\ 
\hline  FC5& $C_{8}$,$C_{7}$ & $g_{3},g_{5}$ \\ 
\hline FC6 & $C_{8}$,$C_{3}$,$C_{7}$,$C_{4}$ & $g_{3}$ \\ 
\hline FC7 & $C_{11}$,$C_{4}$ & $g_{1},g_{6}$ \\ 
\hline FC8 & $C_{15}$,$C_{9}$ & $g_{2},g_{4}$ \\
\hline FC9 & $C_{6}$,$C_{1}$,$C_{7}$,$C_{8}$,$C_{2}$ & $g_{5}$ \\
\hline FC10 & $C_{10}$,$C_{4}$,$C_{1}$,$C_{3}$,$C_{11}$ & $g_{6}$ \\
\hline FC11 & $C_{13}$,$C_{4}$,$C_{9}$,$C_{11}$ & $g_{1}$ \\
\hline FC12 & $C_{14}$,$C_{9}$,$C_{15}$,$C_{5}$,$C_{12}$ & $g_{4}$ \\
\hline 
\end{tabularx}
\captionof{table}{Formal concepts extracted from binary context given in Table \ref{binairebifca}.}
\label{conceptbifca}
\end{table}
\end{center}  
   
\item \textbf{Phase 4:}                                         Using the example of concept5 and concept6(example from Table \ref{conceptbifca}), we compute the bond measure :\\
                                          
bond($I_{5}$,$I_{6}$)= $\frac{ | \{C_{7},C_{8}\} \bigcap  \{C_{3},C_{4},C_{7},C_{8}\} |}{|\{C_{7},C_{8}\} \bigcup  \{C_{3},C_{4},C_{7},C_{8}\}|} $

bond($I_{5}$,$I_{6}$)=$\frac{2}{4} $ = 0.50.

Taking 0.6 as a \textit{Bond} threshold, we consider both concepts $FC_{5}$ and $FC_{6}$ as non overlapping biclusters.
                                                                                    While taking 0.3 as a \textit{Bond} threshold, we consider only $FC_{6}$ as a bicluster, i.e. which has the highest number of conditions.
\end{itemize}

\section{Experimental results}
\label{chapteexppositive}
In this section we provide the experimental results of using our algorithm on three well-known real-life datasets. The evaluation of the biclustering algorithms and their comparison are based on two criteria:\textit{ Statistical} and \textit{Biological} (\textit{cf.} Section \ref{Statistical_validation_description}).
\subsection{Description of used datasets }
\label{useddatabases}
In order to assess the performance of our proposed algorithm and analyze its results, we carry out a series of experimentations on the following real-life gene expression datasets:

\begin{itemize}
\item \textit{Yeast Cell-Cycle dataset:}  The Yeast Cell-Cycle\footnote{Available at \url{http ://arep.med.harvard.edu/biclustering/}.} is a very popular dataset  in the gene expression analysis community. In fact, it is one of the most studied organisms, and the functions of each gene are well known. We use the \textit{Yeast Cell-Cycle dataset} which was described in \cite{Tavazoieand1999}, processed in \cite{DBLP:conf/ismb/ChengC00} and publicly available from \cite{Cheng2006}. It contains 2884 genes and 17 samples. 

\item \textit{Saccharomyces Cerevisiae dataset:} The Saccharomyces Cerevisiae dataset\footnote{Available at \url{ http://www.tik.ethz.ch/sop/bimax/}.} contains the expression levels of 2993 genes under 173 samples. 
\item \textit{Human B-cell Lymphoma dataset:} The Human B-cell Lymphoma dataset \cite{Alizadeh2000} contains the expression levels of 4026 genes under 96 samples\footnote{Available at \url{ http://arep.med.harvard.edu/biclustering/ }.}. 

\end{itemize}

\subsection{Experimental protocol}
\label{Expprotocol}
The first series of experiments concerns statistical validation. In this series, we compute the coverage for the Yeast Cell-Cycle and Human B-cell Lymphoma datasets and compute the adjusted \textit{p-value} for the Yeast Cell-Cycle and Saccharomyces Cerevisiae datasets. The second series of experiments is applied the to Yeast Cell-Cycle and Saccharomyces Cerevisiae datasets in order to study the biological significance of extracted biclusters.

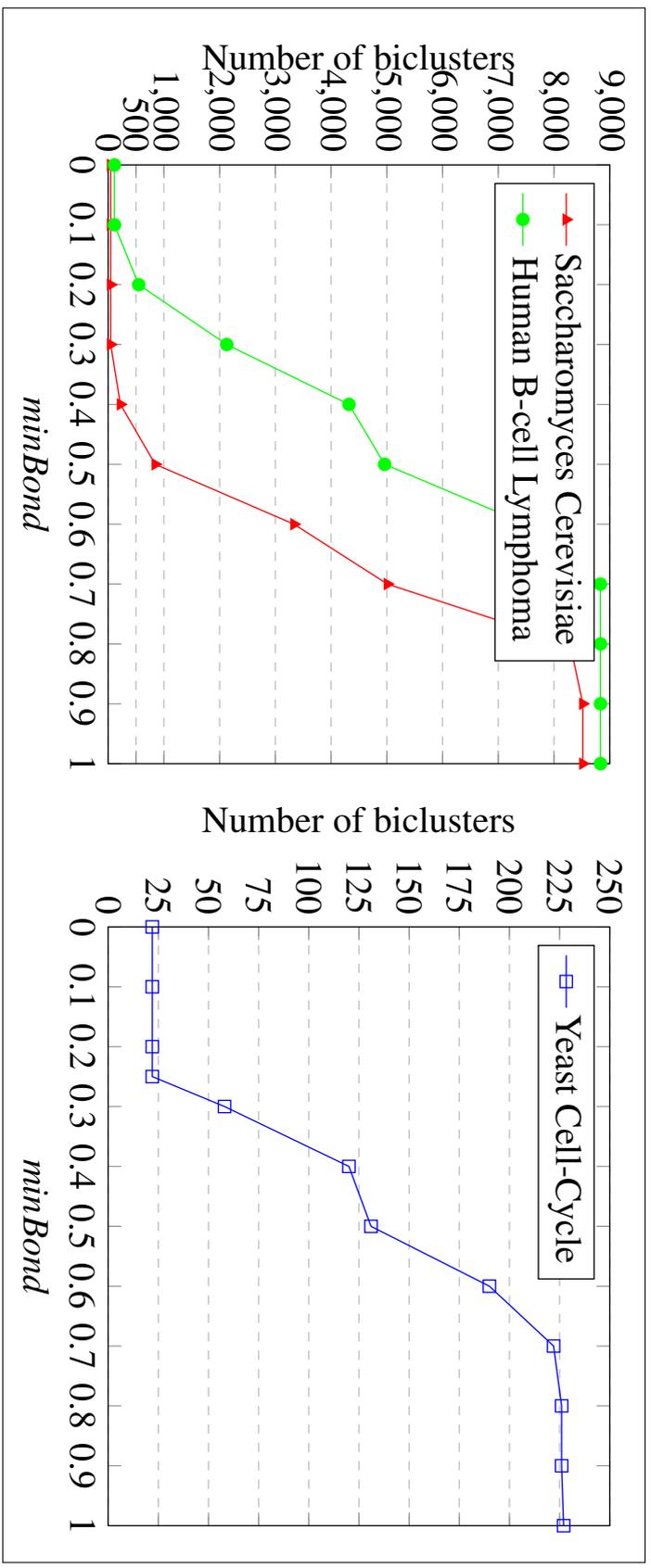
\begin{sidewaysfigure}[p] 

          \begin{center}
          
             \fbox{
             \scalebox{1.25}{
               \begin{tikzpicture}
                 \begin{axis}[
                     xlabel={$minBond$},
                     ylabel={Number of biclusters },
                     xmin=0, xmax=1,
                     ymin=0, ymax=9000,
                     xtick={0,0.10,0.20,0.30,0.4,0.5,0.6,0.7,0.8,0.9,1},
                     ytick={0,500,1000,2000,3000,4000,5000,6000,7000,8000,9000},
                     legend pos=north west,
                     ymajorgrids=true,
                     grid style=dashed,
                 ]

                 \addplot[
                             color=red,
                             mark=triangle*,
                              ]
                coordinates {
              (1,8516)(0.9,8516)(0.8,8289)(0.7,5012)(0.6,3335)(0.5,845)(0.4,221)(0.3,46)(0.2,46)(0.1,46)(0,46)
                                            
                             };           
                  \addplot[
                                          color=green,
                                          mark= otimes*,
                                           ]
               coordinates {
          (1,8835)(0.9,8834)(0.8,8834)(0.7,8832)(0.6,7505)(0.5,4960)(0.4,4321)(0.3,2127)(0.2,549)(0.1,113)(0,113)
                                                         
                                          }; 
              {\footnotesize  \legend{Saccharomyces Cerevisiae,Human B-cell Lymphoma}}

                 \end{axis}
                 \end{tikzpicture}

                 \begin{tikzpicture}
                              \begin{axis}[
                                  xlabel={$minBond$},
                                  ylabel={Number of biclusters },
                                  xmin=0, xmax=1,
                                  ymin=0, ymax=250,
                                  xtick={0,0.10,0.20,0.30,0.4,0.5,0.6,0.7,0.8,0.9,1},
                                  ytick={0,25,50,75,100,125,150,175,200,225,250},
                                  legend pos=north west,
                                  ymajorgrids=true,
                                  grid style=dashed,
                              ]
                             \addplot[
                             color=blue,
                             mark=square,
                              ]
                             coordinates {
                             (1,227)(0.9,226)(0.8,226)(0.7,222)(0.6,190)(0.5,131)(0.4,120)(0.3,58)(0.25,22)(0.2,22)(0.1,22)(0,22)
                                            
                             };
                            
                            \legend{Yeast Cell-Cycle}

                              \end{axis}
                              \end{tikzpicture}}
                              }
                  \end{center}
                \caption{Number of BiFCA+ biclusters w.r.t. $minBond$ variations.}
                \label{courbebond}
       
         \end{sidewaysfigure}

Table \ref {parametresalgopos} summarizes the different values of the parameters used by our algorithms. The values of these parameters are chosen experimentally. Regarding the compared algorithms, we report the results of some algorithms from \cite{wassimayadi2011}, whereas we execute the Trimax algorithm.
   \begin{table*}[htbp]
     \centering
    
       \begin{tabular}{|l|l|l|}
       \toprule
       \textbf{Algorithms} & \textbf{Algorithms}  & \textbf{Parameters}  \\
       \midrule
       Yeast cell cycle & BiARM  &  minsupp =20\%\\
        &   & minconf =80\% \\
         &   & minjaccard=25\% \\
          &   &  \\
                        & BiFCA+             & minbond = 0.25\\
                          &   &  \\
                        & BiFCA               & minbond=20\% \\
                          &   &  minconditions=3\\
                          &   &  \\ 
                        & Trimax               & $\theta$=1, minobject=115, minattribute=2\\
                                                  &   & maxobject=2884, maxattribute=17\\ 
        \midrule
         Saccharomyces Cerevisiae & BiARM  &  minsupp =30\%\\
                 &   & minconf =90\% \\
                  &   & minjaccard=20\% \\
                   &   &  \\
                                & BiFCA+             & minbond = 0.3\\
                                  &   &  \\
                                & BiFCA               & minbond=25\%\\ 
                                  &   & minconditions=5\\
                                    &   &  \\
                                & Trimax               & $\theta$=1, minobject=210, minattribute=20\\
                                                                                  &   & maxobject=2993, maxattribute=173\\  
        \midrule
          Human B-cell Lymphoma & BiARM  &  minsupp =20\%\\
                           &   & minconf =90\% \\
                            &   & minjaccard=25\% \\
                             &   &  \\
                                & BiFCA+             & minbond = 0.1\\
                                  &   &  \\
                                & BiFCA               & minbond=30\% \\
                                 &   & minconditions=14 \\
                                  &   &  \\
                                & Trimax               & $\theta$=1, minobject=300, minattribute=2\\
                                                                                                                  &   & maxobject=4025, maxattribute=96\\    
    
       \bottomrule
       
       \end{tabular}
       \caption{Setting algorithms for real-life datasets for different algorithms.}
     \label{parametresalgopos}%
   \end{table*}%
 The experiments are carried out on different datasets. For instance for the BiFCA+ algorithms, according to the obtained experimental results, interesting reductions in the number of biclusters are obtained as far as the value of $minBond$ is lowered. Representative results are plotted by Figure \ref{courbebond}, where the $minBond$ is set when there is a significant decrease in the number of obtained biclusters.
\subsection{Statistical relevance }
\label{statisticalresultspos}  
In this section, we show the results of applying our approaches on three well-known real-life datasets. The evaluation of biclustering algorithms and their comparison are based on two criteria: \textit{Coverage} and \textit{P-value}. We compare the results obtained by our algorithm versus the state-of-the-art biclustering algorithms as well as the Trimax algorithm\footnote{Available at \url{https://github.com/mehdi-kaytoue/trimax}.} \cite{Kaytoue2014}, which also relies on FCA.
\subsubsection{Coverage:}
\label{sectioncoveragepos}
As carried out by \cite{Ayadi2012,Liu2009} and \cite{Liu2008}, we use the coverage criterion which is defined as the total number of cells in a microarray data matrix covered by the obtained biclusters.  In the biclustering domain, validation using coverage is considered as worthy of interest since a large coverage of a dataset is very important in several applications that rely on biclusters \cite{Freitas20013}. In fact, the higher the number of highlighted correlations, the greater the amount of extracted information. Consequently, the higher the coverage, the lower the overlapping in biclusters. We compare the results of our algorithm versus those obtained by Trimax \cite{Kaytoue2014} and those reported by \cite{wassimayadi2011}. In the latter reference, the following algorithms were considered: CC \cite{DBLP:conf/ismb/ChengC00}, BiMine \cite{Ayadi2009}, BiMine+ \cite{Ayadi2012a}, BicFinder \cite{Ayadi2012}, MOPSOB \cite{Liu2008}, MOEA \cite{Mitra2006} and  SEBI  \cite{Divina2006}.

In the literature, this test has been applied respectively on the Yeast Cell-Cycle and Human B-cell Lymphoma datasets. \footnote{The Human B-cell Lymphoma dataset version that we have does not contain the names of genes to perform other tests. }

  Table \ref{coveragelymphomapos} (resp. Table \ref{coverageyastpos}) presents the coverage of the obtained biclusters. At a glance, we remark that most of the algorithms have  relatively close results. For the Human B-cell Lymphoma (respectively the Yeast Cell-Cycle) dataset, the biclusters extracted by our algorithm BiFCA+ cover 100\% (respectively 80.12\%) of the genes, 100\% of the conditions and 67.84\% (respectively 57.07\%) of the cells of the expression data matrix. Trimax is largely outperformed, since it only covers respectively 8.50 \% of cells, 46.32 \% of genes and 11.46 \% of conditions for the Human B-cell Lymphoma. It is also worth mentioning that for Yeast Cell-Cycle, the  CC algorithm obtains the best results since it masks groups extracted with random values. Thus, it prohibits the genes/conditions that were previously discovered from being selected during the next search process. This type of mask leads to high coverage. Furthermore, The Yeast dataset only contains positive integer values. Thus, one can use the Mean Squared Residue (MSR) \cite{DBLP:conf/ismb/ChengC00} to extract large biclusters. By contrast, the Human B-cell Lymphoma dataset contains integer values including negative ones. This means that the application of MSR on this dataset does not lead the extraction of large biclusters.  This implies that our algorithm can generate biclusters with high coverage of a data matrix. This outstanding coverage is caused by the discretization phase as well as the extraction of biclusters without focusing on a specific type of biclusters.

  As well as, for the BiARM algorithm, for the Human B-cell Lymphoma dataset, the biclusters extracted by our algorithm cover 99.97\% of the genes, 100\% of the conditions and 73.12\% of the cells in the initial matrix. However, Trimax has a low performance since it covers only 8.50 \% of cells, 46.32 \% of genes and 11.46 \% of conditions. This implies that our algorithm can generate biclusters with high coverage of a data matrix due to the discretisation phase where the combinations of all the paired conditions give useful information since a bicluster may be composed of a subset of non contiguous conditions.
  
   \begin{table*}[htbp]
     \centering
    
       \begin{tabular}{|l|r|r|r|}
       \toprule
       \multicolumn{4}{|c|}{\textbf{Human B-cell Lymphoma}} \\
       \midrule
       \textbf{Algorithms} & \textbf{Total coverage} & \textbf{Gene coverage} & \textbf{Condition coverage} \\
       \midrule
       BiMine \cite{Ayadi2009} & 8.93\% & 26.15\% & 100\% \\
       BiMine+ \cite{Ayadi2012a}& 21.19\% & 46.26\% & 100\% \\
       BicFinder \cite{Ayadi2012} & 44.24\% & 55.89\% & 100\% \\
       MOPSOB \cite{Liu2008}& 36.90\% & \_    & \_ \\
       MOEA \cite{Mitra2006} & 20.96\% & \_    & \_ \\
       SEBI \cite{Divina2006} & 34.07\% & 38.23\% & 100\% \\
       CC \cite{DBLP:conf/ismb/ChengC00}    & 36.81\% & 91.58\% & 100\% \\
      Trimax \cite{Kaytoue2014}& 8.50\% & 46.32\% & 11.46\% \\
       \midrule
       \textbf{BiARM} &\textbf{ 73.12 \%} & \textbf{99.97\%} & \textbf{100\%} \\
      \textbf{BiFCA+} &\textbf{ 67.84 \%} & \textbf{100\%} & \textbf{100\%} \\
      \textbf{BiFCA} &\textbf{76.08\%} & \textbf{100\%} & \textbf{100\%} \\
       \bottomrule
       
       \end{tabular}
       \caption{Human B-cell Lymphoma coverage for different algorithms.}
     \label{coveragelymphomapos}%
   \end{table*}%
    
     \begin{table*}[htbp]
        \centering
         
          \begin{tabular}{|l|r|r|r|}
          \toprule
          \multicolumn{4}{|c|}{\textbf{Yeast Cell-Cycle}} \\
          \midrule
          \textbf{Algorithms} & \textbf{Total coverage} & \textbf{Gene coverage} & \textbf{Condition coverage} \\
          \midrule
          BiMine \cite{Ayadi2009} & 13.36\% & 32.84\% & 100\% \\
          BiMine+ \cite{Ayadi2012a}& 51.76\% & 68.65\% & 100\% \\
          BicFinder \cite{Ayadi2012}& 55.43\% & 76.93\% & 100\% \\
          MOPSOB \cite{Liu2008}& 52.40\% & \_    & \_ \\
          MOEA \cite{Mitra2006} & 51.34\% & \_    & \_ \\
          SEBI \cite{Divina2006} & 38.14\% & 43.55\% & 100\% \\
          CC \cite{DBLP:conf/ismb/ChengC00}   & \textbf{81.47}\% & \textbf{97.12}\% & 100\% \\
           Trimax \cite{Kaytoue2014} & 15.32\% & 22.09\% & 70.59\% \\
           \midrule
           \textbf{BiARM} &\textbf{72.03\%} & \textbf{98.2\%} & \textbf{100\%} \\
         \textbf{BiFCA+} & \textbf{57.07 \%} & \textbf{80.12\%} & \textbf{\textbf{100\%}} \\
        \textbf{BiFCA} &\textbf{ 75.32\%} & \textbf{100\%} & \textbf{100\%} \\
          \bottomrule
          
          \end{tabular}
      \caption{Yeast Cell-Cycle coverage for different algorithms.}
        \label{coverageyastpos}%
      \end{table*}%

\subsubsection{P-value:}
To assess the quality of the extracted biclusters, we use the web tool \textit{\textbf{FuncAssociate}} \cite{Berriz2003} in order to compute the adjusted significance scores for each bicluster (adjusted \textit{p-value}\footnote{The adjusted significance scores asses genes in each bicluster, which indicates how well they match with the different GO categories.}). In fact,  the best biclusters have an adjusted \textit{p-value} less than 0.001\%. The results of our algorithm are compared versus those obtained by Trimax \cite{Kaytoue2014} as well as those concerning CC \cite{DBLP:conf/ismb/ChengC00}, ISA \cite{DBLP:journals/bioinformatics/IhmelsBB04}, OSPM \cite{DBLP:journals/jcb/Ben-DorCKY03} and Bimax \cite{Prelic2006}, reported by \cite{wassimayadi2011}.

In the literature, this test is applied respectively on the Yeast Cell-Cycle and Saccharomyces Cerevisiae datasets.

\begin{figure*}[htb]
         \begin{center}
        \fbox{  \includegraphics[width=1.00\textwidth]{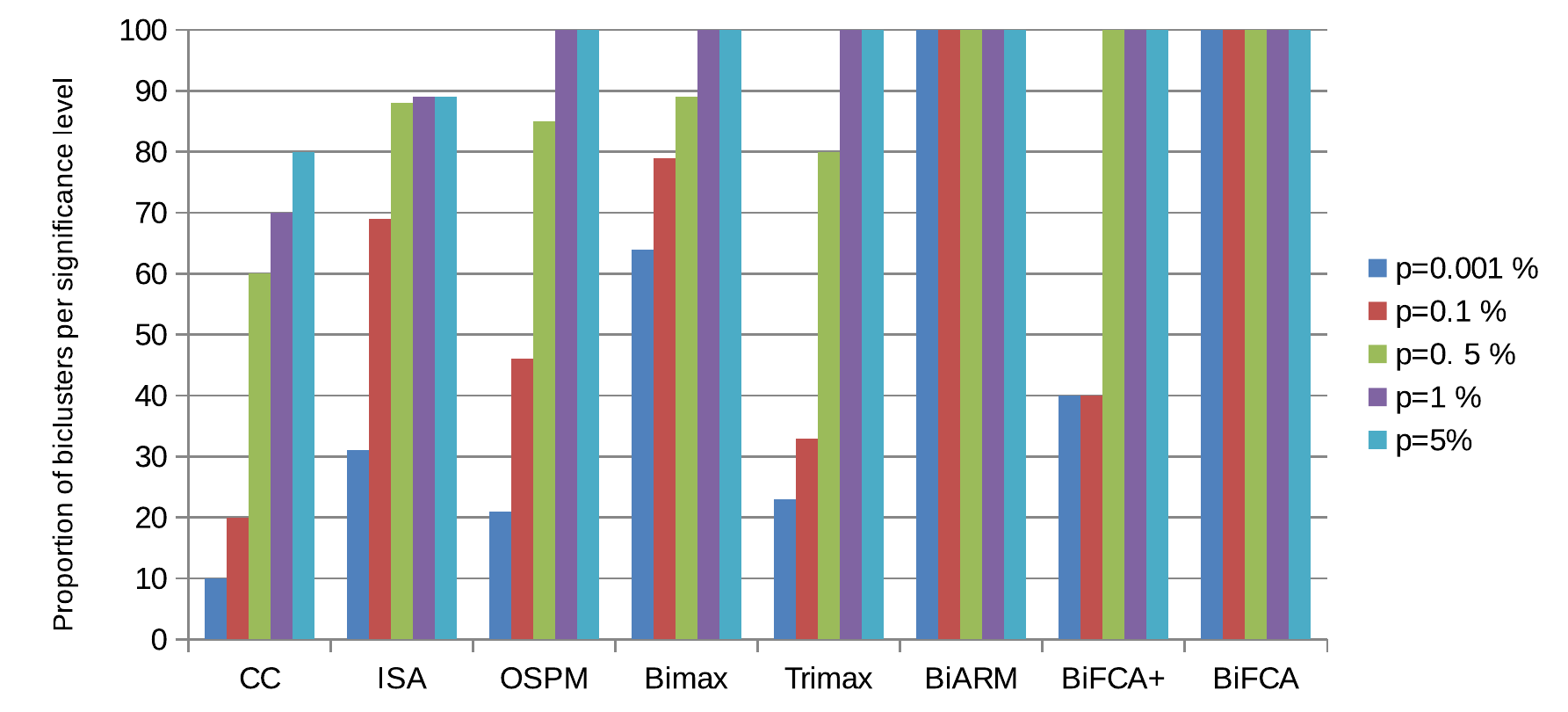}} 
         \end{center}
         \caption{Proportions of biclusters significantly enriched by GO annotations (Yeast cell-cycle dataset)}
         \label{pvaleuryeastpos}
         \end{figure*}

\begin{figure*}[htb]                                                                                                     \begin{center}
      \fbox{ \includegraphics[width=1.0\textwidth]{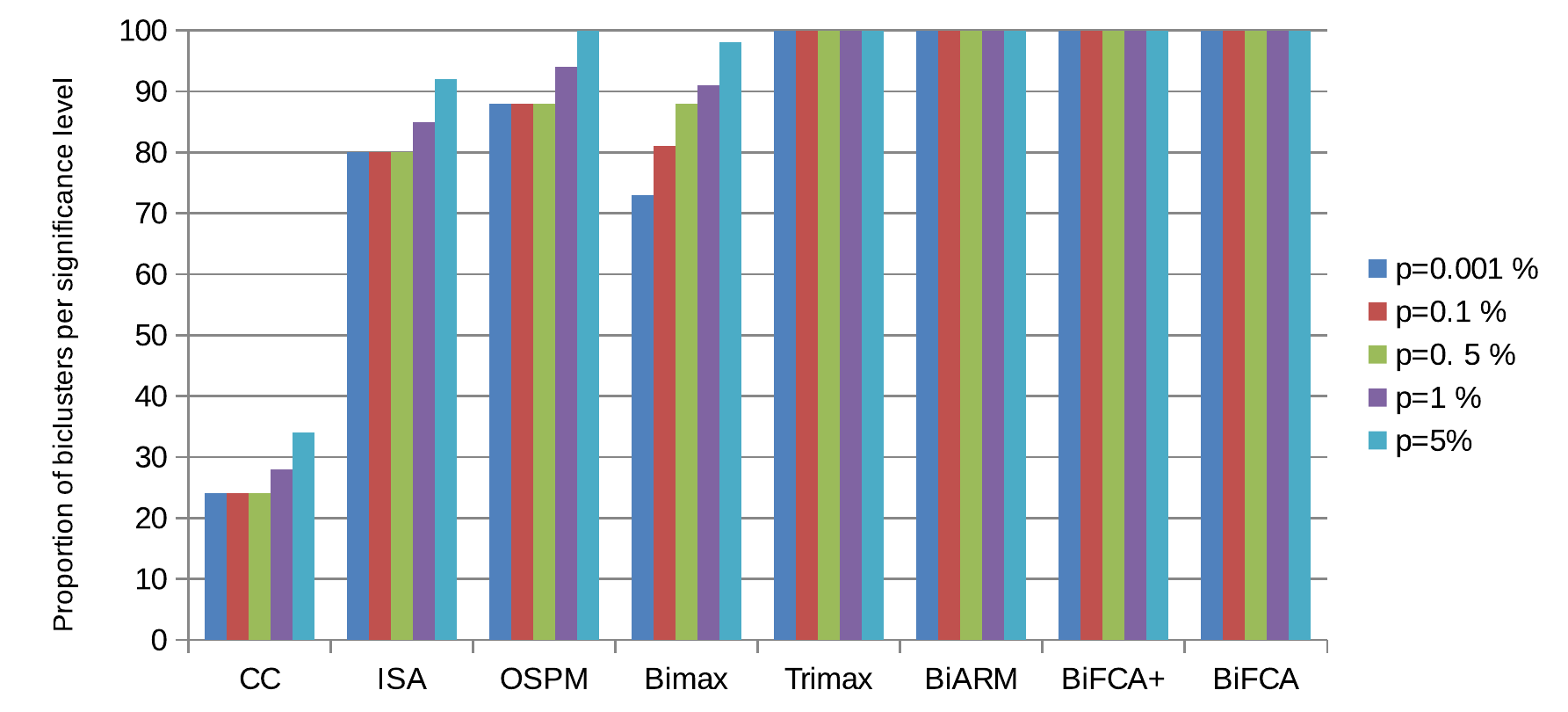}}    
      \end{center}
      \caption{Proportions of biclusters significantly enriched by GO annotations (Saccharomyces Cerevisiae dataset)}
      \label{pvaleursacpos}
    
      \end{figure*}

    The obtained results of the Yeast Cell Cycle and the Saccharomyces Cerevisiae datasets for different adjusted  \textit{p-values} ($p$ = 5\%; 1\%; 0.5\%; 0.1\%; 0.001\%), for each algorithm over the percentage of total biclusters, are respectively depicted in Figure \ref{pvaleuryeastpos} and Figure \ref{pvaleursacpos}.

    For the Saccharomyces Cerevisiae dataset (Figure \ref{pvaleursacpos}), the BiFCA+ (resp. BiARM and BiFCA)  and Trimax results show that 100\% of extracted biclusters are statistically significant with the adjusted \textit{p-value} equal to $0.001 \%$. It is important to note that Bimax achieves its best results whenever $p<0.1\%$, while CC, ISA and OSPM have a reasonable performance with $p<0.5\%$.

    On the other hand, for the Yeast Cell Cycle (Figure \ref{pvaleuryeastpos}), 100\% of the extracted biclusters by BiFCA+ are statistically significant when $p <0.5\%$, while only 80\% of extracted biclusters by Trimax are statistically significant for the same \textit{p-value}. By contrast, Trimax achieves 100\% of extracted biclusters when $p <1 \%$. Our results, then, sharply outperform those of Trimax. However, Bimax scored better when $p <0.001\% $ and $p <0.1 \%$.

  Whereas, for the BiARM alogortihm, the obtained results of the Yeast Cell Cycle dataset for the different adjusted \textit{p-values} (p = 5\%; 1\%; 0,5\%; 0,1\%; 0,001\%) for each algorithm over the percentage of total biclusters are depicted in Figure \ref{pvaleuryeastpos}. The \textit{BiARM} results show that 100\% of the extracted biclusters are statistically significant with the adjusted \textit{p-value} $p <0.001 \%$. Contrarily, Trimax achieves 100\% of statistically significant biclusters when $p <1 \%$. It is important to note that Bimax achieves its best results when $p<0.1\%$, while CC, ISA and OSPM have a reasonable performance with $p<0.5\%$.

 For the BiFCA algorithm, the obtained results for the different adjusted \textit{p-values} for each algorithm over the percentage of total biclusters are depicted in Figure\ref{pvaleuryeastpos}. The BiFCA result shows that 100\% of extracted biclusters are statistically significant with an adjusted \textit{p-value}, where $p <0.001 \%$. Worthy of mention, the best of the other compared algorithms is BiMine when $p<0.1\%$, while CC, ISA and OSPM have a reasonable performance with $p<0.5\%$.
                                                                                                          
\subsection{Biological relevance }
\label{biologicalresultspos}
The biological criterion allows measuring the quality of resulting biclusters, by checking whether the genes of a bicluster have common biological characteristics.
                                                                                                    
                                                                                                     This test is applied respectively on the Yeast Cell-Cycle and Saccharomyces Cerevisiae datasets.                   
                                                                                                         
\begin{figure*}[htb]  
         
           \begin{center}

              
              \scalebox{1.0}{
        \includegraphics[width=0.9\textwidth]{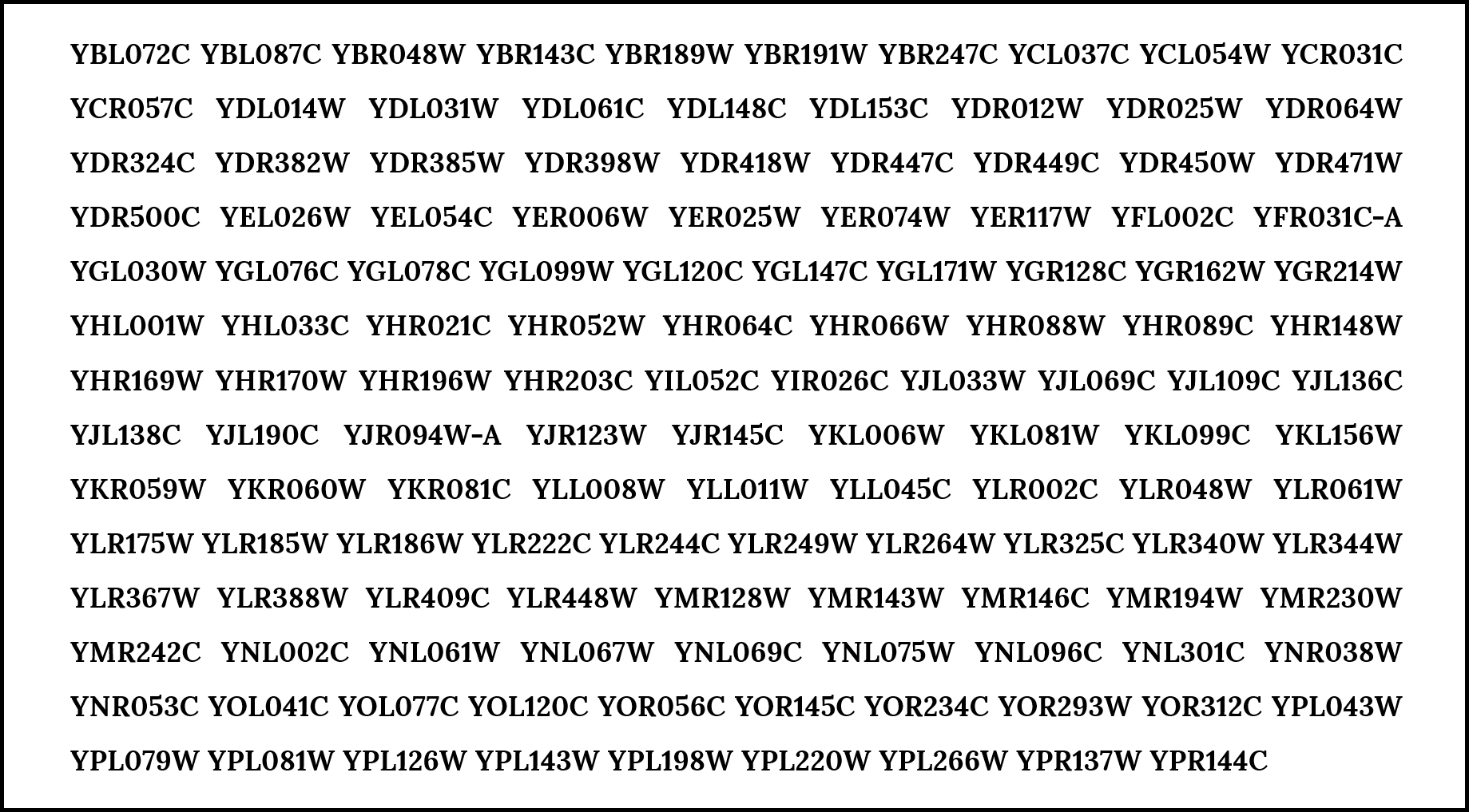}
                           }
                               
                   \end{center}
                 \caption{List of genes which concern the \textit{GO} term \textit{"ribonucleoprotein complex"} (\textit{cellular component}) for the first bicluster (Saccharomyces Cerevisiae dataset).}
                 \label{gene_list}
          \end{figure*}

To evaluate the quality of the extracted biclusters and identify their biological annotations, we use \textbf{\textit{GOTermFinder}}, which is designed to search for the significant shared \textit{Gene Ontology} (GO) terms of a group of genes. The GO is organized according to 3 axes: \textit{biological process}, \textit{molecular function} and \textit{cellular component}\footnote{http://geneontology.org/}. We show in Tables \ref{GOsacc} and \ref{goyeastbifca+} the biological annotations of two randomly selected biclusters in terms of above cited axis, where we report the most significant GO terms. For instance, with the first bicluster extracted from the Saccharomyces Cerevisiae dataset (Table \ref{GOsacc}), the list of genes is illustrated in Figure \ref{gene_list}.
These genes concern the \textit{GO} term \textit{"ribonucleoprotein complex"}, in terms of \textit{cellular component} with a \textit{p-value} equal to 1.39$e^{-65}$ (highly significant) and a background of 10.7\%.

 The results on these real-life datasets show that our proposed algorithm identifies biclusters with a high biological relevance.
 \begin{landscape}
      \begin{table*}[htbp]
             \centering
            
               \begin{tabular}{l|r|r}
               \toprule
                     & \multicolumn{1}{c|}{\textbf{Bicluster 1}} & \multicolumn{1}{c}{\textbf{Bicluster 2}} \\
               \midrule
               \textbf{Biological process} & \multicolumn{1}{l|}{cytoplasmic translation \textbf{(2.4\%, 1.24e-06)}} & \multicolumn{1}{l}{single-organism process \textbf{(49.3\%, 0.09388)}} \\
                     & \multicolumn{1}{l|}{single-organism process \textbf{(49.3\%, 1.82e-05)}} & \multicolumn{1}{l}{} \\
                     & \multicolumn{1}{l|}{cell cycle process \textbf{(8.4\%, 5.39e-05)}} & \multicolumn{1}{l}{} \\
                     & \multicolumn{1}{l|}{} & \multicolumn{1}{l}{} \\
                      \midrule
               \textbf{Molecular function} & \multicolumn{1}{l|}{structural molecule activity \textbf{(4.8\%, 0.00238)} } & \multicolumn{1}{l}{N-methyltransferase activity \textbf{(0.5 \%, 0.2794)}} \\
                     & \multicolumn{1}{l|}{structural constituent of ribosome \textbf{(3.1\%, 0.00310)}} & \multicolumn{1}{l}{transferase activity, transferring one-carbon  } \\
                     & \multicolumn{1}{l|}{} & \multicolumn{1}{l}{groups \textbf{(1.4\%, 0.06158)}} \\
                \midrule
              \textbf{ Cellular component} & \multicolumn{1}{l|}{cytosolic ribosome \textbf{(2.5\%, 7.85e-07)}} & \multicolumn{1}{l}{nuclear chromatin \textbf{(1.8\%, 0.00817)}} \\
                     & \multicolumn{1}{l|}{non-membrane-bounded organelle \textbf{(18.3\%, 3.05e-06)}} & \multicolumn{1}{l}{cytosolic part \textbf{(3.4 \%, 0.01229)}} \\
                     & \multicolumn{1}{l|}{intracellular non-membrane-bounded organelle} & \multicolumn{1}{l}{cytosolic ribosome \textbf{(2.5\%, 0.05227)}} \\
                     & \multicolumn{1}{l|}{\textbf{(18.3\%, 3.05e-06)}} & \multicolumn{1}{l}{} \\
               \bottomrule
               \end{tabular}%
                \caption{Significant GO terms (process, function, component) for two biclusters, extracted from Yeast Cell-Cycle data set using BiFCA+.}
             \label{goyeastbifca+}%
           \end{table*}%
     \end{landscape}      
           
          \begin{landscape} 
              \begin{table*}[htbp]
                \centering
                
                  \begin{tabular}{l|r|r}
                  \toprule
                  \multicolumn{1}{l|}{} & \multicolumn{1}{c|}{\textbf{Bicluster 1}} & \multicolumn{1}{c}{\textbf{Bicluster 2}} \\
                  \midrule
                  \multicolumn{1}{c|}{\textbf{Biological process}} & \multicolumn{1}{l|}{ribosome biogenesis \textbf{(5.8\%, 2.17e-61)}} & \multicolumn{1}{l}{single-organism process \textbf{(49.3\%, 7.64e-43)}} \\
                  \multicolumn{1}{c|}{} & \multicolumn{1}{l|}{ncRNA processing \textbf{(5.8\%, 8.44e-57)}} & \multicolumn{1}{l}{single-organism cellular process (\textbf{43.0\%, 4.61e-30)}} \\
                  \multicolumn{1}{c|}{} & \multicolumn{1}{l|}{ribonucleoprotein complex biogenesis} & \multicolumn{1}{l}{single-organism metabolic process \textbf{(25.5\%, 7.13e-25)}} \\
                  \multicolumn{1}{c|}{} & \multicolumn{1}{l|}{\textbf{(6.9\%,  2.02e-55)}} & \multicolumn{1}{l}{} \\
                  \multicolumn{1}{c|}{} & \multicolumn{1}{l|}{} & \multicolumn{1}{l}{} \\
                  \midrule
                  \multicolumn{1}{c|}{\textbf{Molecular function}} & \multicolumn{1}{l|}{structural constituent of ribosome } & \multicolumn{1}{l}{oxidoreductase activity \textbf{(3.9\%, 3.45e-14)}} \\
                  \multicolumn{1}{c|}{} & \multicolumn{1}{l|}{\textbf{(3.1\%,  4.03e-42)}} &  \\
                  \multicolumn{1}{c|}{} & \multicolumn{1}{l|}{structural molecule activity\textbf{(4.8\%, 9.14e-33)}} & \multicolumn{1}{l}{transmembrane transporter activity \textbf{(4.5\%, 7.98e-14)}} \\
                  \multicolumn{1}{c|}{} & \multicolumn{1}{l|}{RNA helicase activity \textbf{(0.6\%, 2.37e-13)}} & \multicolumn{1}{l}{substrate-specific transmembrane transporter } \\
                        & \multicolumn{1}{l|}{} & \multicolumn{1}{l}{activity \textbf{(4.1\%, 5.67e-12)}} \\
                        &       &  \\
                        \midrule
                  \multicolumn{1}{c|}{\textbf{Cellular component}} & \multicolumn{1}{l|}{ribonucleoprotein complex\textbf{(10.7\%, 1.39e-65)}} & \multicolumn{1}{l}{mitochondrial part \textbf{(7.2\%,  1.42e-19)} } \\
                  \multicolumn{1}{c|}{} & \multicolumn{1}{l|}{preribosome \textbf{(2.4\%, 3.28e-61)}} & \multicolumn{1}{l}{mitochondrion \textbf{(16.2\%, 1.51e-19)}} \\
                  \multicolumn{1}{c|}{} & \multicolumn{1}{l|}{cytosolic ribosome (\textbf{2.5\%, 4.67e-55)}} & \multicolumn{1}{l}{cell part \textbf{(77.2\%, 3.72e-15)}} \\
                  \bottomrule
                  \end{tabular}%
                  \caption{Significant GO terms (process, function, component) for two biclusters, extracted from Saccharomyces Cerevisiae data using BiFCA+.} 
               \label{GOsacc}%
              \end{table*}%
                \end{landscape}

We show in Table \ref{yeastBiARM} the result of a random selected set of genes for the biological process, the molecular function and the cellular component. We report the most significant GO terms. The values within parentheses after each GO term in Table \ref{yeastBiARM}, such as (15.0\%, 2.4\%,8.39e-51) in the first bicluster, indicate the cluster frequency, the background frequency and the statistical significance,  respectively. The cluster frequency shows that for the first bicluster 15.0\% of genes belong to this process, while background frequency shows that this bicluster contains 2.4\% of the number of genes in the background set and the statistical significance is provided by a \textit{p-value} of 8.39e-51 (highly significant).

The results on these real datasets demonstrate that our proposed algorithm can identify biclusters with a high biological relevance. 
 \begin{landscape} 
  \begin{table*}[htbp]
     \centering
     
     \begin{tabular}{rrr}
        \toprule
              & \multicolumn{1}{c}{\textbf{Bicluster 1}} & \multicolumn{1}{c}{\textbf{Bicluster2}} \\
        \midrule
        \multicolumn{1}{c}{\multirow{5}[2]{*}{\textbf{Biological process}}} & \multicolumn{1}{l}{cytoplasmic translation \textbf{(15.0\%, 2.4\%,8.39e-51)}} & \multicolumn{1}{l}{single-organism process \textbf{(55.6\%, 49.5\%, 9.74e-13)}} \\
        \multicolumn{1}{c}{} & \multicolumn{1}{l}{DNA repair \textbf{(8.8\%, 3.4\% ,7.55e-08 )}} & \multicolumn{1}{l}{cell cycle process \textbf{(11.6\%,  8.3\%, 1.17e-11)}} \\
        \multicolumn{1}{c}{} & \multicolumn{1}{l}{organic substance biosynthetic process } & \multicolumn{1}{l}{single-organism cellular process \textbf{(49.7\%, 44.0\%,}} \\
        \multicolumn{1}{c}{} & \multicolumn{1}{l}{(\textbf{40.7\%, 29.4\%, 2.98e-07)}} & \multicolumn{1}{l}{ \textbf{5.88e-11 )}} \\
        \multicolumn{1}{c}{} & \multicolumn{1}{l}{} & \multicolumn{1}{l}{} \\
        \bottomrule
        \multicolumn{1}{c}{\multirow{5}[2]{*}{\textbf{Molecular function}}} & \multicolumn{1}{l}{structural constituent of ribosome \textbf{(14.4\%,}} & \multicolumn{1}{l}{structural molecule activity \textbf{(6.8\%, 4.8\%, 8.88e-07)}} \\
        \multicolumn{1}{c}{} & \multicolumn{1}{l}{ \textbf{3.1\% , 3.80e-35)}} & \multicolumn{1}{l}{} \\
        \multicolumn{1}{c}{} & \multicolumn{1}{l}{structural molecule activity  (1\textbf{6.5\%, 4.8\%,} } & \multicolumn{1}{l}{structural constituent of cytoskeleton \textbf{(0.8\%,}} \\
        \multicolumn{1}{c}{} & \multicolumn{1}{l}{\textbf{2.25e-28)}} & \multicolumn{1}{l}{ \textbf{0.3\% 0.00984)}} \\
        \multicolumn{1}{c}{} & \multicolumn{1}{l}{} & \multicolumn{1}{l}{} \\
        \bottomrule
        \multicolumn{1}{c}{\multirow{7}[2]{*}{\textbf{Cellular component}}} & \multicolumn{1}{l}{cytosolic ribosome \textbf{(15.2\%, 2.4\% , 1.90e-51)}} & \multicolumn{1}{l}{non-membrane-bounded organelle \textbf{(22.9\%, 18.3\%,}} \\
        \multicolumn{1}{c}{} & \multicolumn{1}{l}{} & \multicolumn{1}{l}{ \textbf{5.34e-12)}} \\
        \multicolumn{1}{c}{} & \multicolumn{1}{l}{cytosolic part \textbf{(15.7\%, 3.2\%, 4.53e-41)}} & \multicolumn{1}{l}{intracellular non-membrane-bounded organelle } \\
        \multicolumn{1}{c}{} & \multicolumn{1}{l}{} & \multicolumn{1}{l}{\textbf{(22.9\%, 18.3\%, 5.34e-12)}} \\
        \multicolumn{1}{c}{} & \multicolumn{1}{l}{cytosolic small ribosomal subunit (\textbf{7.3\%, 0.9\%,}} & \multicolumn{1}{l}{organelle \textbf{(65.4\%, 60.4\%, 2.60e-09)}} \\
        \multicolumn{1}{c}{} & \multicolumn{1}{l}{ \textbf{3.43e-30)}} & \multicolumn{1}{l}{} \\
        \multicolumn{1}{c}{} & \multicolumn{1}{l}{} & \multicolumn{1}{l}{} \\
        \bottomrule
        \end{tabular}%
        \caption{Significant GO terms (process, function, component) for two biclusters on Yeast Cell-Cycle data extracted by \textit{BiARM}.}
      \label{yeastBiARM}%
   \end{table*}%
\end{landscape}                            

For the BiFCA algorithm, we show in Table \ref{goyeastBiFCA} the result of a selected set of genes for the biological process, the molecular function and the cellular component. We report the most significant GO terms. With the first bicluster (Table \ref{goyeastBiFCA}), the genes {\footnotesize (YBL027W, YBL072C, YBL087C, YBL092W, YBR031W, YBR048W, YBR079C, YBR084C-A, YBR181C, YBR191W, YCR031C, YDL061C, YDL075W, YDL081C, YDL082W, YDL083C, YDL130W, YDL136W, YDL191W, YDL229W, YDR012W, YDR025W, YDR064W, YDR382W, YDR418W, YDR447C, YDR450W, YDR471W, YDR500C, YER074W, YER102W, YER117W, YER131W, YGR214W, YHL001W, YHR141C, YIL069C, YJL177W, YJL189W, YJL190C, YJR094W-A, YJR123W, YJR145C, YKL056C, YKL156W, YKL180W, YKR057W, YKR094C, YLL045C, YLR048W, YLR075W, YLR167W, YLR185W, YLR325C, YLR340W, YLR344W, YLR367W, YLR388W, YLR406C, YLR441C, YLR448W, YML024W, YML026C, YML063W, YML073C, YMR121C, YMR143W, YMR146C, YMR194W, YMR230W, YMR242C, YNL067W, YNL096C, YNL162W, YNL301C, YNL302C, YOL039W, YOL040C, YOL127W, YOR167C, YOR234C, YOR293W, YOR312C, YOR369C, YPL081W, YPL090C, YPL143W, YPL198W, YPR043W, YPR102C, YPR163C)} concern the \textit{cytoplasmic translation} \footnote{Available at \url{http://www.yeastgenome.org/go/GO:0002181/overview }} with a \textit{p-value} of 1.08$e^{-51}$(Highly significant). 
\begin{landscape} 
  \begin{table*}[htbp]
     \centering
     
     \begin{tabular}{rrr}
        \toprule
              & \multicolumn{1}{c}{\textbf{Bicluster 1}} & \multicolumn{1}{c}{\textbf{Bicluster 2}} \\
        \midrule
        \multicolumn{1}{c}{\multirow{3}[2]{*}{\textbf{Biological process}}} & \multicolumn{1}{l}{cytoplasmic translation  \textbf{(\textit{p-value} 1.08e-51)}} & \multicolumn{1}{l}{Translation \textbf{(\textit{p-value} 4.35e-11)}} \\
          \multicolumn{1}{l}{ } \\
        \multicolumn{1}{c}{} & \multicolumn{1}{l}{ribosomal small subunit biogenesis \textbf{(\textit{p-value} 6.41e-13)}} & \multicolumn{1}{l}{cell cycle process \textbf{(\textit{p-value} 1.17e-11)}} \\
        \multicolumn{1}{c}{} & \multicolumn{1}{l}{ } & \multicolumn{1}{l}{} \\
      
        \bottomrule
        \multicolumn{1}{c}{\multirow{4}[2]{*}{\textbf{Molecular function}}} & \multicolumn{1}{l}{structural constituent of ribosome \textbf{(\textit{p-value} 1.16e-40)}} & \multicolumn{1}{l}{structural molecule activity \textbf{(\textit{p-value} 1.10e-35)}} \\
   \multicolumn{1}{l}{ } \\
        \multicolumn{1}{c}{} & \multicolumn{1}{l}{structural molecule activity  \textbf{(\textit{p-value} 2.25e-32)} } & \multicolumn{1}{l}{kinase regulator activity \textbf{(p-value 9.91e-05)}} \\

        \multicolumn{1}{l}{} \\
        \bottomrule
        \multicolumn{1}{c}{\multirow{3}[2]{*}{\textbf{Cellular component}}} & \multicolumn{1}{l}{cytosolic ribosome \textbf{(p-value 1.22e-54)}} & \multicolumn{1}{l}{cytosolic part \textbf{(p-value 2.37e-40)}} \\
          \multicolumn{1}{l}{ } \\
        \multicolumn{1}{c}{} & \multicolumn{1}{l}{ribosome \textbf{(\textit{p-value} 2.13e-36)}} & \multicolumn{1}{l}{ribosomal subunit \textbf{(p-value 1.69e-38)} } \\
        \multicolumn{1}{l}{} \\
     
        \bottomrule
        \end{tabular}%
        \caption{Significant GO terms (process, function, component) for two biclusters, extracted from Yeast Cell-Cycle data using BiFCA.}
      \label{goyeastBiFCA}%
   \end{table*}

\end{landscape} 
 \subsection{Run time performances}
 \label{time}
 Table \ref{executiontime} presents the comparison of the run time (in seconds) of our algorithms versus those respectively obtained by Trimax, BicFinder and BiMine. We note that for the Human B-cell Lymphoma and Saccharomyces Cerevisiae datasets, \textit{BiFCA+} is the fastest, while for Yeast Cell Cycle, Trimax outperforms other algorithms. In addition, BiMine is the costlier in execution time.
 
          \begin{table*}[htbp]
             \centering
           
             \resizebox{1.0 \linewidth}{!}{  \begin{tabular}{rr|r|r}
               \toprule
                     & \multicolumn{3}{c}{\textbf{Execution time (secondes)}} \\
               \midrule
               \multicolumn{1}{l|}{\textbf{Algorithms}} & \textit{\textbf{Yeast Cell Cycle}} & \textit{\textbf{Saccharomyces Cerevisiae}} & \textit{\textbf{Human B-cell Lymphoma}} \\
                \midrule
               \multicolumn{1}{l|}{BiMine \cite{Ayadi2009}} & 2 days & 5 days & 6 days \\
               \multicolumn{1}{l|}{BicFinder \cite{Ayadi2012}} & 300   & 29040 & 4680 \\
               \multicolumn{1}{l|}{Trimax \cite{Kaytoue2014}} & \textbf{1.33}  &  250.83   & 63.96 \\
                \midrule
                \bottomrule
                \multicolumn{1}{l|}{\textit{BiARM}} &  99 & 280.02 & 279 \\
                \multicolumn{1}{l|}{\textit{BiFCA+}} & 3.7   & \textbf{180.53} & \textbf{8.10} \\
                \multicolumn{1}{l|}{\textit{BiFCA}} &   86 & 332.24  & 200 \\
               \bottomrule
               \end{tabular}}
            \caption{Execution time of \textit{BiFCA+}, Trimax, BicFinder and BiMine algorithms.}
             \label{executiontime}
           \end{table*}%

\section{Conclusion}
Throughout this chapter, we have presented the BiARM algorithm, which allows extracting biclusters of positive correlations using ARM. After that, we have introduced the BiFCA+ and BiFCA algorithms which deal with FCA. This chapter has been concluded by presenting the results of applying our algorithms on three well-known real-life datasets and the evaluation and comparison of biclustering algorithms through \textit{ statistical criteria} as well as a \textit{biological criterion}. In the next chapter, we focus on the proposition of other approaches, allowing the extraction of negatively correlated biclusters.

\chapter{Identifying biclusters of negative correlations}
\minitoc
\section{Introduction}
In this chapter, we put the focus on biclustering gene expresssion data based on negative correlations. The second section (Section \ref{chapterNBic-ARM}) is devoted to the description of the NBic-ARM algorithm. We continue the third section (Section \ref{chapterNBF}) to describe the NBF algorithm. We evaluate our proposed biclustering algorithms in Section \ref{chapterexpnegative} 
\section{NBic-ARM: Identifying negative correlations using ARM}
\label{chapterNBic-ARM}
A majority of existing biclustering algorithms for microarray data focuses only on extracting biclusters with positive correlations of genes. Nevertheless, biological studies show that a group of biologically significant genes may exhibit negative correlations. In this section, we propose a new biclustering algorithm, called \textbf{NBic-ARM} (\textbf{N}egative \textbf{Bic}lusters using \textbf{A}ssociation \textbf{R}ule \textbf{M}ining). Based on generic association rules, our algorithm identifies negatively-correlated genes.
\subsection{NBic-ARM algorithm}
In this section, we introduce \textit{NBic-ARM} \cite{Houari2017}, a negative biclustering algorithm for gene expression data based on generic association rules. The \textit{NBic-ARM} method has four main phases: (1) preprocessing of a gene expression data matrix, (2) extracting biclusters of positive correlations using generic association rules, (3) extracting negatively-correlated genes, and (4) extracting maximal negatively-correlated genes.

 The pseudo-code description of \textit{NBic-ARM} is presented in Algorithm \ref{algoNBic-ARM}.
The  phases of \textit{NBic-ARM} are thoroughly described in the following subsections.
\begin{algorithm}[H] 
\caption{\textsc{NBic-ARM} Algorithm} \label{algoNBic-ARM}
\begin{algorithmic}[1]
\STATE Input: A gene expression matrix $M_{1}$, \textit{ $\mathcal{\alpha}$1},\textit{ $\mathcal{\alpha}$2}, \textit{minsupp} and \textit{minconf};\\
\STATE Output: The set of biclusters $\mathcal{\beta}$;\\
\STATE \textbf{Begin}
\STATE $\mathcal{\beta}$ := $\emptyset$ ;\\
/* \textbf{Phase 1} */\\
\STATE
 Discretize $M_{1}$ using Equation \ref{discr1bifca+} to obtain $M_{2}$ \label{step1.1} \ ;\\
\STATE  Discretize $M_{2}$ using Equation \ref{discr2Nbic-arm} to obtain $M_{3}^{+}$ \label{step1.2} \ ;\\
 \STATE Discretize $M_{2}$ using Equation \ref{discr3Nbic-arm} to obtain $M_{3}^{-}$ \label{step1.3} \ ;\\

/* \textbf{Phase 2} */\\

\STATE Extract $ \mathcal IGB^{+}$ the set of all association rules from $ M_{3}^{+}$\label{step2.1}; // {\footnotesize Candidate biclusters of positive correlations};\\
\STATE Extract $ \mathcal IGB^{-}$ the set of all association rules from $ M_{3}^{-}$\label{step2.2};  // {\footnotesize Candidate biclusters of positive correlations} ;\\

/* \textbf{Phase 3} */\\
\STATE Remove redundant conditions' candidate biclusters from $ \mathcal IGB^{+}$ and $ \mathcal IGB^{-}$.
\STATE  Extract the genes for each obtained generic AR from $M_{3}^{+}$ and from $M_{3}^{-}$ separately
     
/* \textbf{Phase 4} */\\
 \STATE \textbf{for} each two biclusters B$_{i}^{+*}$,B$_{j}^{-*}$ \textbf{do}
 \STATE \text{Compute the condition-intersection-size }\\
 \STATE \textbf{if} condition-intersection-size (B$_{i}^{+*}$,B$_{j}^{-*}$) $>$ $\alpha$1 \textbf{then}  
         \STATE \hspace{0.24cm} Bicluster-conditions= the $\bigcap $ set between conditions of B$_{i}^{+*}$ and B$_{j}^{-*}$\\ 
          \STATE \hspace{0.24cm} Bicluster-genes= the $\bigcup$ set between genes of B$_{i}^{+*}$ and B$_{j}^{-*}$\label{step6_1} 
 \STATE  $\mathcal{\beta}$\`{}= $\mathcal{\beta}$\`{}$\bigcup$ $\{$Bic (Bicluster-conditions, Bicluster-genes)$\}$ 
 \STATE  \textbf{endfor} \\
 /* \textbf{Phase 5} */\\

 \STATE \textbf{for} each two biclusters Bic$_{i}$,Bic$_{j}$ of $\mathcal{\beta}$\`{} \textbf{do}
  \STATE \text{Compute the gene-intersection-size }\\
  \STATE \textbf{if} gene-intersection-size (Bic$_{i}$,Bic$_{j}$) $>$ $\alpha$2 \textbf{then}  
          \STATE \hspace{0.24cm}  Bicluster-conditions= the $\bigcup$ set between conditions of Bic$_{i}$ and Bic$_{j}$\\ 
           \STATE \hspace{0.24cm}   Bicluster-genes= the $\bigcap $ set between genes of Bic$_{i}$ and Bic$_{j}$\label{step6_2}\\ 
           \STATE \hspace{0.24cm}    $C urrent-bicluster$= (Bicluster-conditions, \text{ }Bicluster-genes)\\
           \STATE \hspace{0.24cm}       $\mathcal{\beta}$= $\mathcal{\beta}$$\bigcup$ $\{$$Current-bicluster$$\}$\\
         \STATE \textbf{else}
                  \STATE \hspace{0.24cm} $\beta$ = $\beta$ $\bigcup$ $\{Bic_{i}  \text { }and \text { }Bic_{j}\}$\label{step6_3}              
  \STATE  \textbf{endfor} \\  
\STATE \textbf{Return }\textbf{\textbf{$\mathcal{\beta}$}\;}\; \label{output1}\\
\STATE\textbf{End}
\end{algorithmic}
\end{algorithm}
\subsubsection{Phase 1: Preprocessing of gene expression data matrix}
First of all, our method applies a preprocessing phase to transform the original data matrix $M_{1}$ into two binary ones. This phase is divided into two steps:
\begin{enumerate}
 \item The first one leads to the transformation of the original data matrix $M_{1}$ into a -101 data matrix. This step aims to highlight the trajectory pattern of genes. In microarray data analysis, genes are added into a bicluster if their trajectory patterns of expression levels are similar across a set of conditions. Thus, the obtained matrix $M_{2}$ provides useful information for the mining of relevant biclusters.

 Formally, matrix $M_{2}$ (behavior data matrix) is defined using Equation \ref{discr1bifca+}
                     
 \item Second, from the previously obtained matrix ($M_{2}$), we build two binary data matrices in order to extract association rules. In fact, the two matrices ($M_{3}^{+}$ and $M_{3}^{-}$) allow us to extract positive biclusters. Negative ones can later be extracted by location members of $M_{3}^{+}$ and $M_{3}^{-}$ which are opposite to each other.

  Formally, we define the binary matrices as follows$:$
  \begin{multicols}{2}
  \begin{align}
                                               \label{discr2Nbic-arm}
                                               M_{3}^{+}=
                                              \begin{cases}
                                              	1  & \text{if   } \text{ } \text{ }       M_{2}[i,j]=1   \\
                                                 	0  & \text{ otherwise    }    
                                              \end{cases}
 \end{align}

  \begin{align}
                                              \label{discr3Nbic-arm}
                                              M_{3}^{-}=
                                             \begin{cases}
                                             	1  & \text{if   } \text{ } \text{ }       M_{2}[i,j]=-1   \\
                                                	0  & \text{ otherwise    }    
                                             \end{cases}
                                                                                                                                                                                                                                                                        \end{align}
                                                                                                                                                                                                        
  \end{multicols} 
  with: $i$ $\in$ $\mathcal [1 \ldots p];$ $l$ $\in$ $\mathcal [1 \ldots k-1]$   
 \end{enumerate}
\subsubsection{Phase 2: Extracting biclusters}
\label{extract_biclusters}
Association Rule Mining (ARM) can be viewed as a kind of biclustering for binary data. It provides extracting patterns (\textit{biclusters}) from a binary context. To perform this task they divide the problem into two sub-problems: (1) Finding all association rules that represent biclusters' samples/genes. In fact, they consider items of both the premise and conclusion of an association rule. (2) Extracting the supporting transactions of these items.

In this work, we use also the $\mathcal{IGB}$ representation of the set of valid ARs.

In this respect, after preparing the binary data matrices $M_{3}^{+}$ and $ M_{3}^{-}$, we move to extract association rules (biclusters' conditions) from them. This task is divided into two sub-problems:

\begin{enumerate}
\item Finding all the generic ARs that represent bicluster's conditions (From $M_{3}^{+}$ and $ M_{3}^{-}$).
\item Extracting the genes for each obtained generic AR from $M_{3}^{+}$ and from $M_{3}^{-}$ separately.\\ 
\end{enumerate}

This phase (biclusters extraction) outputs biclusters of positive correlations extracted from $M_{3}^{+}$ and $ M_{3}^{-}$, which will be of use in the next phase (extracting biclusters of negative correlations).

\subsubsection{Phase 3: Extracting negatively-correlated biclusters}
\label{extracting_negative_correlated_genes}
Given the biclusters obtained from the previous phase (Section \ref{extract_biclusters}), and in order to get negatively-correlated genes, for each bicluster obtained from $M_{3}^{+}$ we do the intersection of its conditions with the  biclusters' conditions obtained from $M_{3}^{-}$. After that, we keep only the biclusters that have conditions' numbers greater or equal to  $\alpha1$. In fact,  $\alpha1$ represents the proportion of similarity between biclusters in terms of conditions obtained from generic ARs.

\subsubsection{Phase 4: Extracting maximal biclusters with negative correlations}
Given the negative biclusters obtained from the previous phase (Section \ref {extracting_negative_correlated_genes}), we want to extract maximal biclusters of negative correlations that have maximal sets of genes and conditions. A bicluster is maximal if neither a gene nor a condition can be added without violating the negative correlation criteria.

For each bicluster, we do the intersection of its genes with those of the other biclusters. After that, we keep maximal biclusters that have genes' numbers greater or equal to $\alpha2$ . In fact, $\alpha2$ represents the proportion of similarity between the biclusters obtained from the previous phase in terms of genes. Otherwise, we leave the biclusters as they are.

\subsection{Illustrative example}
Let us consider the data matrix given by Table \ref{exemple}. Each column represents all the gene expression levels from a single experiment, and each row represents the expression of a gene across all experiments. The extraction of the negatively-correlated genes from this matrix goes as follows:
 \begin{itemize}
 \item \textbf{Phase 1:} The pre-processing phase goes as follows:
 \begin{enumerate}
  \item First, we transform the numerical data (Table \ref{exemple}) into the -101 data matrix (Table \ref{exemple1}). This is done using Equation \ref{discr1bifca+}.
  \item Second, we create the binary data matrices, using the -101 data matrix. Let us consider the -101 data matrix given by Table \ref{exemple1}. To build the binary data matrices $M_{3}^{+}$ and $M_{3}^{-}$ (Tables \ref{binarypos1} and \ref{binaryneg1}), we use Equations \ref{discr2Nbic-arm} and \ref{discr3Nbic-arm}.
 \end{enumerate} 
 
\begin{center} \begin{table}[H]
\centering

 \scalebox{1.0}{                                                                                                         \begin{tabularx}{\linewidth}{|X|X|X|X|X|X|X|X|}
\hline       &$c_{1}$  & $c_{2}$ &$c_{3}$  &$c_{4}$  &$c_{5}$  &$c_{6}$ & $c_{7}$ \\ 
\hline  $g_{1}$ & 10   & 20   & 8    & 12   & 9    & 16  & 10    \\ 
\hline  $g_{2}$ & 5    & 10   & 6    & 14   & 8    & 18  & 9      \\ 
\hline  $g_{3}$ & 2    & 2    & 2    &  2   &  2   & 2   & 2       \\ 
\hline  $g_{4}$ & 20   & 10   & 14   & 9    & 16   & 10  & 13       \\ 
\hline  $g_{5}$ & 10   & 5    & 8    & 5    & 10   & 9   & 11        \\ 
 
\hline 
\end{tabularx}}
\captionof{table}{Example of gene expression matrix ($M_{1}$).}  
\label{exemple}
\end{table}
\end{center}   

\begin{center} \begin{table}[H]
\centering
 
 \scalebox{1.0}{                                                                                                         \begin{tabularx}{\linewidth}{|X|X|X|X|X|X|X|}
\hline       &$C_{1}$  & $C_{2}$ &$C_{3}$  &$C_{4}$  &$C_{5}$  &$C_{6}$ \\ 
\hline  $g_{1}$ & 1    & -1   & 1    & -1   & 1    & -1  \\ 
\hline  $g_{2}$ & 1    & -1   & 1    & -1   & 1    & -1  \\ 
\hline  $g_{3}$ & 0    & 0    & 0    &  0   & 0    & 0   \\ 
\hline  $g_{4}$ & -1   & 1    & -1   & 1    & -1   & 1  \\ 
\hline  $g_{5}$ & -1   & 1    & -1   & 1    & -1   & 1    \\ 
 
\hline 
\end{tabularx}}
\captionof{table}{-101 data matrix ($M_{2}$).} 
\label{exemple1}
\end{table}
\end{center}

\begin{center} \begin{table}[H]
\centering
 
 \scalebox{1.0}{                                                                                                         \begin{tabularx}{\linewidth}{|X|X|X|X|X|X|X|}
\hline       &$C_{1}$  & $C_{2}$ &$C_{3}$  &$C_{4}$  &$C_{5}$  &$C_{6}$ \\ 
\hline  $g_{1}$ & 1    & 0    & 1    & 0    & 1    & 0  \\ 
\hline  $g_{2}$ & 1    & 0    & 1    & 0    & 1    & 0  \\ 
\hline  $g_{3}$ & 0    & 0    & 0    & 0    & 0    & 0   \\ 
\hline  $g_{4}$ & 0    & 1    & 0    & 1    & 0    & 1  \\ 
\hline  $g_{5}$ & 0    & 1    & 0    & 1    & 0    & 1    \\ 
 
\hline 
\end{tabularx}}
\captionof{table}{Binary data matrix ($M_{3}^{+}$).} 
\label{binarypos1}
\end{table}
\end{center}

\begin{center} \begin{table}[H]
\centering

 \scalebox{1.0}{                                                                                                         \begin{tabularx}{\linewidth}{|X|X|X|X|X|X|X|}
\hline      &$C_{1}$  & $C_{2}$ &$C_{3}$  &$C_{4}$  &$C_{5}$  &$C_{6}$\\ 
\hline  $g_{1}$ & 0    & 1    & 0    & 1    & 0   & 1  \\ 
\hline  $g_{2}$ & 0    & 1    & 0    & 1    & 0   & 1  \\ 
\hline  $g_{3}$ & 0    & 0    & 0    & 0    & 0   & 0   \\ 
\hline  $g_{4}$ & 1    & 0    & 1    & 0    & 1   & 0  \\ 
\hline  $g_{5}$ & 1    & 0    & 1    & 0    & 1   & 0    \\
 
\hline 
\end{tabularx}}
\captionof{table}{Binary data matrix ($M_{3}^{-}$).}  
\label{binaryneg1}
\end{table}
\end{center}

  \begin{minipage}{.5\textwidth}\centering

   \scalebox{1.2}{
   \begin{tabular}{|l|c|}
     \hline  Transactions & Items \\ 
     \hline $1$ & 1 3 5       \\ 
     \hline $2$ & 1 3 5         \\ 
     \hline $3$ &  \_       \\ 
     \hline $4$ & 2 4 6    \\ 
     \hline $5$ & 2 4 6       \\  
     \hline 
    \end{tabular}}
   \captionof{table}{Transactional representation of binary data set given in Table \ref{binarypos1}.}
    \label{base de transactionpos}
  \end{minipage}
  \begin{minipage}{.5\textwidth}\centering
  
    \scalebox{1.2}{
     \begin{tabular}{|l|c|}
      \hline  Transactions & Items \\ 
      \hline $1$ & 2 4 6           \\ 
      \hline $2$ & 2 4 6           \\
      \hline $3$ &  \_             \\  
      \hline $4$ & 1 3 5           \\ 
      \hline $5$ & 1 3 5           \\  
      \hline 
     \end{tabular}}
  \captionof{table}{Transactional representation of binary data set given in Table \ref{binaryneg1}.}
  \label{base de transactionneg}
  \end{minipage}

\vspace{+0.5 cm} 
\item \textbf{Phase 2:} After preparing the binary data matrices, we move to extract the $\mathcal{IGB}$'s generic rules, i.e. the biclusters' conditions, from the matrices $M_{3}^{+}$ and $M_{3}^{-}$.
 By using the previous example, we obtain as a result the $\mathcal{IGB}$' basis of association rules presented in Table \ref{IGBbasis}. Thus, the obtained biclusters are:\\
$B1$$^+$=$<g_{1}g_{2}, C_{1}C_{3}C_{5}>$;\\    $B2$$^+$=$<g_{4}g_{5}, C_{2}C_{4}C_{6}>$; \\ $B1$$^-$=$<g_{4}g_{5}, C_{1}C_{3}C_{5}>$; \\ $B2$$^-$=$<g_{1}g_{2}, C_{2}C_{4}C_{6}>$. \\
\item \textbf{Phase 3:} Let $\alpha 1$ =90\%. Using these biclusters, we compute the condition-intersection between $B$$^+$ and $B$$^-$. We obtain as a result : $B1$\`{}=$<g_{1}g_{2}g_{4}g_{5}, C_{1}C_{3}C_{5}> $ and $B2$\`{}=$<g_{1}g_{2}g_{4}g_{5}, C_{2}C_{4}C_{6}>$.

\item \textbf{Phase 4:} $B1$\`{} and $B2$\`{} have negative correlations, but they are not maximal. Let $\alpha 2$ =90\%. The obtained bicluster will, thus, be: Bic=$<g_{1}g_{1}g_{4}g_{5}, C_{1}C_{2}C_{3}C_{4}C_{5}C_{6}>$ the maximal bicluster of negative correlations.
\end{itemize}

The profile of this bicluster is sketched in Figure \ref{example_courbe}. From this figure, negative correlation patterns can be observed. In particular, two positive correlated genes (g$_{1}$, g$_{2}$) are plotted and two other genes (g$_{4}$, g$_{5}$) show negative correlations with the two aforementioned genes.
\begin{center}
\fbox{ \scalebox{1.0}{\includegraphics{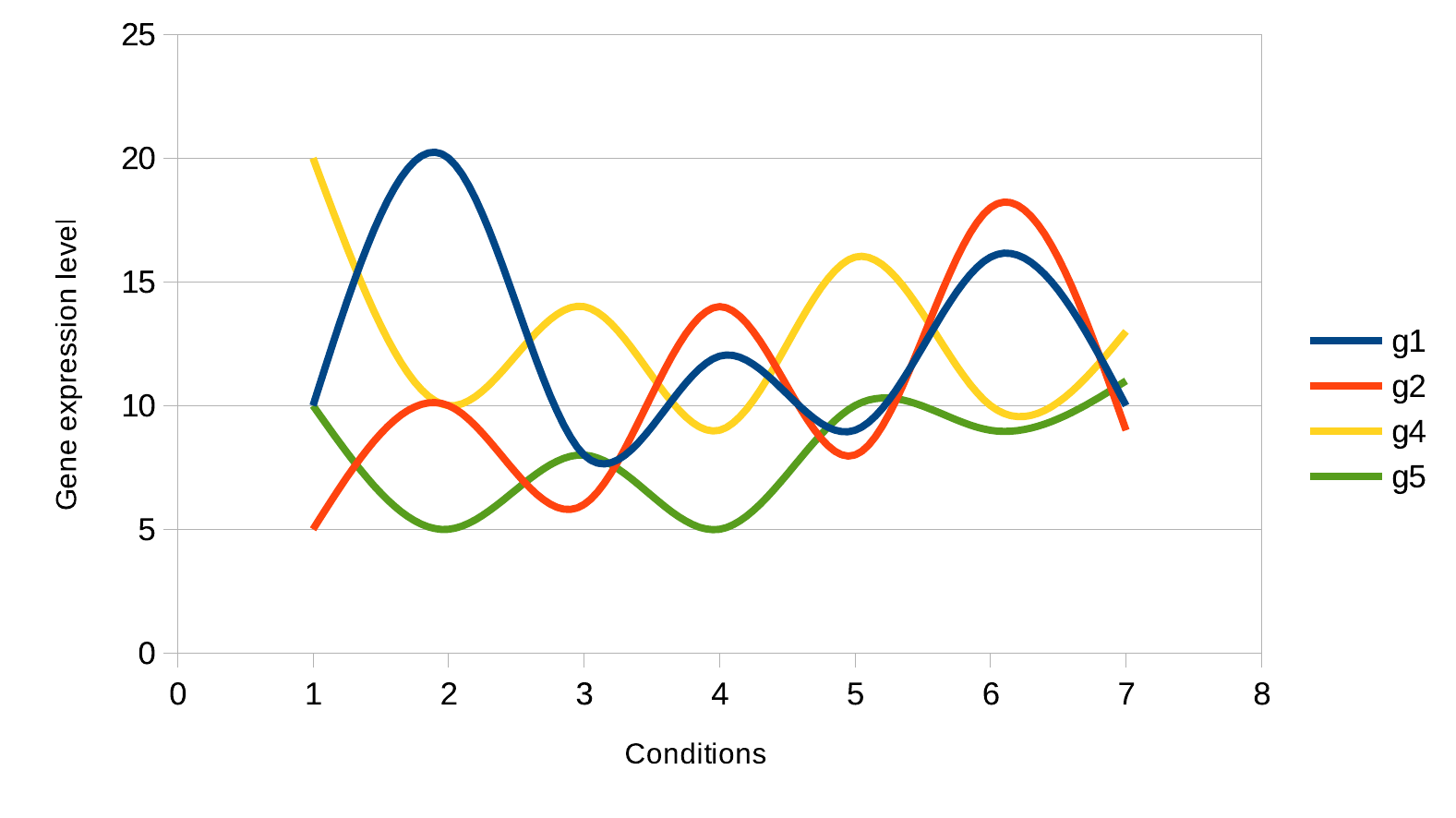}}}
\captionof{figure}{Resulting bicluster profile obtained through NBic-ARM algorithm on our example.}
\label{example_courbe}

\end{center}
 
   \begin{tabular}{|l|r|r||l|r|r|} \toprule \multicolumn{6}{|c|}{\textbf{Generic association rules (ARs)}} \\ \midrule \multicolumn{3}{|c||}{\textbf{$ M_{3}^{+}$}} & \multicolumn{3}{|c|}{\textbf{$ M_{3}^{+}$}} \\  \midrule ARs & \multicolumn{1}{|c|}{Support} & \multicolumn{1}{|c||}{Confidence} & ARs & \multicolumn{1}{|c|}{Support} & \multicolumn{1}{|c|}{Confidence} \\   \midrule AR 1$^+$: 1 $\Longrightarrow$ 3, 5 & 0.4 & 1& AR 1$^-$: 1 $\Longrightarrow$ 3, 5 & 0.4 & 1 \\ AR 2$^+$: 3 $\Longrightarrow$ 1, 5& 0.4 & 1 & AR 2$^-$: 3 $\Longrightarrow$ 1, 5& 0.4 & 1 \\ AR 3$^+$: 5 $\Longrightarrow$ 1, 3& 0.4 & 1& AR 3$^-$: 5 $\Longrightarrow$ 1, 3& 0.4 & 1 \\AR 4$^+$: 2 $\Longrightarrow$ 4, 6& 0.4 & 1& AR 4$^-$: 2 $\Longrightarrow$ 4, 6& 0.4 & 1 \\AR 5$^+$: 4 $\Longrightarrow$ 2, 6& 0.4 & 1& AR 5$^-$: 4 $\Longrightarrow$ 2, 6& 0.4 & 1 \\AR 6$^+$: 6 $\Longrightarrow$ 2, 4& 0.4 & 1& AR 6$^-$: 6 $\Longrightarrow$ 2, 4& 0.4 & 1 \\ \bottomrule 
   \end{tabular}
   \captionof{table}{\textit{IGB} basis of association rules extracted from Table \ref{base de transactionpos} and Table \ref{base de transactionneg} ($minsupp=0.2, minconf=0.9$).}
\label{IGBbasis}
\subsection{Discussion} 
First, we focus on the extraction of negatively correlated biclusters where we start by proposing a new ARM-based biclustering method (\textit{\textbf{NBic-ARM}}) as a novel biclustering algorithm for discovering negative biclusters from gene expression data. Our algorithm relies on the extraction of generic association rules from the dataset by discretizing this latter into two binary data matrices. 
 However, a significant number of redundant rules is found. Therefore, and to remedy this problem, we introduce our second contribution. This latter consists in proposing a new solution based on FCA. In addition, and during the discretization phase, we use only adjacent conditions. Consequently, in the next section, we present NBF, a new FCA-based biclustering method, for discovering negatively correlated genes from gene expression data. Our approach consists in extracting formal concepts from a dataset after having discretized it into two binary data matrices with combining in pairs, for each gene, all conditions' pairs.
 
\section{NBF: Identifying negative correlations in gene expression data through FCA }
\label{chapterNBF}
We introduce in this section a new algorithm, called \textbf{\textit{NBF}} (\textbf{N}egative \textbf{B}icluster \textbf{F}inder). One of the NBF features is to discover biclusters of negative correlations using FCA. We carry out exhaustive experiments on three real-life datasets to assess the performance of NBF. Our results prove the ability of NBF to identify statistically and biologically significant biclusters.
\subsection{NBF algorithm}
In this section, we present our method \cite{Houari2018a}, called NBF, to find biclusters with negatively-correlated patterns in microarray data. The NBF method has five main phases: (1) preprocessing of gene expression data matrix, (2) extracting formal concepts, (3) filtering out the obtained formal concept set (4) extracting negatively-correlated genes, and (5) extracting maximal negatively-correlated genes. We are going to describe these phases in the following sub-sections.
\begin{algorithm}[H] \small
\caption{\textsc{NBF} Algorithm} \label{algoNBF}
\begin{algorithmic}[1]
\STATE Input: A gene expression matrix $ M_{1}$, \textit{ $\mathcal{\alpha}$1},\textit{ $\mathcal{\alpha}$2} and  \textit{minStability};\\
\STATE Output: The set of biclusters $\mathcal{\beta}$;\\
\STATE \textbf{Begin}
\STATE $\mathcal{\beta}$ := $\emptyset$ ;\\
/* \textbf{Phase 1} */\\
\STATE
 Discretize $M_{1}$ using Equation \ref{disc1biarm} to obtain $M_{2}$ \label{step1.1} \ ;\\
\STATE  Discretize $M_{2}$ using Equation \ref{discr2Nbic-arm} to obtain $M_{3}^{+}$ \label{step1.2} \ ;\\
 \STATE Discretize $M_{2}$ using Equation \ref{discr3Nbic-arm} to obtain $ M_{3}^{-}$ \label{step1.3} \ ;\\

/* \textbf{Phase 2} */\\

\STATE Extract $ \mathcal FC^{+}$ the set of all formal concepts from $M_{3}^{+}$\label{step2.1}; // {\footnotesize Candidate biclusters of positive correlations};\\
\STATE Extract $ \mathcal FC^{-}$ the set of all formal concepts from $ M_{3}^{-}$\label{step2.2};  // {\footnotesize Candidate biclusters of positive correlations} ;\\

/* \textbf{Phase 3} */\\
\STATE
 Compute $\varepsilon$ the stability of each formal concept of the set $ \mathcal FC^{+}$\ ;\\
\STATE \textbf{for} each formal concept in $ \mathcal FC^{+}$ \textbf{do} 
\STATE \textbf{if} $\varepsilon$ $>$ \textit{minstability} \textbf{then}  
        \STATE \hspace{0.24cm} $\mathcal FC^{+*}$ = $\mathcal FC^{+*}$ $\bigcup$ $\{current concept\}$ \label{step4_1}// {\footnotesize Selected biclusters of positive correlations.} 
\STATE \textbf{else}
         \STATE \hspace{0.24cm} Remove current concept\label{step4_2};
\\ \STATE  \textbf{endfor}  \\

\STATE Compute $\varepsilon$ the stability of each formal concept of the set $ \mathcal FC^{-}$\ ;\\
\STATE \textbf{for} each formal concept in $ \mathcal FC^{-}$ \textbf{do} 
\STATE \textbf{if} $\varepsilon$ $>$ \textit{minstability} \textbf{then}  
        \STATE \hspace{0.24cm} $\mathcal FC^{-*}$ = $\mathcal FC^{-*}$ $\bigcup$ $\{current concept\}$ \label{step5_1}// {\footnotesize Selected biclusters of negative correlations.} 
\STATE \textbf{else}
         \STATE \hspace{0.24cm} Remove current concept\label{step5_2};
\\ \STATE  \textbf{endfor}  \\      

/* \textbf{Phase 4} */\\
 \STATE \textbf{for} each two concepts FC$_{i}^{+*}$,FC$_{j}^{-*}$ \textbf{do}
 \STATE \text{Compute the condition-intersection-size }\\
 \STATE \textbf{if} condition-intersection-size (FC$_{i}^{+*}$,FC$_{j}^{-*}$) $>$ $\alpha$1 \textbf{then}  
         \STATE \hspace{0.24cm} bicluster-conditions= the $\bigcap $ set between conditions of FC$_{i}^{+*}$ and FC$_{j}^{-*}$\\ 
          \STATE \hspace{0.24cm} bicluster-genes= the $\bigcup$ set between genes of FC$_{i}^{+*}$ and FC$_{j}^{-*}$\label{step6_1} 
 \STATE  $\mathcal{\beta}$\`{}= $\mathcal{\beta}$\`{}$\bigcup$ $\{$Bic (bicluster-conditions,bicluster-genes)$\}$ 
 \STATE  \textbf{endfor} \\
 /* \textbf{Phase 5} */\\

 \STATE \textbf{for} each two biclusters Bic$_{i}$,Bic$_{j}$ of $\mathcal{\beta}$\`{} \textbf{do}
  \STATE \text{Compute the gene-intersection-size }\\
  \STATE \textbf{if} gene-intersection-size (Bic$_{i}$,Bic$_{j}$) $>$ $\alpha$2 \textbf{then}  
          \STATE \hspace{0.24cm}  bicluster-conditions= the $\bigcup$ set between conditions of Bic$_{i}$ and Bic$_{j}$\\ 
           \STATE \hspace{0.24cm}   bicluster-genes= the $\bigcap $ set between genes of Bic$_{i}$ and Bic$_{j}$\label{step6_2}\\ 
           \STATE \hspace{0.24cm}    $current-bicluster$= (bicluster-conditions, \text{ }bicluster-genes)\\
           \STATE \hspace{0.24cm}       $\mathcal{\beta}$= $\mathcal{\beta}$$\bigcup$ $\{$$current-bicluster$$\}$\\
         \STATE \textbf{else}
                  \STATE \hspace{0.24cm} $\beta$ = $\beta$ $\bigcup$ $\{Bic_{i}  \text { }and \text { }Bic_{j}\}$\label{step6_3}              
  \STATE  \textbf{endfor} \\  
\STATE \textbf{Return }\textbf{\textbf{$\mathcal{\beta}$}\;}\; \label{output1}\\
\STATE\textbf{End}
\end{algorithmic}
\end{algorithm}
 The pseudo-code description of \textit{NBF} is shown in Algorithm \ref{algoNBF}. To describe formally the NBF algorithm, let us define some variables:
\begin{description}
 \item $M_{1}$(resp. $M_{2}$, and $ M_{3}^{+}$ and $M_{3}^{-}$): data matrix (resp. discretized data matrix, and binary data matrices),
\item $\mathcal{\beta}$: set of maximal negatively-correlated biclusters,
\item  $\mathcal{\beta}$\`{}: set of negatively-correlated biclusters,
\item Bic: bicluters of positive correlations (formal concepts) extracted from $M_{3}^{+}$ and $M_{3}^{-}$,
\item  $\varepsilon$: stability of formal concept,
 \item  $\mathcal{\alpha}$1,  $ \mathcal{\alpha}$2, \textit{minStability}: quality thresholds according to negative correlation, maximal negative correlation and stability measure, respectively, 
 \item $ \mathcal FC^{+}$(resp.$ \mathcal FC^{-}$): set of extracted biclusters from $M_{3}^{+}$(resp.$ M_{3}^{-}$)
\end{description}

\subsubsection{\textbf{Phase 1:} Preprocessing of gene expression data matrix}

\label{descritize1}
Our method applies a preprocessing phase to transform the original data matrix $M_{1}$ into two binary ones.
First, we discretize the original data into a 3-state data matrix $M_{2}$ (behavior matrix). This latter is  discretized into two binary data matrices, positive and negative binary matrices.

In our case, the purpose of using the discretized matrix is to identify biclusters with negative correlation genes.
 
 This phase is divided into two steps:                                                                                                                          \begin{enumerate}
                                                                                                                          \item First, we discretize the initial data matrix. The discretization process generates the 3-state data matrix. 
                                                                                                                                                                                                                                                      In our case, each column of the 3-state data matrix represents the meaning of the variation in genes between a pair of conditions of $M_{1}$.
                                                                                                                                                                                                                                                      
                                                                                                                                                                                                            Formally, we can incrementally build matrix $M_{2}$ (3-state data matrix) through the merger of a couple of columns from the input data matrix $M_{1}$. Since $M_{1}$ has $n$ rows and $m$ columns, there are $m(m-1)/2$ distinct combinations between columns. So, $M_{2}$ has $n$ rows and $m(m-1)/2$ columns. $M_{2}$ is defined using Equation \ref{disc1biarm}
                                                                                                                       
\item For the second step of the preprocessing phase, we build two binary data matrices in order to extract formal concepts from $M_{3}^{+}$ and  $M_{3}^{-}$. This discretization is used in order to extract genes which show an opposite change tendency over a subset of experimental conditions.

Formally, we define the binary matrices using Equations \ref{discr2Nbic-arm} and \ref{discr3Nbic-arm}.
                                                                                                                                                                                                                                           \end{enumerate}

\subsubsection{\textbf{Phase 2:} Extracting formal concepts}
\label{extractingconcepts}


The presence of local patterns in biological data motivated the wide study of biclustering in dealing with them using pattern-mining-based searches. Given our promising results following the use of FCA in extracting positively correlated biculsters, we also adopt it in dealing with the extraction of negatively correlated ones.

In this context, the second phase after the preprocessing one is the phase of extracting formal concepts.  
As mentioned before, the extraction of formal concepts is carried out through the LCM algorithm.

 \subsubsection{\textbf{Phase 3:} Coherency measure (Stability)}
 \label{stabilitymeasure}
 The huge number of extracted formal concepts  (candidate biclusters) represent a genuine hindrance for their effective use. To remove the non-coherent formal concepts from such endless list, the stability metric is the best option. Stability was introduced for the first time by Kuznestov \cite{Kuznetsov1990} and revisited in \cite{Kuznetsov2007a,Kuznetsov2007}.
 It is the most widely used around the FCA community. The intentional stability measure for a given formal concept highlights the proportion of subsets of its objects whose closure is equal to the intent of this formal concept. This measure captures the dependence between the intent on particular objects of the extent. In our work, we adopt the stability measure in order to measure the coherency of a given formal concept. Finally, we consider only the obtained concepts for which the stability measure is verified.

 Worthy of mention, computing the stability measure is an NP-complete task. Hence, in our experiments, we use the DFSP algorithm \cite{Dimassi2014} to compute the stability of the obtained formal concepts. The latter is considered a unique algorithm that efficiently and straightforwardly handles a set of formal concepts for stability computation.

  \subsubsection{\textbf{Phase 4:} Extracting negatively-correlated genes}
   \label{extract_negative_correlated_genes}
   It is of paramount importance to extract negatively-correlated biclusters since most of the existing biclustering algorithms identify only positively-correlated genes despite the fact that recent biological studies have turned to a trend focusing on the notion of negative correlations.

   In this section, we present the problem of extracting biclusters of negative correlations using FCA. We should not take into account all possible coherent formal concepts in $M_{3}^{+}$ and $M_{3}^{-}$, but rather coherent formal concepts having an intersection size greater or equal to a given intersection threshold $\alpha 1$. Intuitively, we consider the formal concepts presented in the previous sub-section (Sub-section \ref{stabilitymeasure}) to get negatively-correlated genes for a given\textbf{ $\alpha 1$ }only where we compute the proportion of similarity between coherent formal concepts from $M_{3}^{+}$ and $ M_{3}^{-}$ in terms of conditions. In other words, coherent formal concepts with an intersection size above or equal to the threshold belong to the same bicluster, while those with an intersection value below it do not.\\
   \subsubsection{\textbf{Phase 5:} Extracting maximal negatively-correlated genes}
    \label{extract_maximal_negative_correlated_genes}
    We consider here maximal biclusters of negative correlations, denoted by $(X,Y)$, where $X$ and $Y$ are respectively maximal sets of objects and attributes, such that the values taken by these attributes for these objects have negative correlations. A bicluster is maximal if neither an object nor an attribute can be added without violating the negative correlation criteria. We take into consideration biclusters having an intersection size greater or equal to a given intersection threshold\textbf{ $\alpha$}2. Here we compute the proportion of similarity between the biclusters obtained from the previous phase (Section \ref{extract_negative_correlated_genes}) in terms of genes in order to obtain maximal biclusters. That is to say, we compute the intersection between two biclusters. If their similarity value is above or equal to $\alpha$2, we construct a maximal bicluster. Otherwise, we leave the biclusters as they are.

 
\subsection{Illustrative example}
\begin{figure*}[t]
           \begin{center}
            \fbox{\includegraphics[width=1.0\textwidth]{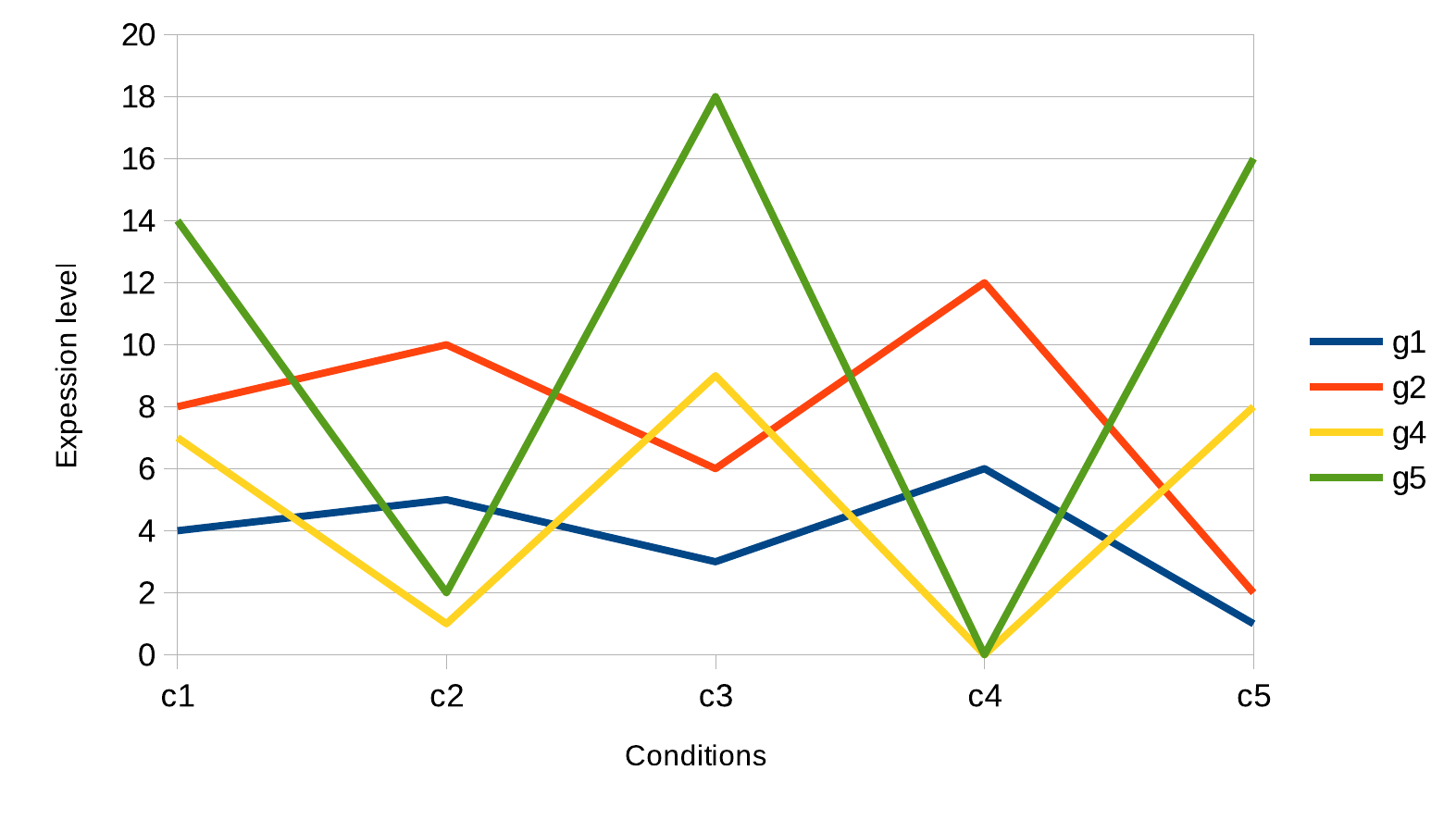}} 
           \end{center}
           \caption{Resulting bicluster profile obtained through NBF algorithm on our example.}
           \label{exampleprofile}
           \end{figure*}

Let us consider the dataset given by Table \ref{example}.
 \begin{itemize}
 \item \textbf{Phase 1:} The preprocessing phase is applied as follows:
 
\begin{center} \begin{table}[H]
\centering
 
 \scalebox{1.0}{                                                                                                         \begin{tabularx}{\linewidth}{|X|X|X|X|X|X|}
\hline  &$ c_{1}$  & $ c_{2}$ &$ c_{3}$  &$ c_{4}$  &$ c_{5}$ \\ 
 \hline  $g_{1}$& 4    & 5    & 3    & 6    & 1 \\ 
 \hline  $g_{2}$& 8    & 10   & 6    & 12   & 2 \\ 
 \hline  $g_{3}$& 3    & 3    & 3    & 3    & 3 \\ 
 \hline  $g_{4}$& 7    & 1    & 9    & 0    & 8  \\ 
 \hline  $g_{5}$& 14   & 2    & 18   & 0    & 16  \\  
 
\hline 
\end{tabularx}}
\captionof{table}{Example of gene expression matrix ($M_{1}$).} 
 \label{example}
\end{table}
\end{center}
 
By applying Equation \ref{disc1biarm}, we represent the 3-state data matrix (Table \ref{table3}).

\begin{center}
 \begin{table}[H]
 
  \centering

  \resizebox{1.0 \linewidth}{!}{
 
  \begin{tabularx}{\linewidth}{|X|X| X|X|X|X|X|X|X|X|X|}
  \hline   & C$_{1}$ & C$_{2}$ & C$_{3}$ & C$_{4}$ & C$_{5}$ & C$_{6}$ & C$_{7}$ & C$_{8}$ & C$_{9}$ & C$_{10}$ \\ 
  \hline   $g_{1}$ & 1 & -1 & 1 & -1 & -1 & 1 & -1 & 1 & -1 & -1   \\ 
   \hline   $g_{2}$& 1 & -1 & 1 & -1 & -1 & 1 & -1 & 1 & -1 & -1  \\ 
   \hline   $g_{3}$& 0 & 0 & 0 & 0 & 0 & 0 & 0 & 0 &  0 & 0 \\ 
   \hline   $g_{4}$& -1 & 1 & -1 & 1 & 1 & -1 & 1 & -1 & -1 & 1 \\ 
   \hline   $g_{5}$& -1 & 1 & -1 & 1 & 1 & -1 & 1 & -1 & -1 & 1  \\ 
   \hline 
  \end{tabularx}}
  \caption{3-state data matrix ($M_{2}$).}
 \label{table3}
 \end{table}
 \end{center}

 Let $M_{2}$ be a 3-state data matrix (Table \ref{table3}). By using Equations \ref{discr2Nbic-arm} and \ref{discr3Nbic-arm}, we obtain the binary matrices (Tables \ref{binaryneg} and \ref{binarypos})
 \begin{center}
 \begin{table}[H]
 
  \centering

  \resizebox{1.0 \linewidth}{!}{

   \begin{tabularx}{\linewidth}{|X|X| X|X|X|X|X|X|X|X|X|}
    \hline   & C$_{1}$ & C$_{2}$ & C$_{3}$ & C$_{4}$ & C$_{5}$ & C$_{6}$ & C$_{7}$ & C$_{8}$ & C$_{9}$ & C$_{10}$ \\ 
   \hline   $g_{1}$& 1 & 0 & 1 & 0 & 0  & 1& 0& 1& 0 & 0  \\ 
   \hline   $g_{2}$& 1 & 0 & 1 & 0 & 0  & 1& 0& 1& 0 & 0 \\ 
   \hline   $g_{3}$& 0 & 0 & 0 & 0 & 0  & 0& 0& 0& 0  & 0\\ 
   \hline   $g_{4}$& 0 & 1 & 0 & 1 & 1 & 0 & 1& 0& 0  & 1 \\ 
   \hline   $g_{5}$& 0 & 1 & 0 & 1 & 1  & 0& 1& 0&  0& 1 \\ 
   \hline 
  \end{tabularx}} 
  \caption{Positive binary data matrix ($ M_{3}^{+}$).}
 \label{binarypos}
 \end{table}
 \end{center}

 \begin{center}
 \begin{table}[H]

  \centering

  \resizebox{1.0 \linewidth}{!}{

   \begin{tabularx}{\linewidth}{|X|X| X|X|X|X|X|X|X|X|X|}
    \hline   & C$_{1}$ & C$_{2}$ & C$_{3}$ & C$_{4}$ & C$_{5}$ & C$_{6}$ & C$_{7}$ & C$_{8}$ & C$_{9}$ & C$_{10}$ \\ 
      \hline   $g_{1}$& 0 & 1 & 0 & 1 & 1 & 0& 1& 0 & 1 & 1 \\ 
      \hline   $g_{2}$& 0 & 1 & 0 & 1 & 1 & 0& 1& 0 & 1 & 1 \\
      \hline   $g_{3}$& 0 & 0 & 0 & 0 & 0 & 0& 0& 0 & 0 & 0\\ 
      \hline   $g_{4}$& 1 & 0 & 1 & 0 & 0 & 1& 0& 1 & 1 & 0\\ 
      \hline   $g_{5}$& 1 & 0 & 1 & 0 & 0 & 1& 0& 1 & 1 & 0\\ 
   \hline 
  \end{tabularx}}
   \caption{Negative binary data matrix ($ M_{3}^{-}$).} 
 \label{binaryneg}
 \end{table}
 \end{center}

 \item \textbf{Phase 2:} After preparing the binary data matrices, we move to extract formal concepts (Candidate biclusters) from the matrices $ M_{3}^{+}$ and $ M_{3}^{-}$ (Table \ref{binaryneg} and Table \ref{binarypos}).
 
 By using the binary data given in Table \ref{binaryneg} and Table \ref{binarypos}, we obtain as a result the formal concepts sketched in Table \ref{formal_concepts}.
 
 \begin{table*}[htbp] \centering  
 \resizebox{1.0 \linewidth }{!}{
  \begin{tabular}{|l|l|l|r||l|l|l|r|} \toprule     \multicolumn{8}{|c|}{Formal concepts (Fcs) (candidate biclusters)}\\
       \hline
       \multicolumn{4}{|c||}{\textbf{$ M_{3}^{+}$}} & \multicolumn{4}{c|}{\textbf{$ M_{3}^{-}$}}\\
       \hline
       Id concept & Extent (bicluster's genes) & Intent (bicluster's conditions) & Stability & Id concept & Extent (bicluster's genes) & Intent (bicluster's conditions) & Stability\\
       \hline
       FC1$^+$ & g$_{1}$,g$_{2}$ & C$_{1}$,C$_{3}$,C$_{6}$,C$_{8}$ & 0.75 & FC1$^-$ & g$_{4},g$$_{5}$ & C$_{1}$,C$_{3}$,C$_{6}$,C$_{8}$,C$_{9}$ & 0.75\\
       \hline
       FC2$^+$ & g$_{4}$,g$_{5}$ & C$_{2}$,C$_{4}$,C$_{5}$,C$_{7}$,C$_{10}$ & 0.75 & FC2$^-$ & g$_{1}$,g$_{2}$ & C$_{2}$,C$_{4}$,C$_{5}$,C$_{7}$,C$_{9}$,C$_{10}$ & 0.75 \\
       \hline
           &     &     &     & FC3$^-$ & g$_{1}$,g$_{2}$,g$_{4}$,g$_{5}$ & C$_{9}$  & 0.56 \\ \bottomrule 
  \end{tabular}}
  \caption{Extracted formal concepts from formal context presented in Table \ref{binaryneg} and Table \ref{binarypos}.} 
 \label{formal_concepts}%
 \end{table*}

   \item \textbf{Phase 3:} In this phase, we consider only coherent formal concepts.
    Using the example of concepts (example from Table \ref{formal_concepts}), we compute the stability measure. The results are sketched in Table \ref{formal_concepts}.

    In this example, the stability measure threshold is set equal to 0.6. Therefore, we consider $FC$ $1^{+}, FC2^{+}, FC1^{-}, FC2^{-}$ and ignore the $FC$ $3^{-}$. 
    \item \textbf{Phase 4:}
     Suppose that $\alpha 1$ =70\% and using our example, we have: \\
     $FC1^{+} \bigcap FC1^{-}= \{C_{1}, C_{3}, C_{6}, C_{8}\} $;\\
     $FC1^{+} \bigcap FC2^{-}= \emptyset $;\\
     $FC1^{+} \bigcap FC3^{-}=  \emptyset $;\\
     and\\
     $FC2^{+} \bigcap FC1^{-}= \emptyset $;\\
     $FC2^{+} \bigcap FC2^{-}= \{C_{2}, C_{4}, C_{5}, C_{7}, C_{10}\} $; \\
     $FC2^{+} \bigcap FC3^{-}= \emptyset $.

     \item \textbf{Phase 5:}
      For example, $Bic1=\langle g_{1}g_{2}g_{4}g_{5}, C_{1}C_{3}C_{6}C_{8}\rangle$ is a bicluster with negative correlation genes, but it is not maximal.
     Then for genes, we consider the union of the two genes' sets; so for the previous concepts the biclusters become: \\
      $Bic1= \langle g_{1}g_{2}g_{4}g_{5},C_{1}C_{3}C_{6}C_{8}\rangle$;\\ and \\$Bic2= \langle g1g2g4g5, C_{2}C_{4}C_{5}C_{7}C_{10}\rangle.$

      Suppose that $\alpha 2$ =70\%. Thus the obtained bicluster is: \\
      $maxbic (Bic1,Bic2)=\\(genes-intesection,conditions-union).$\\
       $maxbic(Bic1,Bic2)$$= $$\langle g_{1}g_{2}g_{4}g_{5}$$,$ $C_{1}C_{2}C_{3}C_{4}C_{5}C_{6}C_{7}C_{8}C_{10}\rangle.$

       At the end, using Equation ~\ref{disc1biarm} on page~\pageref{disc1biarm}, we obtain the bicluster $\langle g_{1} g_{2}g_{4}g_{5}$, $c_{1}c_{2}c_{3}c_{4}c_{5}$$\rangle$.
      
 \end{itemize}      
      The profile of this bicluster is sketched in Figure \ref{exampleprofile}. From this figure, negative correlation patterns can be observed. In particular, two positively-correlated genes $(g_{1}, g_{2})$  are plotted and two other genes $(g_{4}, g_{5})$ show negative correlations with the two aforementioned genes. 
\section{Experimental results}
\label{chapterexpnegative}
In this section, we provide the experimental results of using our algorithm on three well-known real-life datasets. The evaluation of the biclustering algorithms and their comparison are based on two criteria:\textit{ Statistical} and \textit{Biological}. We compare our algorithm with the state-of-the-art biclustering algorithms, the Trimax algorithm\footnote{Available at \url{https://github.com/mehdi-kaytoue/trimax}.} \cite{Kaytoue2014}  that uses FCA, and the MBA algorithm \cite{Ayadi2014} which extracts biclusters with negative correlations. 
\subsection{Experimental protocol}
\label{Expprotocolneg}
As for Chapter \ref{chapter3} (Section \ref{Expprotocolneg} on page ~\pageref{Expprotocolneg}), the first series of experiments concerns statistical validation, where we compute the coverage for the Yeast Cell-Cycle and Human B-cell Lymphoma datasets and we compute also the adjusted \textit{p-value} for the Yeast Cell-Cycle and Saccharomyces Cerevisiae datasets. Whereas, the second series of experiments is applied to the Yeast Cell-Cycle and Saccharomyces Cerevisiae datasets in order to study the biological significance of extracted biclusters.

\hspace{+ 0.5 cm}Table \ref {parametresalgoneg} summarizes the different values of the parameters used by our algorithms. The values of these parameters are chosen experimentally.

\begin{figure*}[htb]  
          
            \begin{center}
            
               \fbox{
               \scalebox{1.5}{
                 \begin{tikzpicture}
                   \begin{axis}[
                       xlabel={$\alpha$2 (\%)},
                       ylabel={Number of biclusters },
                       xmin=50, xmax=100,
                       ymin=0, ymax=10120,
                      xtick={50,55,60,65,70,75,80,85,90,95,100},
                       ytick={500,1000,2000,3000,4000,5000,6000,7000,8000,9000,10000,10120},
                       legend pos=north east,
                       ymajorgrids=true,
                       grid style=dashed,
                   ]

                   \addplot[
                               color=blue,
                               mark=triangle*,
                                ]
                  coordinates {
                 (55,10112)(60,7015)(65,4341)(70,2279)(75,1041)(80,482)(85,247)(90,133)(95,41)(100,0)
                                              
                               };           
                 
                {\footnotesize  \legend{Number of biclusters}}

                   \end{axis}
                   \end{tikzpicture}
            
                                }
                                }
                    \end{center}
                  \caption{Number of biclusters with varying $\alpha$2 values in Yeast Cell-Cycle dataset.}
                      \label{varying_alpha2}
           \end{figure*}
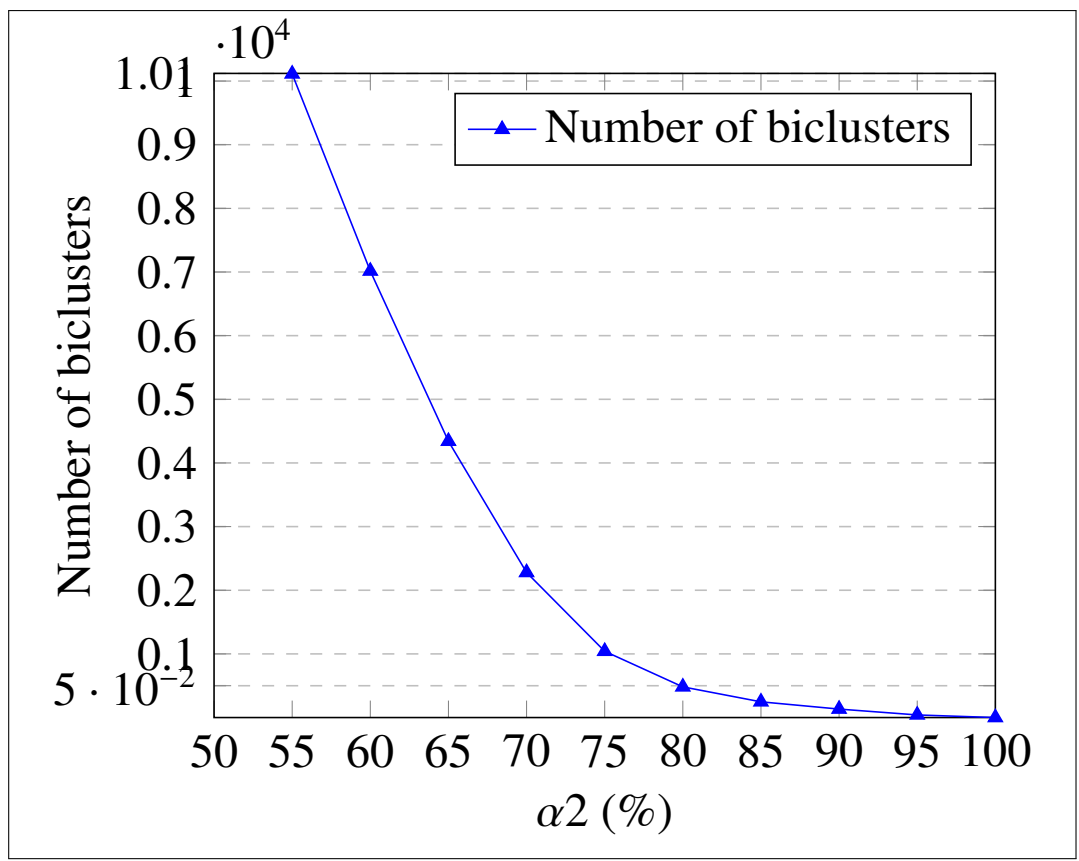

 \hspace{+ 0.5 cm} Figure \ref{varying_alpha2} depicts the impact of the \textbf{$\alpha$}2  threshold on the number of the obtained biclusters for the Yeast Cell-Cycle dataset.
  
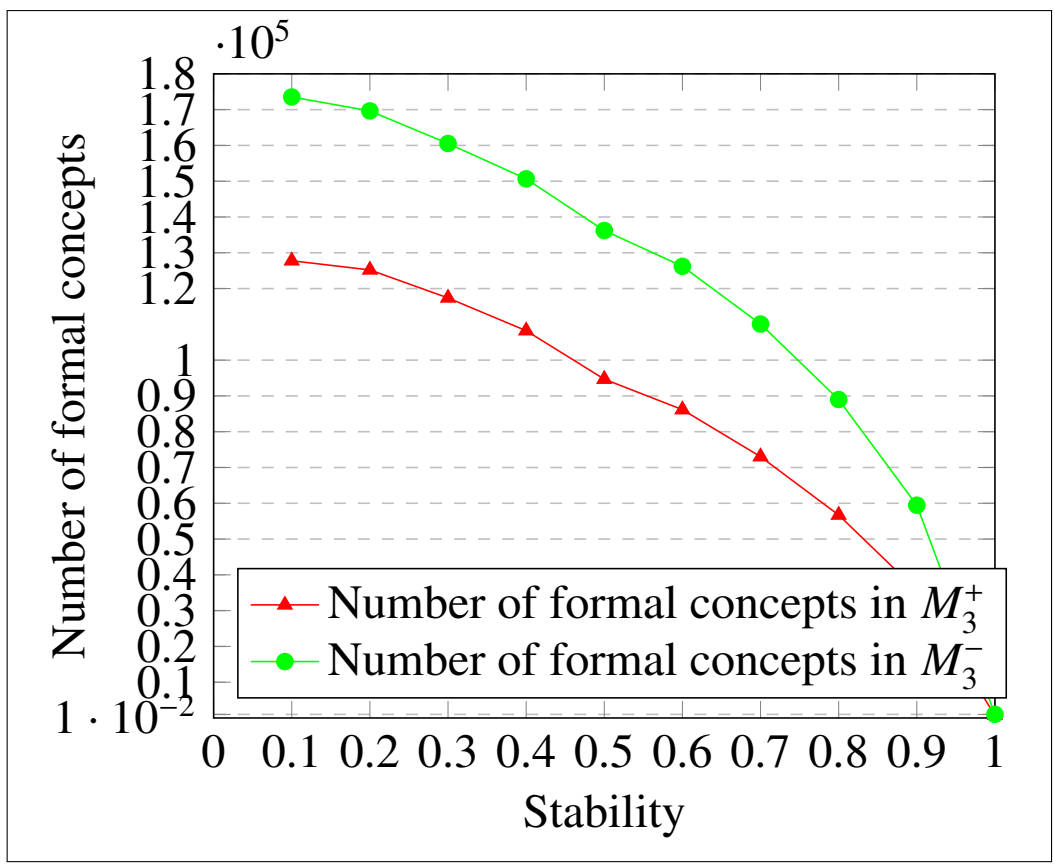
\begin{figure*}[htb]  
                  
                    \begin{center}
                    
                       \fbox{
                       \scalebox{1.5}{
                         \begin{tikzpicture}
                           \begin{axis}[
                               xlabel={Stability},
                               ylabel={Number of formal concepts },
                               xmin=0, xmax=1,
                               ymin=0, ymax=180000,
                               xtick={0,0.10,0.20,0.30,0.4,0.5,0.6,0.7,0.8,0.9,1},
                               ytick={1000,10000,20000,30000,40000,50000,60000,70000,80000,90000,100000,120000,130000,140000,150000,160000,170000,180000},
                               legend pos=south west,
                               ymajorgrids=true,
                               grid style=dashed,
                           ]

                           \addplot[
                                       color=red,
                                       mark=triangle*,
                                        ]
                          coordinates {
                         (0.1,127771)(0.20,125207)(0.30,117377)(0.40,108223)(0.50,94682)(0.60,86162)(0.70,73019)(0.80,56714)(0.90,35650)(1,766)
                                                      
                                       };           
                        \addplot[
                                                                 color=green,
                                                                 mark= otimes*,
                                                                  ]
                                      coordinates {
                                 (1,1027)(0.9,59462)(0.8,88986)(0.7,110081)(0.6,126214)(0.5,136214)(0.4,150662)(0.3,160533)(0.2,169644)(0.1,173542)
                                                                                
                                                                 }; 
                                                                 
                        {\tiny  \legend{Number of formal concepts in $ M_{3}^{+}$,Number of formal concepts in $  M_{3}^{-}$}}

                           \end{axis}
                           \end{tikzpicture}
                    
                                        }
                                        }
                            \end{center}
                          \caption{Variednumber of formal concepts (candidate biclusters) w.r.t. stability variations in Yeast Cell-Cycle dataset.}
                          \label{varying_stability}
                   \end{figure*}

 \hspace{+ 0.5 cm} Figure \ref{varying_stability} depicts more clearly the impact of the stability measure on the number of formal concepts for the Yeast Cell-Cycle dataset.
   \begin{table*}[htbp]
     \centering
    
       \begin{tabular}{|l|l|l|}
       \toprule
       \textbf{Algorithms} & \textbf{Algorithms}  & \textbf{Parameters}  \\
       \midrule
       Yeast Cell Cycle & NBic-ARM &  $\mathcal{\alpha}$1=85\% \\
        &              & $\mathcal{\alpha}$2=95\% \\
         &              & minsupp=20\% \\
          &              & minconf=95\% \\
                        & NBF             &$\mathcal{\alpha}$1=80\% \\
                                &              & $\mathcal{\alpha}$2=95\% \\
                                 &              & minstability=0.9 \\

        \midrule
         Saccharomyces Cerevisiae & Nbic-ARM & $\mathcal{\alpha}$1=80\% \\
                 &              & $\mathcal{\alpha}$2=90\% \\
                  &              & minsupp=20\% \\
                   &              & minconf=97\% \\
                                 & NBF             &$\mathcal{\alpha}$1=85\% \\
                                         &              & $\mathcal{\alpha}$2=90\% \\
                                          &              & minstability=0.8 \\
        \midrule
          Human B-cell Lymphoma & NBic-ARM & $\mathcal{\alpha}$1=85\% \\
                           &              & $\mathcal{\alpha}$2=92\% \\
                            &              & minsupp=10\% \\
                             &              & minconf=90\% \\
                                           & NBF             &$\mathcal{\alpha}$1=80\% \\
                                                   &              & $\mathcal{\alpha}$2=95\% \\
                                                    &              & minstability=1 \\
       \bottomrule
       
       \end{tabular}
       \caption{Setting algorithms for real-life datasets for different algorithms.}
     \label{parametresalgoneg}%
   \end{table*}

\subsection{Statistical relevance }
\label{statisticalresults}  
  To evaluate the statistical relevance of our algorithm, we use the coverage criterion as well as the \textit{p-value} criterion.
\subsubsection{Coverage:}
 As in Section \ref{sectioncoveragepos}, we compare the results of our algorithm versus those of Trimax \cite{Kaytoue2014} and those reported by \cite{wassimayadi2011}, namely, CC \cite{DBLP:conf/ismb/ChengC00}, BiMine \cite{Ayadi2009}, BiMine+ \cite{Ayadi2012a}, BicFinder \cite{Ayadi2012}, MOPSOB \cite{Liu2008}, MOEA \cite{Mitra2006} and SEBI \cite{Divina2006}.

 Table \ref{coveragelymphoma} (resp. Table \ref{coverageyast}) presents the coverage of the obtained biclusters. We can show that most of the algorithms have relatively close results. For the Human B-cell Lymphoma (respectively Yeast Cell-Cycle) dataset, the biclusters extracted by the NBF algorithm cover 100\% (respectively 97.08\%) of genes, 100\% of conditions and 73.15\% (respectively 77.17\%) of cells in the initial matrix. However, Trimax has a low performance since it covers only 8.50 \% of cells, 46.32 \% of genes and 11.46 \% of conditions. This implies that our algorithm can generate biclusters with high coverage of a data matrix due to the discretisation phase, where the combinations of all the paired conditions give useful information since a subset of non contiguous conditions may compose a bicluster.

 For the NBic-ARM algorithm, in the Human B-cell Lymphoma dataset (respectively Yeast Cell-Cycle), the biclusters extracted by our algorithm cover 100\% of genes, 94.74\% (respectively \%)  of conditions and 94.73\%  (respectively \%) of cells in the initial matrix. However, Trimax has a low performance since it covers only 8.50 \% of cells, 46.32 \% of genes and 11.46 \% of conditions. This implies that our algorithm can generate biclusters with high coverage of a data matrix due to the discretization phase and the use of the $\mathcal{IGB}$ (Informative Generic Base) representation.
  \begin{table}[htbp]
    \centering
     
      \scalebox{0.7}{
      \resizebox{1.2 \linewidth}{!}{
      \begin{tabular}{|l|r|r|r|}
      \toprule
      \multicolumn{4}{|c|}{\textbf{Human B-cell Lymphoma}} \\
      \midrule
      \textbf{Algorithms} & \textbf{Total coverage} & \textbf{Gene coverage} & \textbf{Condition coverage} \\
      \midrule
      BiMine & 8.93\% & 26.15\% & 100\% \\
      BiMine+ & 21.19\% & 46.26\% & 100\% \\
      BicFinder & 44.24\% & 55.89\% & 100\% \\
      MOPSOB & 36.90\% & \_    & \_ \\
      MOEA  & 20.96\% & \_    & \_ \\
      SEBI  & 34.07\% & 38.23\% & 100\% \\
      CC    & 36.81\% & 91.58\% & 100\% \\
     Trimax & 8.50\% & 46.32\% & 11.46\% \\
     \toprule
    \textbf{ NBF} &\textbf{ 73.75 \%} & \textbf{100\%} & \textbf{100\%} \\
      \textbf{ NBic-ARM} &\textbf{ 94.73 \%} & \textbf{100\%} & \textbf{94.74}\% \\
      \bottomrule
      \end{tabular} } }
      \caption{Human B-cell Lymphoma coverage for different algorithms.}
  \label{coveragelymphoma}%
  \end{table}%
 
   \begin{table}[htbp]
       \centering
       
         \scalebox{0.7}{
         \resizebox{1.2 \linewidth}{!}{
         \begin{tabular}{|l|r|r|r|}
         \toprule
         \multicolumn{4}{|c|}{\textbf{Yeast Cell-Cycle}} \\
         \midrule
         \textbf{Algorithms} & \textbf{Total coverage} & \textbf{Gene coverage} & \textbf{Condition coverage} \\
         \midrule
         BiMine & 13.36\% & 32.84\% & 100\% \\
         BiMine+ & 51.76\% & 68.65\% & 100\% \\
         BicFinder & 55.43\% & 76.93\% & 100\% \\
         MOPSOB & 52.40\% & \_    & \_ \\
         MOEA  & 51.34\% & \_    & \_ \\
         SEBI  & 38.14\% & 43.55\% & 100\% \\
         CC    & 81.47\% & 97.12\% & 100\% \\
          Trimax & 15.32\% & 22.09\% & 70.59\% \\
          \toprule
        \textbf{NBF} & \textbf{77.17 \%} & \textbf{97.08\%} & \textbf{100\%} \\
         \textbf{ NBic-ARM} &\textbf{60.4 \%} & \textbf{80.33\%} & \textbf{99\%} \\
         \bottomrule
         
         \end{tabular}
       }}
        \caption{Yeast Cell-Cycle coverage for different algorithms.}
       \label{coverageyast}%
     \end{table}
\subsubsection{p-value:}
  As in \cite{DBLP:conf/ismb/ChengC00,Ayadi2014,DBLP:journals/bioinformatics/IhmelsBB04,DBLP:journals/jcb/Ben-DorCKY03,Prelic2006}, to assess the quality of the extracted biclusters, we use the web tool \textit{\textbf{FuncAssociate}}\footnote{Available at  \url{http://llama.mshri.on.ca/funcassociate/}} \cite{Berriz2003} in order to compute the adjusted significance scores for each bicluster (adjusted \textit{p-value}\footnote{The adjusted significance scores assess genes in each bicluster, which indicates how well they match with the different GO categories.}). In this test, we compute the percentage of biclusters having an adjusted \textit{p-value}, i.e. the proportion between the number of biclusters having an adjusted \textit{p-value} and the total number of obtained bicluters. We compute the adjusted \textit{p-value} \cite{Prelic2006} based on the exact value of Fisher test \cite{Fisher1922}, to measure the quality of the obtained biclusters. In fact, the biclusters having a \textit{p-value} lower than 5\% are considered as over-represented; i.e., the majority of genes of a bicluster have common biological characteristics. Thus, the best biclusters are those having an adjusted \textit{p-value} less than 0.001\%. The results of our algorithm are compared with CC \cite{DBLP:conf/ismb/ChengC00}, ISA \cite{DBLP:journals/bioinformatics/IhmelsBB04} and Bimax \cite{Prelic2006}. We report the results of the algorithms mentioned before from \cite{wassimayadi2011}. We also compare our algorithm with Trimax \cite{Kaytoue2014} and the MBA algorithm \cite{Ayadi2014}.

\begin{figure*}[!t]
    \begin{center}
  \fbox{  \includegraphics[width=1.0\textwidth]{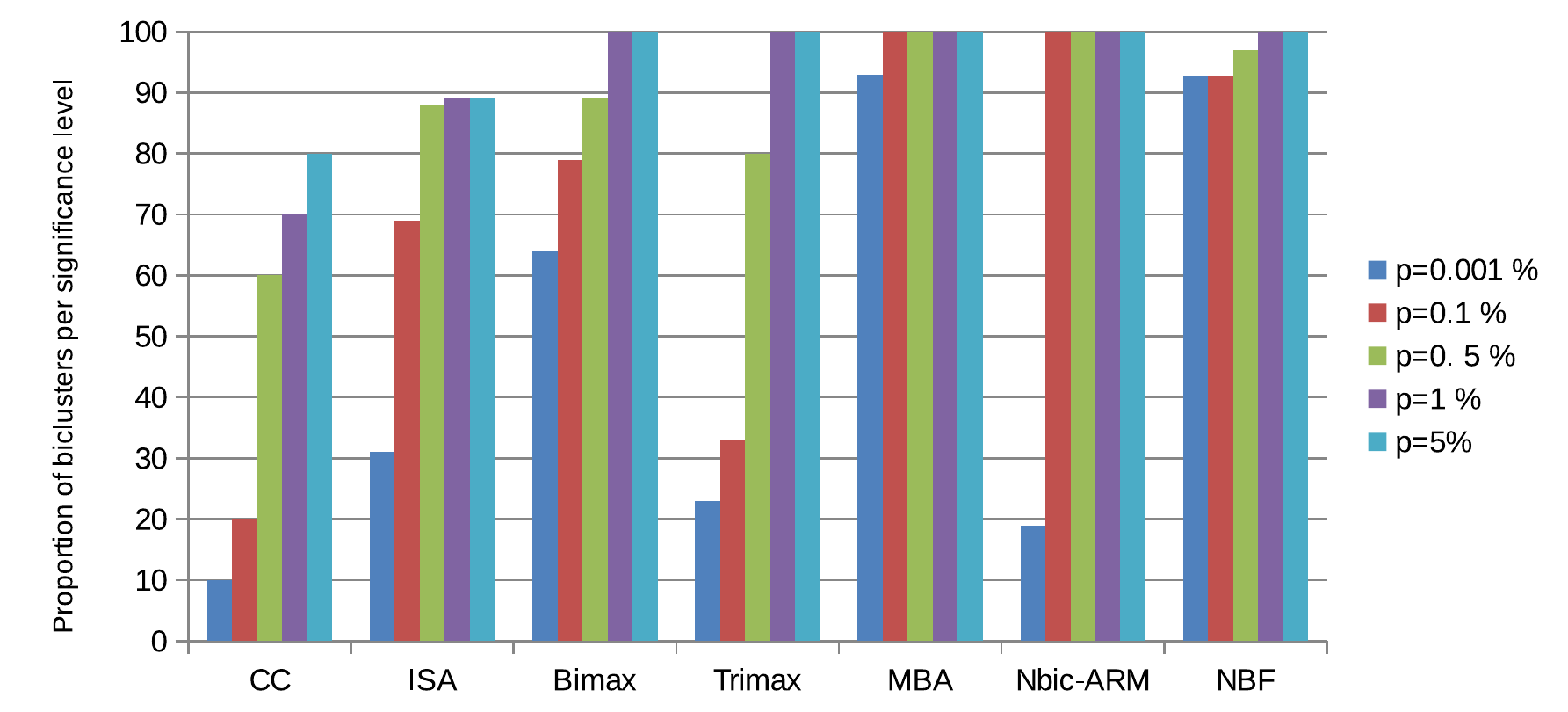} }
    \end{center}
    \caption{Proportions of biclusters significantly enriched by GO annotations (Yeast Cell-Cycle).}
    \label{pvaleuryeastneg}
    \end{figure*}

  The obtained results of the Yeast Cell Cycle dataset for the different adjusted \textit{p-values} (p = 5\%; 1\%; 0.5\%; 0.1\%; 0.001\%) for each algorithm over the percentage of total biclusters are depicted in Figure \ref{pvaleuryeastneg}. The \textit{NBF} results show that 92.68\% of the extracted biclusters are statistically significant with the adjusted \textit{p-value} $p <0.001 \%$ and $p <0.1 \%$. By contrast, Trimax achieves 23\% of statistically significant biclusters when $p <0.001 \%$. With the same dataset and utilizing the NBic-ARM algorithm for different significance levels, our algorithm achieves 19\% of statistically significant biclusters when $p <0.001 \%$. MBA, however, outperforms all algorithms in that 93\% of its discovered biclusters are statistically significant with the \textit{p-value} $p <0.001 \%$.

\begin{figure*}[!t]
    \begin{center}
  \fbox{  \includegraphics[width=1.0\textwidth]{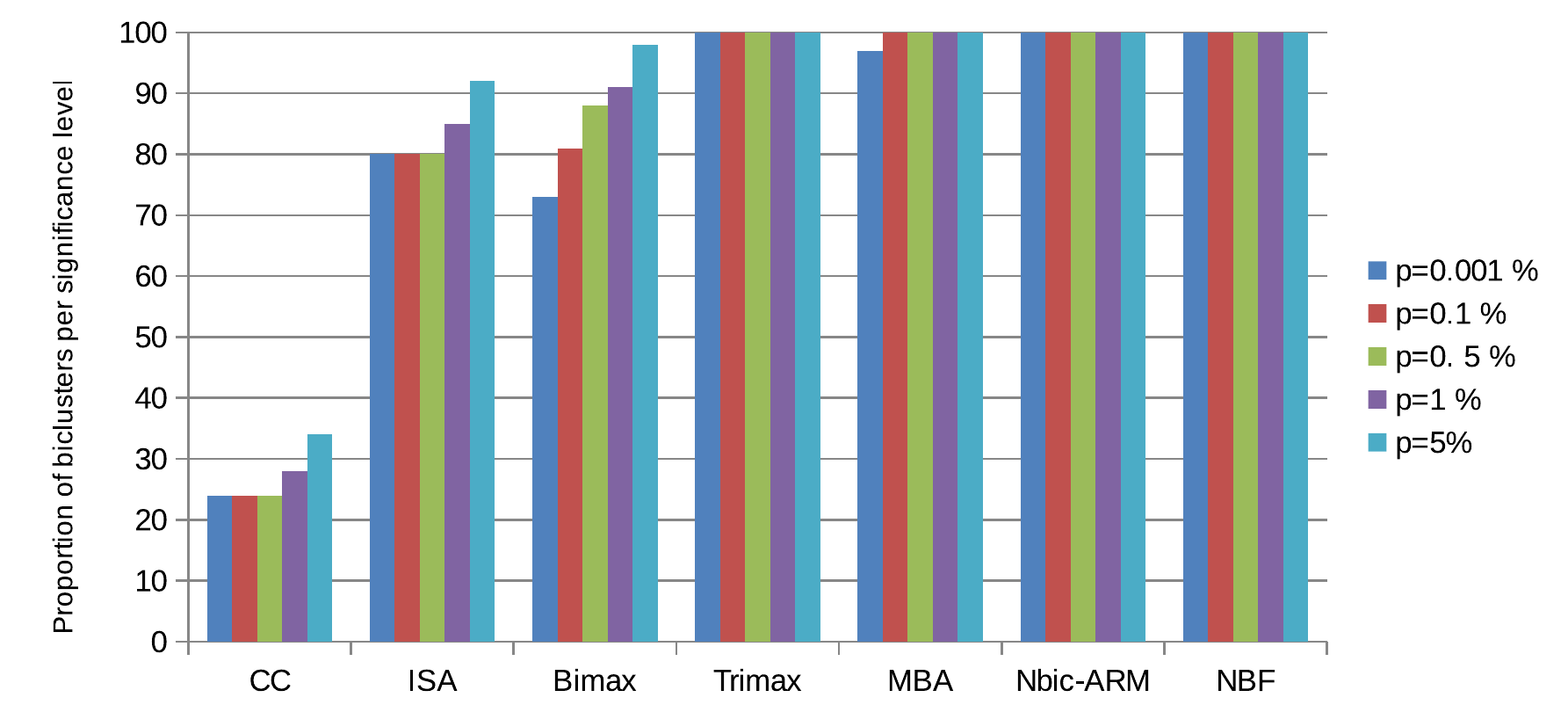} }
    \end{center}
    \caption{Proportions of biclusters significantly enriched by GO annotations (Saccharomyces  Cerevisiae dataset).}
    \label{pvaleursacneg}
    \end{figure*}

 The obtained results of the Saccharomyces Cerevisiae dataset for the different adjusted \textit{p-values} (p = 5\%; 1\%; 0,5\%; 0,1\%; 0,001\%) for each algorithm over the percentage of total biclusters are illustrated in Figure \ref{pvaleursacneg}. The \textit{NBic-ARM} and NBF results indicate that 100\% of the extracted biclusters are statistically significant with the adjusted \textit{p-value} $p <0.001 \%$, which is the same as those results achieved by the Trimax algorithm. On the other hand, MBA achieves its best results whenever $p<0.1\%$. It can be concluded that a high negative correlation among genes implies biologically relevant biclusters according to GO.

   Our algorithms has close results with the other algorithms regarding the Yeast Cell ­Cycle dataset since this dataset contains only positive integer values. By contrast, the Saccharomyces Cerevisiae dataset contains real values (including negative ones) since the Yeast Cell ­Cycle dataset has only a small number of conditions and our algorithsm is sensitive to the number of items due to the discretization phase.\\ 
    
\subsection{Biological relevance }
\label{biologicalresults}

 As in \cite{Ayadi2014,Freitas20013,Mitra2006} and \cite{Ayadi2009}, we use the biological criterion which allows measuring the quality of the obtained biclusters, by checking whether the genes of a bicluster have common biological characteristics.\\
To evaluate the quality of the extracted biclusters and identify their biological annotations, we use \textbf{\textit{GOTermFinder}}\footnote{Available at \url{http://db.yeastgenome.org/cgi-bin/GO/goTermFinder }} which is designed to search for the significant shared \textit{GO} terms of a group of genes. The GOs are represented by direct acyclic graphs where GO terms represent nodes and the relationships between them represent edges.\\

 We present in Table \ref{go} the result of a randomly selected biclusters for the \textit{biological process}, the \textit{molecular function} and the \textit{cellular component} and we report the most significant GO terms. The values in parentheses after each GO term in Table \ref{go}, e.g., (11.7\%, 6.9\%, 4.98e-07) in the first bicluster, respectively stand for the cluster frequency, the background frequency and the statistical significance. The cluster frequency shows that for the first bicluster, 11.7\% of genes belong to this process, while the background frequency demonstrates that this bicluster contains 6.9\% of the number of genes in the background set. Finally, the statistical significance is supplied by a \textit{p-value} of 4.98e-07, which is highly significant. 
 
 We show in Tables \ref{go1} and \ref{go2} the biological annotations of two randomly selected biclusters in terms of the above cited axes, where we report the most significant GO terms. For instance, with the first bicluster extracted from the Saccharomyces Cerevisiae dataset (Table \ref{go2}), the values within parentheses after each GO term, such as (93.2\%, 69.1\%, 1.20e-242), indicate that for the first bicluster, 93.2\% of genes belong to this process. The background frequency shows that this bicluster contains 69.1\% of the number of genes in the background set. Finally, the statistical significance is provided by a \textit{p-value} of 1.20e-242 (highly significant).

The results on these real-life datasets show that our proposed algorithm can identify biclusters with a high biological relevance.

\begin{landscape}
\begin{center}

   \begin{table*}[htbp]
     \centering
    
      \scalebox{1.0}{
      \vspace{+ 3.6 cm}
    \begin{tabular}{l|l|l} \toprule &\textbf{ Bicluster 1} & \textbf{Bicluster 2 }\\ \midrule \multirow{8}[2]{*}{\textbf{Biological process}} & ribonucleoprotein complex biogenesis & cytoplasmic translation \\ & \textbf{11.7\%, 6.9\%, 4.98e-07}  & \textbf{6.1\%, 2.5\%, 2.67e-09} \\ & maturation of SSU-rRNA & ribosomal large subunit biogenesis \\ & \textbf{4.1\%, 1.6\%, 1.81e-06} & \textbf{3.4\%, 1.4\%, 7.50e-05} \\ & maturation of SSU-rRNA from & cellular component biogenesis \\ & tricistronic rRNA transcript & \textbf{20.6\%, 15.5\%, 0.00245} \\  & \textbf{3.7\%, 1.5\%, 1.85e-05} & \\ \midrule \multirow{5}[2]{*}{\textbf{Molecular function}} & structural constituent of ribosome & structural molecule activity \\ & \textbf{7.2\%, 3.2\%, 3.94e-10} & \textbf{9.3\%, 4.9\%, 2.27e-08} \\& structural molecule activity & structural constituent of ribosome \\ & \textbf{9.1\%, 4.9\%, 6.03e-08} & \textbf{7.0\%, 3.2\%, 9.30e-09} \\ \midrule \multirow{5}[2]{*}{\textbf{Cellular component}} & cytosolic ribosome & cytosolic part \\ & \textbf{6.9\%, 2.5\%, 1.16e-14} & \textbf{7.3\%, 3.3\%, 1.31e-09} \\ & ribosomal subunit & non-membrane-bounded organelle \\ & \textbf{7.8\%, 3.5\%, 4.86e-11} & \textbf{25.2\%, 18.4\%, 4.46e-07} \\ \bottomrule
       \end{tabular}} 
      
       \caption{Significant GO terms (process, function, component) for two biclusters on Yeast Cell-Cycle dataset extracted by \textit{NBF}.}
       \label{go}
   \end{table*}                                                                    
  \end{center}
  \end{landscape}
 \vspace{+3.3 cm} 
\begin{landscape}
  \begin{table*}[htbp]
     \centering
    
      \scalebox{0.9}{
     \begin{tabular}{rrr}
        \toprule
              & \multicolumn{1}{c}{\textbf{Bicluster1}} & \multicolumn{1}{c}{\textbf{Bicluster2}} \\
        \midrule 
        \multicolumn{1}{c}{\multirow{3}[2]{*}{\textbf{Biological process}}} & \multicolumn{1}{l}{cellular response to stress  \textbf{(12.3\%, 8.5\%,6.37e-09)}} & \multicolumn{1}{l}{cell cycle process \textbf{(12.1\%, 8.6\%, 6.28e-09)}} \\
         \multicolumn{1}{c}{} & \multicolumn{1}{l}{ } & \multicolumn{1}{l}{} \\   
        \multicolumn{1}{c}{} & \multicolumn{1}{l}{cellular response to stimulus \textbf{(17.0\%, 12.6\%, 7.87e-09)}} & \multicolumn{1}{l}{single-organism process \textbf{(48.1\%, 42.7\%, 3.46e-07)}} \\
        \multicolumn{1}{c}{} & \multicolumn{1}{l}{} & \multicolumn{1}{l}{} \\
        \bottomrule
        \multicolumn{1}{c}{\multirow{3}[2]{*}{\textbf{Molecular function}}} & \multicolumn{1}{l}{structural molecule activity \textbf{(6.7\%,}} & \multicolumn{1}{l}{structural molecule activity  \textbf{(6.6\%, 4.9\%, 0.00752)}} \\
        \multicolumn{1}{c}{} & \multicolumn{1}{l}{ \textbf{4.9\% , 0.00655)}} & \multicolumn{1}{l}{} \\
         \multicolumn{1}{c}{} & \multicolumn{1}{l}{ } & \multicolumn{1}{l}{} \\   
        \bottomrule
        \multicolumn{1}{c}{\multirow{3}[2]{*}{\textbf{Cellular component}}} & \multicolumn{1}{l}{replication fork  \textbf{(1.9\%, 0.8\% , 2.91e-06 )}} & \multicolumn{1}{l}{non-membrane-bounded organelle \textbf{(23.3\%, 19.0\%, 1.14e-07)}} \\
         \multicolumn{1}{c}{} & \multicolumn{1}{l}{ } & \multicolumn{1}{l}{} \\   
        \multicolumn{1}{c}{} & \multicolumn{1}{l}{nuclear chromosome  \textbf{(6.0\%, 3.9\%, 4.98e-06)}} & \multicolumn{1}{l}{chromosomal part \textbf{(7.3\%, 5.0\%, 7.26e-07)} } \\
         \multicolumn{1}{c}{} & \multicolumn{1}{l}{ } & \multicolumn{1}{l}{} \\   
        \multicolumn{1}{c}{} & \multicolumn{1}{l}{organelle \textbf{(69.1\%, 64.4\%, 3.74e-05)}} & \multicolumn{1}{l}{cytosolic small ribosomal subunit \textbf{(1.9\%, 0.9\%, 1.97e-05)}} \\
         \multicolumn{1}{c}{} & \multicolumn{1}{l}{ } & \multicolumn{1}{l}{} \\   
        \bottomrule
        \end{tabular} }
       \caption{Significant GO terms (process, function, component) for two biclusters on Yeast Cell-Cycle dataset extracted by \textit{NBic-ARM}.}
       \label{go1}
   \end{table*}%
\end{landscape}

\begin{landscape} 
 \begin{table*}[htbp]
                                                                                                         \centering
                                                                                                        
\scalebox{1.0}{                                        
\begin{tabular}{rrr}
                                                                                                            \toprule
                                                                                                                  & \multicolumn{1}{c}{\textbf{Bicluster1}} & \multicolumn{1}{c}{\textbf{Bicluster2}} \\
                                                                                                            \midrule 
                                                                                                            \multicolumn{1}{c}{\multirow{5}[2]{*}{\textbf{Biological process}}} & \multicolumn{1}{l}{cellular process  \textbf{(93.2\%, 69.1\%,1.20e-242)}} & \multicolumn{1}{l}{primary metabolic process \textbf{(73.1\%,  48.3\%, 2.26e-167)}} \\
                                                                                                             \multicolumn{1}{c}{} & \multicolumn{1}{l}{ } & \multicolumn{1}{l}{} \\   
 \multicolumn{1}{c}{} & \multicolumn{1}{l}{metabolic process \textbf{(78.0\%, 53.3\%, 9.23e-195)}} & \multicolumn{1}{l}{cellular metabolic process \textbf{(75.2\%,  50.7\%, 9.74e-165)}} \\
                         \multicolumn{1}{c}{} & \multicolumn{1}{l}{ } & \multicolumn{1}{l}{} \\                                                                                   
                                                                                                            \multicolumn{1}{c}{} & \multicolumn{1}{l}{organic substance metabolic process } & \multicolumn{1}{l}{organic substance biosynthetic process \textbf{(47.8\%, 29.7\%,}} \\
                                                                                                            \multicolumn{1}{c}{} & \multicolumn{1}{l}{\textbf{(76.4\%, 51.3\% ,6.84e-198 )}} & \multicolumn{1}{l}{ \textbf{1.23e-99 )}} \\
                  \multicolumn{1}{c}{} & \multicolumn{1}{l}{ } & \multicolumn{1}{l}{} \\                                                                                           
                                                                                                            \bottomrule
                                                                                                            \multicolumn{1}{c}{\multirow{4}[2]{*}{\textbf{Molecular function}}} & \multicolumn{1}{l}{structural molecule activity\textbf{ (10.5\%,}} & \multicolumn{1}{l}{structural constituent of ribosome (\textbf{7.7\%, 3.2\%, 3.29e-38)}} \\
                                                                                                            \multicolumn{1}{c}{} & \multicolumn{1}{l}{ \textbf{4.9\% , 5.37e-46)}} & \multicolumn{1}{l}{} \\
            \multicolumn{1}{c}{} & \multicolumn{1}{l}{ } & \multicolumn{1}{l}{} \\                                                                                                \multicolumn{1}{c}{} & \multicolumn{1}{l}{transferase activity  \textbf{(19.1\%, 11.7\% ,} } & \multicolumn{1}{l}{hydrolase activity \textbf{(20.0\%,12.4\%,2.95e-32)}} \\
                                                                                                            \multicolumn{1}{c}{} & \multicolumn{1}{l}{\textbf{9.78e-37)}} & \multicolumn{1}{l}{ } \\
                                                                                                             \multicolumn{1}{c}{} & \multicolumn{1}{l}{ } & \multicolumn{1}{l}{} \\   
\bottomrule
                                                                                                            \multicolumn{1}{c}{\multirow{3}[2]{*}{\textbf{Cellular component}}} & \multicolumn{1}{l}{cell part \textbf{(97.1\%, 77.5\% , 7.16e-212 )}} & \multicolumn{1}{l}{intracellular part \textbf{(92.6\%, 73.5\%,5.43e-147 )}} \\
                                                                                                       
                                                                                                            \multicolumn{1}{c}{} & \multicolumn{1}{l}{} & \multicolumn{1}{l}{intracellular organelle \textbf{(80.5\%, 60.6\%, 8.48e-118 )} } \\
                                                                                                     \multicolumn{1}{c}{} & \multicolumn{1}{l}{ } & \multicolumn{1}{l}{} \\   
                                                                                                            \multicolumn{1}{c}{} & \multicolumn{1}{l}{intracellular \textbf{(92.3\%, 73.7\%,1.63e-156)}} & \multicolumn{1}{l}{organelle \textbf{(80.6\%, 60.7\%, 1.27e-117)}} \\

      \multicolumn{1}{c}{} & \multicolumn{1}{l}{ } & \multicolumn{1}{l}{} \\                                                                                                          \bottomrule
                                                                                  \end{tabular}}
\caption{Significant GO terms (process, function, component) for two biclusters on Saccharomyces Cerevisiae dataset extracted by \textit{NBic-ARM}.}                                                                                  \label{go2}%
                                                                                                       \end{table*}
\end{landscape}

\begin{figure*}[htb]
                \begin{center}
               \fbox{ \includegraphics[width=0.9\textwidth]{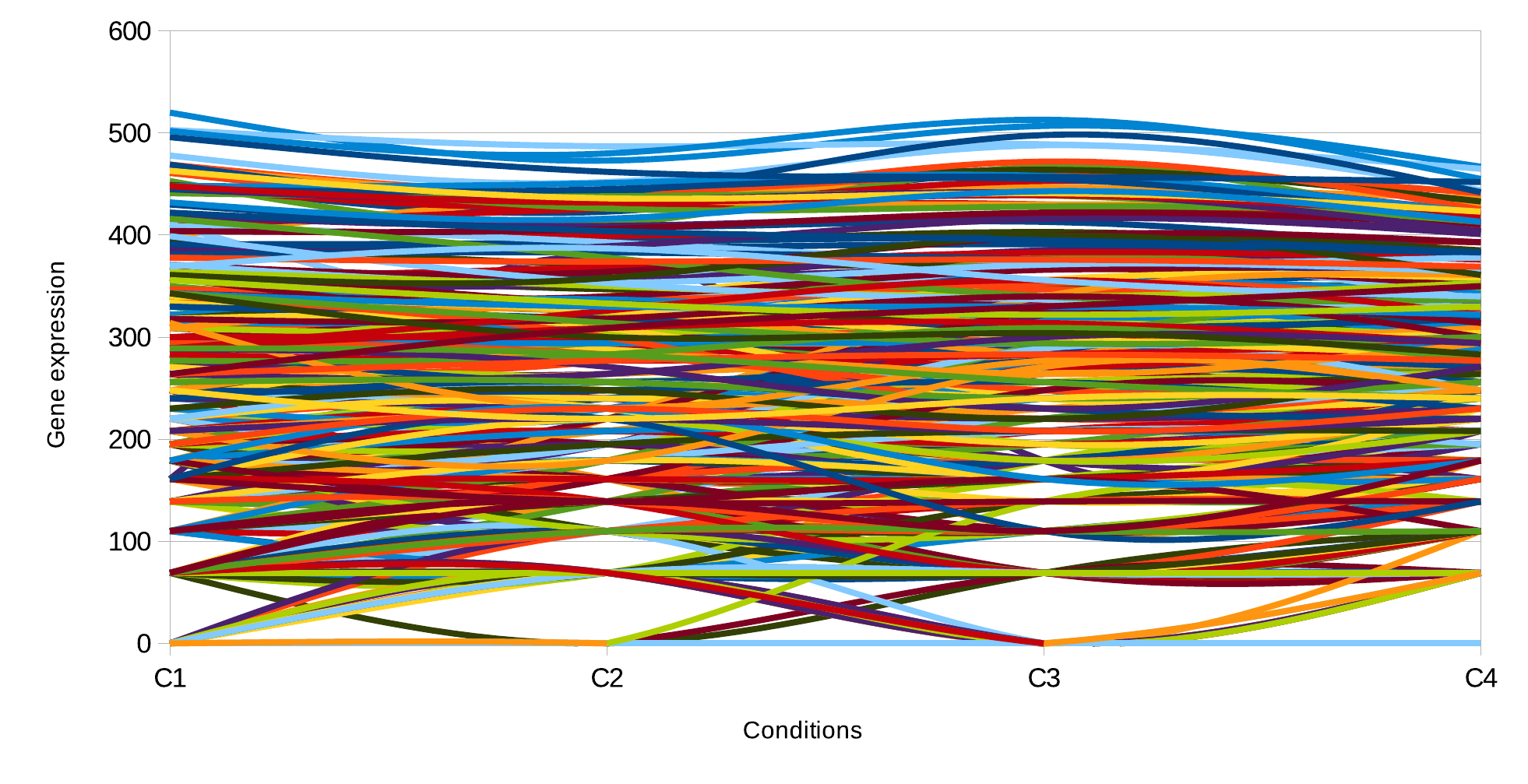}} 
                \end{center}
               
                \caption{Resulting bicluster profile obtained through NBic-ARM algorithm on Yeast Cell-Cycle.}
                 \label{yeast_courbe}
                \end{figure*}

\begin{figure*}[htb]
                \begin{center}
               \fbox{ \includegraphics[width=0.9\textwidth]{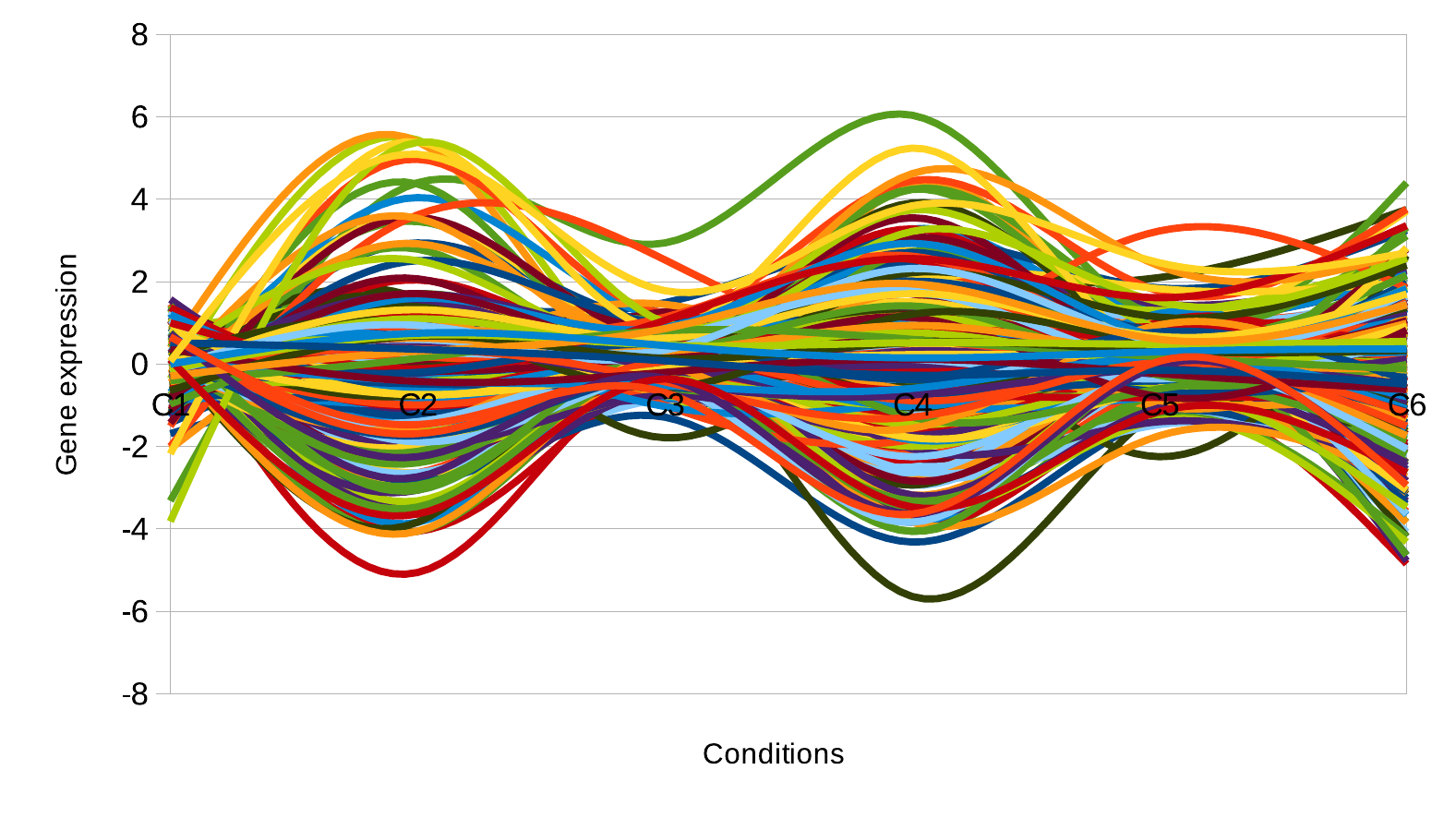}} 
                \end{center}
               
                \caption{Resulting bicluster profile obtained through NBic-ARM algorithm on Saccharomyces Cerevisiae.}
                 \label{sacc_courbe}
                \end{figure*}

\begin{figure*}[htb]
                    \begin{center}
                   \fbox{ \includegraphics[width=0.9\textwidth]{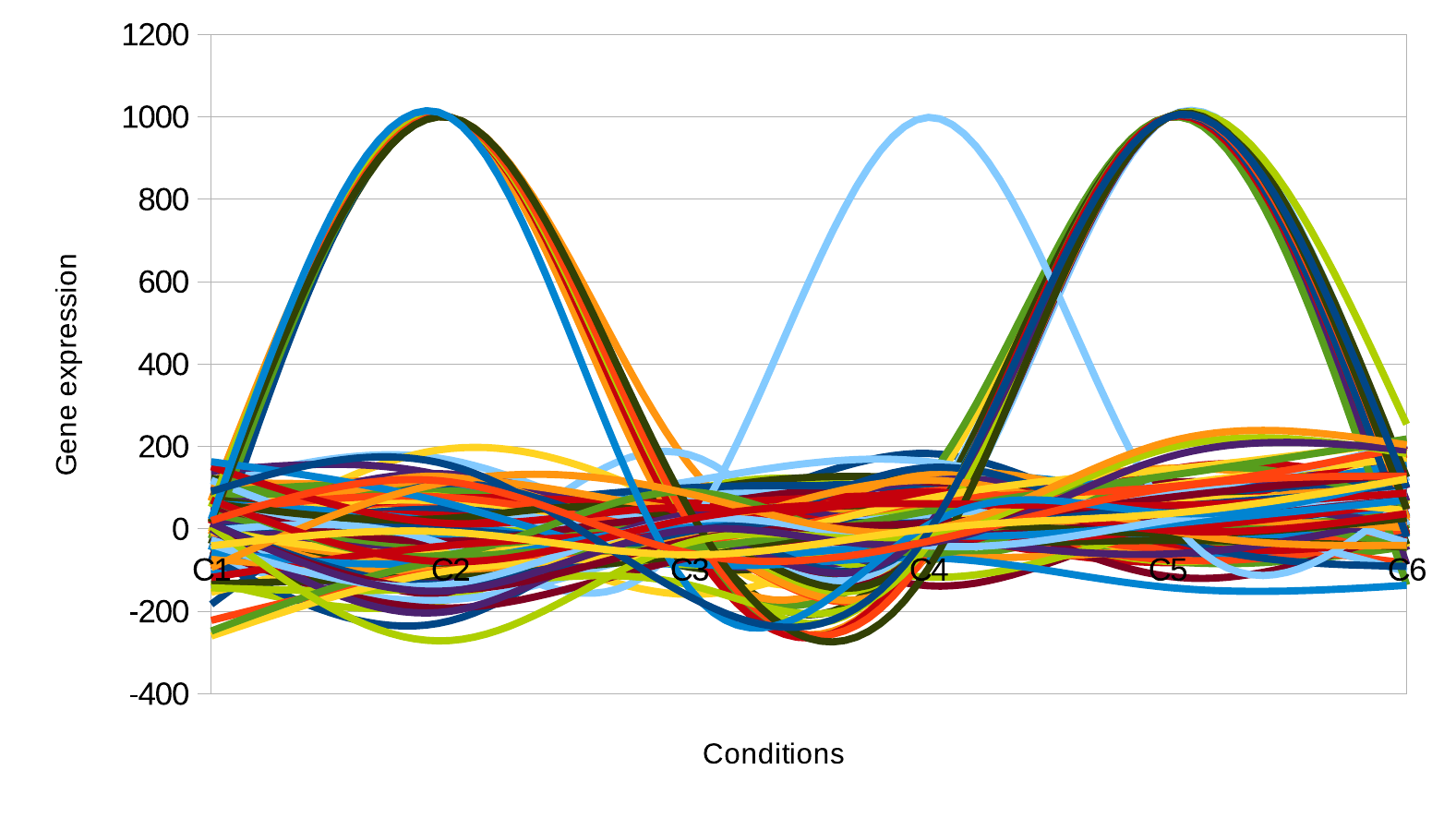}} 
                    \end{center}
                    
                    \caption{Resulting bicluster profile obtained through NBic-ARM algorithm on Human B-cell Lymphoma.}
                     \label{lymphoma_courbe}
                    \end{figure*}

The experiments are carried out on the different datasets. Figure \ref{yeast_courbe} (resp. \ref{sacc_courbe} and \ref{lymphoma_courbe}) depicts the profile of a randomly selected bicluster obtained by our algorithm for the Yeast Cell-Cycle  (resp. Saccharomyces Cerevisiae and Human B-cell Lymphoma) dataset. From these figures, negative patterns can be observed. $\beta$ contains two subsets of genes $\beta$$_{1}$ and $\beta$$_{2}$ showing an opposite changing tendency over the subset of experimental conditions; and the genes in each subset have similar expression tendencies. In fact, if two subsets $\beta$$_{1}$ and $\beta$$_{2}$ show an opposite changing tendency over a subset of experimental conditions (like in Figure \ref{yeast_courbe} (resp. \ref{sacc_courbe} and \ref{lymphoma_courbe})), we assume that $\beta$$_{1}$ is negatively correlated with $\beta$$_{2}$ \cite{Henriques2014,Nepomuceno2015}.\\

\begin{figure}[!t]
    \begin{center}
    \fbox{\includegraphics[width=0.9\textwidth]{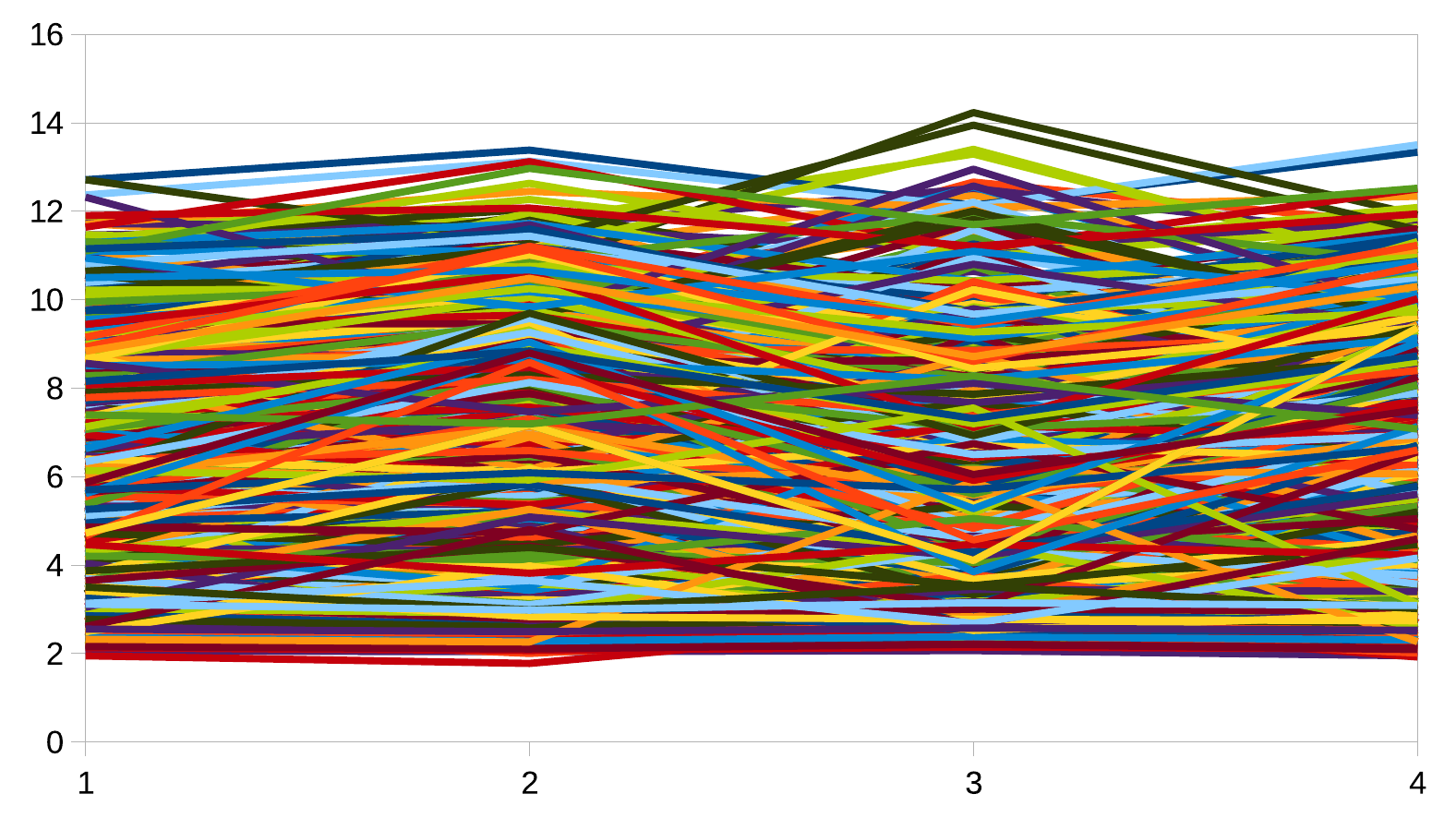}} 
    \end{center}
   
    \caption{Bicluster profile obtained by NBF algorithm for Saccharomyces Cerevisiae dataset.}
   
    \label{profilealzheimer}
    \end{figure}

\begin{figure}[!t]
       \begin{center}
 \fbox{\includegraphics[width=0.9\textwidth]{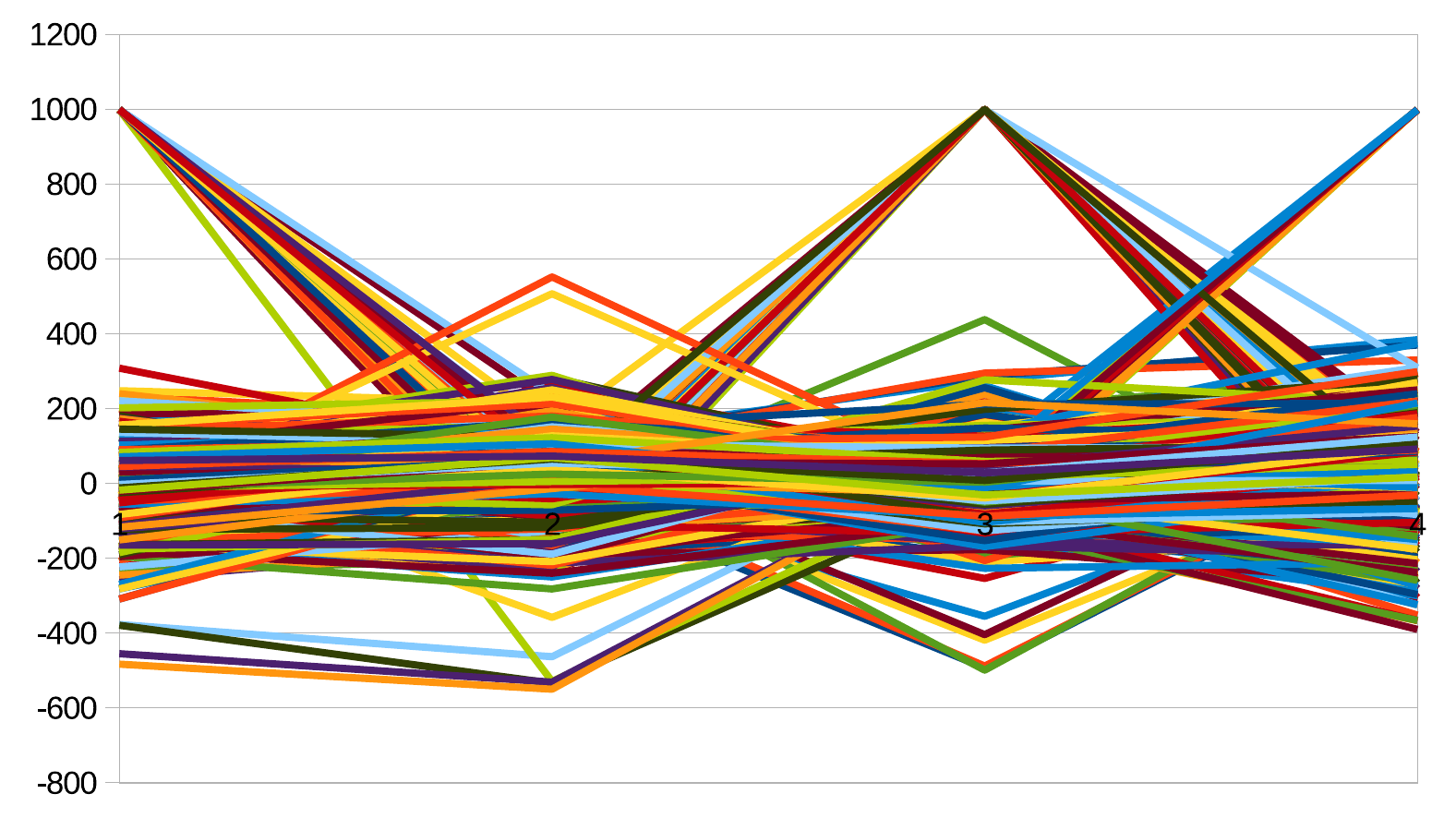}} 
       \end{center}
       
       \caption{Bicluster profile obtained by NBF algorithm for Human B-Cell Lymphoma dataset.}
        \label{profilelymphoma}
       \end{figure}

\begin{figure}[!t]
             \begin{center}
            \fbox{ \includegraphics[width=0.9\textwidth]{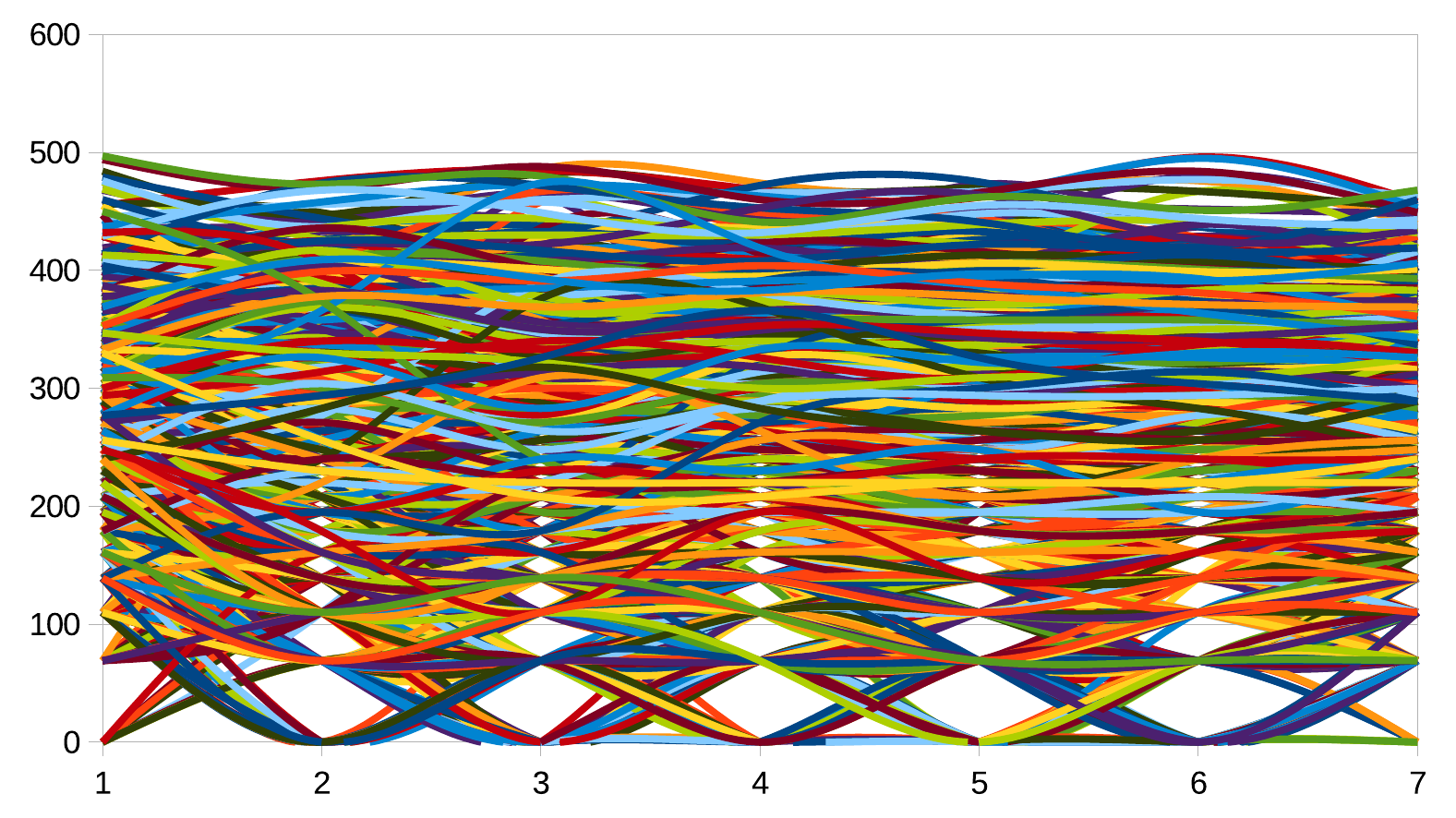}} 
             \end{center}
             
             \caption{Bicluster profile obtained by NBF algorithm for Yeast Cell-Cycle dataset.}
             \label{profileyeast}
             \end{figure}

Furthermore, Figure \ref{profilealzheimer} (resp. \ref{profilelymphoma} and \ref{profileyeast}) illustrates the profile of a bicluster obtained by the NBF algorithm for the Saccharomyces Cerevisiae (resp. Human B-Cell Lymphoma and Yeast Cell-Cycle) dataset. From these figures, negative patterns can be observed.          
   
   \section{Conclusion}
   We have introduced in this chapter the NBic-ARM and the NBF approaches to mine negatively-correlated biclusters from gene expression data. Our approches are based on the key notions of ARM and FCA. Finally, this chapter is concluded by the presentation of our experimental evaluation of NBic-ARM an NBF on real-life datasets according to both statistical and biological aspects.
     

    

\chapter*{Conclusion}

 \markboth{Conclusion}{Conclusion}

\addcontentsline{toc}{chapter}{Conclusion}
\section*{Thesis summary}
Throughout this dissertation, we have presented the problem of biclustering gene expression data. This thesis report is partitioned into two different parts. The first part is dedicated to present the theoretical concepts used in this thesis. In this regard, we have started the first chapter of this part by introducing the basic notions related to Formal Concept Analysis (FCA) and Association Rule Mining (ARM) which offer the basis for the proposition of our approaches. Thereafter, we have studied in the second chapter of this part, the related work dealing with the biclustering problem.

The second part is dedicated to the presentation of our proposed approaches. Our contributions concern two main parts related to \textit{(i)} extracting biclusters of positive correlations and \textit{(ii)} extracting negatively correlated biclusters. 
\vspace{+0.7 cm}

\hspace{+ 0.5 cm}In the following, we summarize the contributions made in each part.
\subsection*{Mining biclusters of positive correlations}
First, a new Association Rule Mining (ARM)-based biclustering method (\textit{\textbf{BiARM}}) has been proposed as a new biclustering algorithm for gene expression data. Our algorithm relies on the extraction of ARs from the dataset by discretizing this latter into a binary data matrix. The resulting biclusters were filtered with the help of the similarity measure in order to remove those with a high overlap. The performance of the \textit{BiARM} algorithm is assessed on real-life DNA microarray datasets. These experimentations show that \textit{BiARM} allows extracting high quality biclusters. These biclusters have been evaluated with \textit{Gene Ontology} (GO) annotations which checks the biological significance of biclusters. The obtained results confirm the \textit{BiARM}'s ability to extract significant biclusters. BiARM, however, extracts an enormous number of biclusters which in turn entails an increase in the computation time of the filtering step.

\hspace{+0.5 cm}After that, a new FCA-based biclustering method for gene expression data has been proposed. Our approach consists in extracting formal concepts from a dataset after a discretization into a 3-state data matrix. The 3-state data matrix allows observing the profile of each gene through all pairs of adjacent conditions in the gene expression matrix. This latter discretization is used to extract formal concepts, a mathematical framework for deriving implicit relationships from a set of objects and their attributes. The resulting formal concepts represent biclusters. These biclusters are filtered with the help of the \textit{Bond} correlation measure in order to remove the biclusters that have a high overlap. The performances of the \textit{BiFCA+} algorithm have been assessed on three real-life DNA microarray datasets. These experimentations show that \textit{BiFCA+} enables extracting high quality biclusters. These biclusters have been evaluated with the \textit{GO} annotations which check the biological significance of biclusters. The obtained results confirm the \textit{BiFCA+}'s ability to extract significant biclusters. However, a 3-state data matrix allows observing the profile of each gene through all pairs of adjacent conditions in the gene expression matrix. Nevertheless, a close look at existing studies proves that our results will be much improved if we extend the discretization of all columns and not only those which are adjacent.

\hspace{+0.5 cm} Interestingly enough, we have introduced the \textit{BiFCA} algorithm, an optimized version of the \textit{BiFCA+} algorithm which presents a much better performance than \textit{BiFCA+} over different datasets. Our proposed method has aimed to extract positively-correlated biclusters. The main concept of the proposed algorithm has been to extract formal concepts from a binary data set. That is why we discritize the original data matrix into a -101 data matrix which permits observing the profile of genes through all pairs of conditions. The latter is also discretized to a binary data matrix. The obtained experimental results have highlighted interesting rates compared to its competitors. However, the BiFCA algorithm consumes an anourmous amount of time when run on large datasets. This is even more so when combining genes and condition. Hence, a new method, based on the MapReduce paradigm is necessary.
\subsection*{Mining biclusters of negative correlations}
The second part of our contributions has concerned the extraction of negatively correlated biclusters. We have started by proposing a new ARM-based biclustering method (\textit{\textbf{NBic-ARM}}) as a new biclustering algorithm for discovering negative biclusters from gene expression data. Our algorithm relies on the extraction of generic association rules from a dataset by discretizing this latter into two binary data matrices. 
The performance of the \textit{NBic-ARM} algorithm is assessed on three real-life DNA microarray datasets. These experimentations show that \textit{NBic-ARM} allows extracting high quality biclusters. These biclusters have been evaluated with \textit{GO} annotations which verifies the biological significance of biclusters. The obtained results have shown the fullness of our proposed biclustering algorithm.

\hspace{+0.5 cm}In our last contribution we have introduced the NBF biclustering algorithm, a new FCA-based biclustering method for discovering negatively correlated genes from gene expression data. Our approach consists in extracting formal concepts from a dataset after having discretized it into two binary data matrices, a positive ($ \mathcal M3^{+}$) matrix and a negative ($ \mathcal M3^{-}$) one. These matrices allow us to discover negative correlation genes.The discretization of these latter is used to extract formal concepts. The resulting formal concepts represent formal concepts filtered by a stability measure in order to remove the non-coherent concepts. The performance of our algorithm has been assessed on real-life DNA microarray datasets. These experimentations demonstrate that the NBF permits extracting high-quality negatively correlated biclusters with respect to statistical and biological significance. 

\hspace{+0.5 cm} In the case of negative correlations algorithms, and seeing how the extraction process is done on 2 binary matrices to locate negative correlations, the computational complexity ventures to exponential in function of the number of lines (genes) and columns (conditions) of the original matrix.

\hspace{+0.5 cm} As we conclude this dissertation, some interesting future work has to be mentioned. 
\section*{Future research}
The obtained results in this thesis opens several perspectives. In this section, we present some promising future research paths from which we quote: 
\begin{itemize}
\item Future work will focus on the issue of extracting biclusters from big datasets. In fact, big data mining is a new challenging task since computational requirements are difficult to provide. An interesting solution is to exploit parallel frameworks such as MapReduce \cite{Dean2008} that offers the opportunity to make powerful computing and storage. The major drawback that exists in microarray data analysis is the curse of dimensionality problem. To deal with the above mentioned issues, MapReduce has been designed to support the concept of distributed computing, turning out to be an efficient platform for parallel data mining for large datasets, where the data are distributed on various nodes, which allows making powerful computing and storage units on top of ordinary machines. In this context, mining biclusters from gene expression from big real-life datasets thanks to the MapReduce environment is an up-to-date challenging mining task. 
\item Another fruitful perspective consists in addressing the extensions of the concepts of biclusters and formal concepts to those of triclusters and triconcepts \cite{Ignatov2015}. The basic idea is the extension of FCA to ternary relations: An object has an attribute under a condition \textit{i.e.} Triadic Concept Analysis. The latter provides a powerful mathematical framework for biclustering.
\item Other avenues of future work also concern the extraction of biclusters by integrating biological knowledge during the extraction process. Basically, some a priori biological information is introduced as an input and the search process has a bias to find better biclusters. It can be said that the integration of biological information from several sources is an up-to-date challenge in bioinformatics \cite{Aguilar-Ruiz2005}. In this context, we can use other data sources, not only information from GO. For instance, many data sources can be merged to integrate biological knowledge as protein-protein interaction networks, genome-wide binding data and information from the literature and not only information from GO. 

\end{itemize} 

\section*{Publication list}
The following published, in press or accepted papers are partial outputs of this thesis.
\begin{enumerate}
\subsection*{International journals (1)}

\item  Houari, A., Ayadi, W. \& Ben Yahia, S. Int. J. Mach. Learn. \& Cyber. (2018). https://doi.org/10.1007/s13042-018-0794-9 [IF:1.699, SJR:0.685]

\subsection*{International conferences (5)}
\item NBF: An FCA-based Algorithm to Identify Negative Correlation Biclusters of DNA Microarray Data. (Accepted in AINA 2018) [CORE: B]

\item Houari, A., Ayadi, W., \& Yahia, S. B. (2017). Mining Negative Correlation Biclusters from Gene Expression Data using Generic Association Rules. Procedia Computer Science, 112, 278-287. [CORE: B]

\item Houari, A., Ayadi, W., \& Yahia, S. B. (2016). Biclustering Gene Expression Data: Pattern-Mining-Based Approaches . Proceedings ICICDS’2016. 

\item Houari, A., Ayadi, W., \& Yahia, S. B. (2015). Discovering Low Overlapping Biclusters in Gene Expression Data Through Generic Association Rules. In Model and Data Engineering (pp. 139-153). Springer International Publishing.

\item Houari, A., Ayadi, W., \& Yahia, S. B. (2015). Bi-Clustering Algorithm Using Formal Concept Analysis. Proceedings ICNTC'2015.
\end{enumerate}


\bibliographystyle{apalike}
\bibliography{AminaRefs,Ref8,Ref8neg,Ref8 copy}

\end{document}